\documentclass[mnsc,nonblindrev]{informs3}

\OneAndAHalfSpacedXI 

\usepackage{natbib}
 \bibpunct[, ]{(}{)}{,}{a}{}{,}%
 \def\bibfont{\small}%

\usepackage{amsmath,amssymb,bm,comment,mathrsfs}
\usepackage[ruled]{algorithm}
\usepackage{algorithmicx,algpseudocode,subfigure,caption}    

\newcommand{\vx}{x}
\newcommand{\vy}{y}

\newcommand{\MN}{\mathcal{MN}}

\newcommand{\vep}{\varepsilon}
\newcommand{\E}{\mathbb{E}}

\newcommand{\R}{\mathbb{R}}
\newcommand{\regret}{\mathcal{R}}
\newcommand{\N}{\mathcal{N}}
\newcommand{\J}{\mathcal{J}}
\newcommand{\T}{\mathcal{T}}

\newcommand{\cP}{\mathcal{P}}
\newcommand{\cH}{\mathcal{H}}

\newcommand{\tr}{\textnormal{tr}}
\newcommand{\rev}{\mathtt{REV}}
\newcommand{\eg}{\textit{e.g.}}
\newcommand{\ie}{\textit{i.e.}}

\newcommand{\oracle}{\textnormal{TS}}
\newcommand{\md}{\textnormal{MD}}
\newcommand{\mts}{\textnormal{MDP}}
\newcommand{\mpdp}{\texttt{Meta-DP}~algorithm}
\newcommand{\covmpdp}{\texttt{Meta-DP++}~algorithm}

\TheoremsNumberedThrough     
\ECRepeatTheorems

\EquationsNumberedThrough    

\MANUSCRIPTNO{MS-0001-1922.65}
\allowdisplaybreaks
\begin{document}



\RUNTITLE{Meta Dynamic Pricing}

\TITLE{\Large Meta Dynamic Pricing: Transfer Learning Across Experiments}

\ARTICLEAUTHORS{%
\AUTHOR{Hamsa Bastani}
\AFF{Operations, Information and Decisions, Wharton School,  \EMAIL{hamsab@wharton.upenn.edu}}
\AUTHOR{David Simchi-Levi}
\AFF{Institute for Data, Systems, and Society, Massachusetts Institute of Technology, \EMAIL{dslevi@mit.edu}}
\AUTHOR{Ruihao Zhu}
\AFF{Institute for Data, Systems, and Society, Massachusetts Institute of Technology, \EMAIL{rzhu@mit.edu}} 
} 

\ABSTRACT{%
We study the problem of learning shared structure \emph{across} a sequence of dynamic pricing experiments for related products. We consider a practical formulation where the unknown demand parameters for each product come from an unknown distribution (prior) that is shared across products. We then propose a meta dynamic pricing algorithm that learns this prior online while solving a sequence of Thompson sampling pricing experiments (each with horizon $T$) for $N$ different products. Our algorithm addresses two challenges: (i) balancing the need to learn the prior (\emph{meta-exploration}) with the need to leverage the estimated prior to achieve good performance (\emph{meta-exploitation}), and (ii) accounting for uncertainty in the estimated prior by appropriately ``widening" the estimated prior as a function of its estimation error. We introduce a novel prior alignment technique to analyze the regret of Thompson sampling with a mis-specified prior, which may be of independent interest. Unlike prior-independent approaches, our algorithm's meta regret grows sublinearly in $N$, demonstrating that the price of an unknown prior in Thompson sampling can be negligible in experiment-rich environments (large $N$). Numerical experiments on synthetic and real auto loan data demonstrate that our algorithm significantly speeds up learning compared to prior-independent algorithms.
}%


\maketitle

%


\section{Introduction}
Experimentation is popular on online platforms to optimize a wide variety of elements such as search engine design, homepage promotions, and product pricing. This has led firms to perform an increasing number of experiments, and several platforms have emerged to provide the infrastructure for these firms to perform experiments at scale \citep[see, \eg,][]{Optimizely}. State-of-the-art techniques in these settings employ bandit algorithms (\eg, Thompson sampling), which seek to adaptively learn treatment effects while optimizing performance \textit{within} each experiment \citep{thompson1933likelihood, scott2015multi}. However, the large number of related experiments begs the question: can we transfer knowledge \textit{across} experiments?

We study this question for Thompson sampling algorithms in dynamic pricing applications that involve a large number of related products. Dynamic pricing algorithms enable retailers to optimize profits by sequentially experimenting with product prices, and learning the resulting customer demand \citep{kleinberg2003value, besbes2009dynamic}. Such algorithms have been shown to be especially useful for products that exhibit relatively short life cycles \citep{ferreira2015analytics}, stringent inventory constraints \citep{xu2018designing}, strong competitive effects \citep{fisher2017competition}, or the ability to offer personalized coupons/pricing \citep{zhang2017does, ban2017personalized}. In all these cases, the demand of a product is estimated as a function of the product's price (chosen by the decision-maker) and a combination of exogenous features as well as product-specific and customer-specific features. Through carefully chosen price experimentation, the decision-maker can learn the price-dependent demand function for a given product, and choose an optimal price to maximize profits \citep{qiang2016dynamic, cohen2016feature, javanmard2016dynamic}. Dynamic pricing algorithms based on Thompson sampling have been shown to be particularly successful in striking the right balance between exploring (learning the demand) and exploiting (offering the estimated optimal price), and are widely considered to be state-of-the-art \citep{thompson1933likelihood, agrawal2013thompson, RVR14, FLW18}.

The decision-maker typically runs a separate pricing experiment (\ie, dynamic pricing algorithm) for each product (or for a set of simultaneously-offered products). However, this approach can waste valuable samples re-discovering information that could have been learned from previously-offered related products. For example, students may be more price-sensitive than general customers; as a result, many firms such as restaurants, retailers and movie theaters offer student discounts. This implies that the coefficient of student-specific price elasticity in the demand function is positive for many products (although the specific value of the coefficient likely varies across products). Similarly, winter clothing may have higher demand in the fall and lower demand at the end of winter. This implies that the demand functions of winter clothing may have similar coefficients for the features indicating time of year. In general, there may even be complex correlations between coefficients of the demand functions of products that are shared. For example, the price-elasticities of products are often negatively correlated with their demands, \ie,  customers are willing to pay higher prices when the demand for a product is high. When offering \textit{multiple} products simultaneously, one must additionally learn cross-product price elasticities in the demand function (to model substitution effects), which may also exhibit patterns that can be learned from substitution patterns of related products in historical data. For example, substitution effects may be stronger between more similar products, or among more price-sensitive customers like students.

Thus, one may expect that the demand functions for related products may share some (a priori unknown) common structure, which can be learned \textit{across} products. Note that the demand functions are unlikely to be exactly the same, so a decision-maker would still need to conduct separate pricing experiments for each product. However, accounting for shared structure during these experiments may significantly speed up learning per product (or per set of products, if offering multiple products simultaneously), thereby improving profits.

In this paper, we propose an approach to learn shared structure across pricing experiments. We begin by noting that the key (and only) design decision in Thompson sampling methods is the Bayesian prior over the unknown parameters. This prior captures shared structure of the kind we described above --- \eg, the mean of the prior on the student-specific price-elasticity coefficient may be positive with a small standard deviation. It is well known that choosing a good (bad) prior significantly improves (hurts) the empirical performance of the algorithm \citep{chapelle2011empirical, honda2014optimality, liu2015prior, russo2018tutorial}. However, the prior is typically unknown in practice, particularly when the decision-maker faces a cold start. While the decision-maker can use a \textit{prior-independent} algorithm \citep{agrawal2013thompson}, such an approach achieves poor empirical performance due to over-exploration; we demonstrate a substantial gap between the prior-independent and prior-dependent approaches in our experiments on synthetic and real data. In particular, knowledge of the correct prior enables Thompson sampling to appropriately balance exploration and exploitation \citep{RVR14}. Thus, the decision-maker needs to learn the true prior (\ie, shared structure) \textit{across} products to achieve good performance. We propose a meta dynamic pricing algorithm that efficiently achieves this goal.

We first formulate the problem of learning the true prior online while solving a sequence of pricing experiments for different products. Our meta dynamic pricing algorithm requires two key ingredients. First, for each product, we must balance the need to learn about the prior (``meta-exploration'') with the need to leverage the prior to achieve strong performance for the current product (``meta-exploitation''). In other words, our algorithm balances an additional exploration-exploitation tradeoff across price experiments. Second, a key technical challenge is that finite-sample estimation errors of the prior may significantly impact the performance of Thompson sampling for any given product. In particular, vanilla Thompson sampling may fail to converge with an incorrect prior; as a result, directly using the estimated prior across products can result in poor performance. To this end, we introduce a novel ``prior alignment" technique to analyze the regret of Thompson sampling with a mis-specified prior, which may be of independent interest.

Using our alignment technique, we show surprisingly that \textit{despite} prior mis-specification, greedy updating of the prior is sufficient to learn effectively across pricing experiments when the prior covariance is known. However, when the prior has an unknown covariance matrix, it is beneficial to widen the estimated prior covariance by a term that is a function of the prior's estimated finite-sample error. Thus, we use a more conservative approach (a wide prior) for earlier products when the prior is uncertain; over time, we gain a better estimate of the prior, and can leverage this knowledge for better empirical performance. Our algorithm provides an exact prior correction path over time to achieve strong performance guarantees across all pricing problems. We prove that, when using our algorithm, the price of an unknown prior for Thompson sampling is negligible in experiment-rich environments (\ie, as the number of products grows large).

\subsection{Related Literature}

Experimentation is widely used to optimize decisions in a data-driven manner. This has led to a rich literature on bandits and A/B testing \citep{lai1985asymptotically, auer2002using, dani2008stochastic, rusmevichientong2010linearly, besbes2014stochastic, johari2015always, bhat2019near}. This literature primarily proposes learning algorithms for a single experiment, while our focus is on meta-learning across experiments. Meta-learning can take the form of constructing an empirical Bayesian prior \citep{raina2006constructing, anderer2019adaptive}, data pooling \citep{GuptaK20}, or leveraging low-dimensional structure between problems \citep{bastani2020predicting}. We take an empirical Bayesian approach to sequential decision-making. While there has been some prior work on meta-learning in bandits \citep{hartland2006multi, maes2012meta, wang2018regret, sharaf2019meta} and more generally in reinforcement learning \citep{finn2017model, finn2018probabilistic, yoon2018bayesian}, these papers only provide heuristics for learning exploration strategies given a fixed set of past problem instances. They do not prove any theoretical guarantees on the performance or regret of the meta-learning algorithm. To the best of our knowledge, our paper is the first to propose a meta-learning algorithm in a bandit setting with provable regret guarantees.

We study the specific case of dynamic pricing, which aims to learn an unknown demand curve in order to optimize profits. We focus on dynamic pricing because meta-learning is particularly important in this application, \eg, online retailers such as Rue La La may run numerous pricing experiments for related fashion products. We believe that a similar approach could be applied to multi-armed or contextual bandit problems, in order to inform the prior for Thompson sampling across a sequence of related bandit problems.

Dynamic pricing has been found to be especially useful in settings with short life cycles or limited inventory \citep[\eg, fast fashion or concert tickets, see][]{ferreira2015analytics, xu2018designing}, among online retailers that constantly monitor competitor prices and adjust their own prices in response \citep{fisher2017competition}, or when prices can be personalized based on customer-specific price elasticities \citep[\eg, through personalized coupons, see][]{zhang2017does}. Several papers have designed near-optimal dynamic pricing algorithms for pricing a product by balancing the resulting exploration-exploitation tradeoff \citep{kleinberg2003value, besbes2009dynamic, araman2009dynamic, farias2010dynamic, harrison2012bayesian, broder2012dynamic, den2013simultaneously, keskin2014dynamic}. Recently, this literature has shifted focus to pricing policies that dynamically optimize the offered price with respect to exogenous features \citep{qiang2016dynamic, cohen2016feature, javanmard2016dynamic} as well as customer-specific features \citep{ban2017personalized,ElmachtoubGH20}. We adopt the linear demand model proposed by \cite{ban2017personalized}, which allows for feature-dependent heterogeneous price elasticities.

When sellers offer multiple products simultaneously, one may wish to perform price experiments \textit{jointly} on a set of products to capture substitution effects or overlapping inventory constraints \citep{keskin2014dynamic, agrawal2014bandits, FLW18}. However, in these papers, price experimentation is still performed independently on the current set of products, and any learned parameter knowledge is not shared across future sets of products to inform future demand learning. In contrast, we propose a meta dynamic pricing algorithm that learns the distribution of unknown parameters of the demand function across products. While we focus largely on the single-product setting for ease of exposition, we show how our algorithm and theoretical results carry over straightforwardly for multi-product settings with substitution effects; in fact, transfer learning from historical data may be even more valuable in these settings since the number of parameters (\eg, cross-product elasticities) to learn is much larger.

Our learning strategy is based on Thompson sampling, which is widely considered to be state-of-the-art for balancing the exploration-exploitation tradeoff \citep{thompson1933likelihood}. Several papers have studied the sensitivity of Thompson sampling to prior misspecification. For example, \cite{honda2014optimality} show that Thompson sampling still achieves the optimal theoretical guarantee with an incorrect but uninformative prior, but can fail to converge if the prior is not sufficiently conservative. \cite{liu2015prior} provide further support for this finding by showing that the performance of Thompson sampling for any given problem instance depends on the probability mass (under the provided prior) placed on the underlying parameter; thus, one may expect that Thompson sampling with a more conservative prior (\ie, one that places nontrivial probability mass on a wider range of parameters) is more likely to converge when the true prior is unknown. It is worth noting that \cite{agrawal2013thompson} and \cite{bubeck2013prior} propose a \textit{prior-independent} form of Thompson sampling, which is guaranteed to converge to the optimal policy even when the prior is unknown by conservatively increasing the variance of the posterior over time. However, the use of a more conservative prior creates a significant cost in empirical performance \citep{chapelle2011empirical}. For instance, \cite{bastani2020mostly} empirically find through simulations that the conservative prior-independent Thompson sampling is significantly outperformed by vanilla Thompson sampling \textit{even} when the prior is misspecified.\footnote{We provide some theoretical support for this finding, since we show that limited prior mis-specification does not affect the rate of convergence (\eg, when the prior covariance is known but the mean is unknown).} We empirically find, through experiments on synthetic and real datasets, that learning and leveraging the prior can yield much better performance compared to a prior-independent approach. As such, the choice of prior remains an important design choice in the implementation of Thompson sampling \citep{russo2018tutorial}. We propose a meta-learning algorithm that learns the prior across pricing experiments on related products to attain better performance. We also empirically demonstrate that a naive approach of greedily using the updated prior performs poorly when the prior covariance is unknown, since it may cause Thompson sampling to fail to converge to the optimal policy for some products. Instead, our algorithm gracefully tunes the width of the estimated prior as a function of the uncertainty in the estimate over time.

\subsection{Main Contributions}

We highlight our main contributions below:
\begin{enumerate}
	\item \textit{Model:} We formulate our problem as a sequence of $N$ different dynamic pricing problems, each with horizon $T$. Importantly, the unknown parameters of the demand function for each product are drawn i.i.d. from a shared (unknown) multivariate Gaussian prior.
	\item \textit{Algorithm:} We propose two meta-learning pricing policies, \texttt{Meta-DP} and \texttt{Meta-DP++}. The former learns only the mean of the prior, while the latter learns both the mean and the covariance of the prior across products. Both algorithms balance the need to learn the prior (\emph{meta-exploration}) with the need to leverage the current estimate of the prior to achieve good performance (\emph{meta-exploitation}). \texttt{Meta-DP++} additionally accounts for uncertainty in the estimated prior by conservatively widening the prior as a function of its estimation error.
	\item \textit{Theory:} Unlike standard approaches, our algorithm can leverage shared structure across products to achieve regret that scales sublinearly in the number of products $N$. We prove upper bounds $\tilde{O}(d^2\sqrt{NT}+d^3\sqrt{T})$ and $\tilde{O}(\min\{d^2 N T^{\frac{1}{2}},~ d^4 N^{\frac{1}{2}}T^{\frac{3}{2}}\})=\tilde{O}(d^3(NT)^{\frac{5}{6}})$ on the meta regret of \texttt{Meta-DP} and \texttt{Meta-DP++} respectively. In both cases, our meta-learning approach matches the performance of prior-independent algorithms for small $N$, and outperforms them in experiment-rich experiments (\ie, when $N = \tilde{\Omega}(d)$ and $N = \tilde{\Omega}(d^4 T^2)$ respectively). A key ingredient of our analysis is a ``prior alignment" proof technique that may be of general interest for analyzing the regret of mis-specified Thompson Sampling instances.
	\item \textit{Numerical Experiments:} We demonstrate on both synthetic and real auto loan data that our approach significantly speeds up learning compared to ignoring shared structure (\ie, using prior-independent Thompson sampling).
\end{enumerate}

\section{Problem Formulation} \label{sec:problem}

For ease of exposition, we primarily focus on a seller offering a \textit{single} product at a time. Our approach and results generalize straightforwardly when \textit{multiple} products are offered simultaneously, where a seller must also learn cross-product elasticities to capture substitution effects (see extension in Appendix \ref{sec:multiple_prod}).

\paragraph{Notation:} Throughout the paper, all vectors are column vectors by default. We define $[n]$ to be the set $\{1,2,\ldots,n\}$ for any positive integer $n.$  We use $\|\vx\|_u$ to denote the $\ell_u$ norm of a vector $x\in\R^d,$ but we often omit the subscript when we refer to the $\ell_2$ norm. For a matrix $X\in\R^{d\times d}$ $\|X\|_{op}:=\max_{v\in\R^d:\|v\|=1}|v^{\top}Xv|$ is the operator norm of $X.$ For a positive definite matrix $A\in \R^{d\times d}$ and vectors $\vx,\vy\in\R^d$, let $\|\vx\|_A$ denote the matrix norm $\sqrt{\vx^{\top}A\vx}$ and $\langle \vx, \vy \rangle$ denote the inner product $\vx^\top\vy$. For two matrices $A$ and $B,$ we use $A\otimes B$ to denote their Kronecker product. We also denote $x\vee y$ and $x\wedge y$ as the maximum and minimum between $(x,y)\in\R,$ respectively. We use the standard notation $O(\cdot), \Omega(\cdot)$ and $\Theta(\cdot)$ to characterize the asymptotic growth rate of a function \citep{CLRS09}; when logarithmic factors are omitted, we use $\tilde{O}(\cdot), \tilde{\Omega}(\cdot)$ and $\tilde{\Theta}(\cdot)$. Finally, let $\lambda_{\min}(\cdot)$ and $\lambda_{\max}(\cdot)$ denote the minimum and maximum eigenvalues of a matrix respectively.

\subsection{Model}

We first describe the classical dynamic pricing formulation for a single product; we then formalize our meta-learning formulation over a sequence of $N$ products. 

\paragraph{Classical Formulation:} Consider a seller who offers a single product over a selling horizon of $T$ periods. The seller can dynamically adjust the offered price in each period. At the beginning of each period $t \in [T]$, the seller observes a random feature vector (capturing exogenous and/or customer-specific features) that is independently and identically distributed from an unknown distribution. Upon observing the feature vector, the seller chooses a price for that period. The seller then observes the resulting demand, which is a noisy function of both the observed feature vector and the chosen price. The seller's revenue in each period is given by the chosen price multiplied by the corresponding realized demand. The goal in this setting is to develop a policy $\pi$ that maximizes the seller's cumulative revenue by balancing exploration (learning the demand function) with exploitation (offering the estimated revenue-maximizing price). 

\paragraph{Meta-learning Formulation:} We consider a seller who sequentially offers $N$ related products, each with a selling horizon of $T$ periods. For simplicity, a new product is not introduced until the life cycle of the previous product ends.\footnote{We model epochs as fully sequential for simplicity; if epochs overlap, we would need to additionally model a customer arrival process for each epoch. Our algorithms straightforwardly generalize for overlapping epochs; see remark in \S \ref{ssec:remarks}.} We call each product's life cycle an \textit{epoch}, \ie, there are $N$ epochs that last $T$ periods each. Each product (and corresponding epoch) is associated with a different (unknown) demand function, and constitutes a different instance of the classical dynamic pricing problem described above. We now formalize the problem.

In epoch $i \in [N]$ at time $t \in [T]$, the seller observes a random feature vector $ \vx_{i,t} \in \R^d$, which is independently and identically distributed from a known distribution $\cP_i$. She then chooses a price $p_{i,t}$ for that period. Based on practical constraints, we will assume that the allowable price range is bounded across periods and products, \ie, $p_{i,t} \in [p_{\min}, p_{\max}]$ and $0 < p_{\min} < p_{\max} < \infty$. The seller then observes the resulting induced demand
\[ D_{i,t}(p_{i,t},\vx_{i,t}) = \langle\alpha_i, \vx_{i,t}\rangle + p_{i,t} \langle\beta_i, \vx_{i,t}\rangle + \vep_{i,t} \,,\]
where $\alpha_i \in \R^{d}$ and $\beta_i \in \R^{d}$ are unknown fixed constants throughout epoch $i$, and $\vep_{i,t}\sim\N(0,\sigma^2)$ is i.i.d. Gaussian noise with variance $\sigma^2.$ This demand model was recently proposed by \cite{ban2017personalized}, and captures several salient aspects. In particular, the observed feature vector $\vx_{i,t}$ in period $t$ determines both the baseline demand (through the parameter $\alpha_i$) and the price-elasticity of the demand (through the parameter $\beta_i$) of product $i$.

\begin{example}[Rue La La] \label{ex:ruelala} Rue La La sells a limited set of new products in multi-day ``events" \citep{ferreira2015analytics}. In this case, $T$ is the number of price changes during the event (events are typically 1-4 days, and prices are updated no more than a few times a day), $N$ is the number of events offered so far by the seller (note that $N \gg T$), and $K$ is the number of simultaneously-offered products in an event. For ease of exposition, we primarily consider $K=1$, but Appendix \ref{sec:multiple_prod} provides a straightforward extension to general values of $K$, accounting for substitution effects.
\end{example}

\begin{remark}[Alternative Demand Models]
Our demand model utilizes a continuous outcome variable, motivated by the setting where many customers simultaneously view the same product with the same price in a given time unit. One can alternatively modify the demand model to follow a generalized linear model (\eg, logistic) to consider a binary purchase outcome variable for \textit{each} customer. Our proposed algorithms easily generalize by appropriately modifying our Bayesian posterior update rules; however, we restrict our regret analysis to the linear case since OLS Bayesian posterior updates have a closed form, yielding a tractable analysis.
\end{remark}

\paragraph{Shared Structure:} For ease of notation, we denote $\theta_i =\begin{pmatrix}
\alpha^{\top}_i&~ \beta^{\top}_i
\end{pmatrix}^{\top} \in \R^{2d}$; following the classical formulation of dynamic pricing, $\theta_i$ is the unknown parameter vector that must be learned within a given epoch in order for the seller to maximize her revenues over $T$ periods. When there is no shared structure between the $\{\theta_i \}_{i=1}^N$, our problem reduces to $N$ independent dynamic pricing problems.

However, we may expect that related products share a similar potential market, and thus may have some shared structure that can be learned from previously offered products. We model this relationship by positing that the product demand parameter vectors $\{\theta_i \}_{i=1}^N$ are independent and identically distributed draws from a common unknown distribution, \ie, $\theta_i \sim \N(\theta_*, \Sigma_*)$ for each $i \in [N]$.\footnote{Following the literature on Thompson sampling, we consider a multivariate Gaussian distribution since the posterior has a simple closed form, thereby admitting a tractable theoretical analysis. When implementing such an algorithm in practice, more complex distributions can be considered \citep[\eg, see discussion in][]{russo2018tutorial}.} As discussed earlier, knowledge of the distribution over the unknown demand parameters can inform the prior for Thompson sampling, thereby avoiding the need to use a conservative prior that can result in poor empirical performance \citep{honda2014optimality, liu2015prior}. The mean of the shared distribution $\theta_*$ is unknown; we will consider settings where the covariance of this distribution $\Sigma_*$ is known and unknown.  We propose using meta-learning to learn this distribution from past epochs to inform and improve the current product's pricing strategy.

\begin{remark}[Product Features] \label{rem:prod-features}
A complementary form of shared structure can be captured through product features. However, even after conditioning on observed product features, the demand functions for two products may behave very differently, \eg, two black dresses may cater to very different types of customers or have very different price elasticities due to attributes like fit or design that may be hard to capture as features. To capture product-specific (\ie, SKU-level) demand behaviors, we allow the \textit{coefficients} of the demand function (\eg, price-elasticity) to differ.
\end{remark}

\subsection{Assumptions}\label{sec:assumption}

We now describe some mild assumptions on the parameters of the problem for our regret analysis.

\begin{assumption}[Boundedness]
	\label{assumption:x_norm}
	The support of the features are bounded, \ie,
	\begin{align*}
	\forall i\in[N] \,, \forall t\in[T] \quad\left\|\vx_{i,t}\right\|\leq x_{\max}.
	\end{align*}
	Furthermore, there exists a positive constant $S$ such that $\|\theta_*\|\leq S.$
\end{assumption}
Our first assumption is that the observed feature vectors $\{\vx_{i,t}\}$ as well as the mean of the product demand parameters $\theta_*$ are bounded. This is a standard assumption made in the bandit and dynamic pricing literature, ensuring that the expected regret at any time step is bounded. This is likely satisfied since features and outcomes are typically bounded in practice.

\begin{assumption}[Positive-Definite Feature Covariance]
\label{assumption:pos-def}
	The minimum eigenvalue of the feature covariance matrix $\E_{\vx_{i,t} \sim \cP_i}\left[\vx_{i,t}\vx_{i,t}^{\top}\right]$ in every epoch $i \in [N]$ is lower bounded by some positive constant $\lambda_0$, \ie,
	\begin{align*}
	\min_{i \in [N]} ~\lambda_{\min}\left(\E_{\vx_{i,t} \sim \cP_i} \left[\vx_{i,t}\vx_{i,t}^{\top}\right]\right) ~\geq~ \lambda_0 \,.
	\end{align*}
\end{assumption}
Our second assumption imposes that the covariance matrix of the observed feature vectors $\E\left[\vx_{i,t}\vx_{i,t}^{\top}\right]$ in every epoch is positive-definite. This is a standard assumption for the convergence of OLS estimators; in particular, our demand model is linear, and therefore requires that no features are perfectly collinear in order to identify each product's true demand parameters.

\begin{assumption}[Positive-Definite Prior Covariance]
	\label{assumption:Sigma}
	The maximum and minimum eigenvalues of $\Sigma_*$ are upper and lower bounded by positive constants $\overline{\lambda}$ and $\underline{\lambda},$ respectively \ie,
	\begin{align*}
	\lambda_{\max}\left(\Sigma_*\right)\leq\overline{\lambda},\quad\lambda_{\min}\left(\Sigma_*\right)\geq\underline{\lambda} \,.
	\end{align*}
\end{assumption}
Our final assumption imposes that the covariance matrix of the random product demand parameter $\theta$ is also positive-definite and bounded. Again, this assumption ensures that each product's true demand parameter is identifiable using standard OLS estimators.

\subsection{Background on Thompson Sampling with Known Prior}

In this subsection, we consider the setting where the true prior $\N\left(\theta_*,\Sigma_*\right)$ over the unknown product demand parameters is \textit{known}. This setting will inform our definition of the meta oracle and meta regret in the next subsection. When the prior is known, a natural candidate policy for minimizing Bayes regret is the Thompson sampling algorithm \citep{thompson1933likelihood}. The Thompson sampling algorithm adapted to our dynamic pricing setting for a single epoch $i \in [N]$ is formally given in Algorithm \ref{alg:oracle} below. Since the prior is known, there is no additional shared structure to exploit across products, so we can treat each epoch independently.

We denote TS$\left(\N\left(\theta_*,\Sigma_*\right),\lambda_e\right),$ as the Thompson sampling algorithm with prior $\N\left(\theta_*,\Sigma_*\right)$ and a positive input parameter $\lambda_e$ for initialization. 
In line with pricing algorithms in the literature \citep[see, \eg,][]{keskin2014dynamic, ban2017personalized}, to ensure that we can obtain a well-defined OLS estimate of the underlying parameter at the end of an epoch, our algorithm initially performs random price exploration (alternating between $p_{\min}$ and $p_{\max}$) until the Fisher information matrix 
$V_{i,t}=\sum_{s=1}^{t}\begin{pmatrix}
x^{\top}_{i,s}&~p_{i,s}x_{i,s}^{\top}
\end{pmatrix}^{\top}\begin{pmatrix}
x^{\top}_{i,s}&~p_{i,s}x_{i,s}^{\top}
\end{pmatrix}$
has minimum eigenvalue of at least $\lambda_e$. Let $\T_i$ be the (random) length of this initialization period in epoch $i,$
\begin{align}
\T_i ~=~ \argmin_{t}\lambda_{\min}\left(V_{i,t}\right) ~\geq~ \lambda_e \,.
\end{align}
We show that $\T_i = \tilde{O}(1)$ with high probability (see Lemma \ref{lemma:ini} in Appendix \ref{sec:theorem:oracle}), and therefore this initialization period forms a negligible portion of the epoch.

For each time step after initialization, $t \geq \T_i+1$, the algorithm (1) samples the unknown product demand parameters $\mathring{\theta}_{i,t}=\left[\mathring{\alpha}_{i,t};\mathring{\beta}_{i,t}\right]$ from the posterior $\N\left(\theta^{\oracle}_{i,t},\Sigma^{\oracle}_{i,t}\right)$, and (2) solves and offers the resulting optimal price based on the demand function given by the sampled parameters
\begin{align}
p_{i,t}^{\oracle} = \argmax_{p\in\left[p_{\min},p_{\max}\right]} ~ p \cdot \left\langle\mathring{\alpha}_{i,t}, \vx_{i,t}\right\rangle+ p^2 \cdot \left\langle\mathring{\beta}_{i,t},\vx_{i,t}\right\rangle \,.
\end{align}
Upon observing the actual realized demand $D_{i,t}\left(p_{i,t}^{\oracle},\vx_{i,t}\right)$, the algorithm computes the posterior $\N\left(\theta^{\oracle}_{i,t+1},\Sigma^{\oracle}_{i,t+1}\right)$ for round $t+1$. Specifically, using the update rule for Bayesian linear regression \citep{Bishop06} and letting $m^{\oracle}_{i,t}=(\vx^{\top}_{i,t}, p_{i,t}^{\oracle}\vx^{\top}_{i,t})^{\top}$, the posterior at time $t$  is
\begin{align*}
\theta^{\oracle}_{i,t}=&\left(\Sigma_*^{-1}+\sigma\sum_{s=1}^{t-1}m_{i,s}^{\oracle}(m_{i,s}^{\oracle})^{\top}\right)^{-1}\left(\Sigma_*^{-1}\theta_*+\sigma\sum_{s=1}^{t-1}m^{\oracle}_{i,s}D_{i,s}\right) \,, \quad
\Sigma^{\oracle}_{i,t}=\left(\Sigma_*^{-1}+\sigma\sum_{s=1}^{t-1}m^{\oracle}_{i,s}(m^{\oracle}_{i,s})^{\top}\right)^{-1} \,.
\end{align*}
The same algorithm is applied independently to each epoch $i \in [N]$.

\begin{algorithm}[!ht]
\SingleSpacedXI
	\caption{TS$(\N\left(\theta_*,\Sigma_*\right),\lambda_e):$ Thompson Sampling Algorithm}
	\label{alg:oracle}
	\begin{algorithmic}[1]
		\State \textbf{Input:} The prior mean vector $\theta_*$ and covariance matrix $\Sigma_*,$ the index $i$ of epoch, the length of each epoch $T,$ the noise parameter $\sigma,$ exploration parameter $\lambda_e.$
	\State \textbf{Initialization:} $t\leftarrow1,\left(\theta^{\oracle}_{i,t},\Sigma^{\oracle}_{i,t}\right)\leftarrow\left(\theta_*,\Sigma_*\right)$.
	\While{$\lambda_{\min}\left(\sum_{s=1}^{t-1}\begin{pmatrix}
			x^{\top}_{i,s}&~p_{i,s}x^{\top}
		\end{pmatrix}^{\top}\begin{pmatrix}
			x^{\top}_{i,s}&~p_{i,s}x^{\top}
		\end{pmatrix}\right)<\lambda_e$}
	\State{Observe feature vector $\vx_{i,t},$ and offer price $p^{\oracle}_{i,t}\leftarrow\begin{cases}
		p_{\max}&\text{if $t$ is even,}\\
		p_{\min}&\text{otherwise.}
		\end{cases}$}
	\State{Observe demand $D_{i,t}\left(p^{\oracle}_{i,t},\vx_{i,t}\right),$ and compute the posterior $\N\left(\theta^{\oracle}_{i,t+1},\Sigma^{\oracle}_{i,t+1}\right).$}
		\State{$t\leftarrow t+1$}
	\EndWhile
	\While {$t\leq T$}
	\State{Observe feature vector $\vx_{i,t}.$}
	\State{Sample parameter $\mathring{\theta}_{i,t}\leftarrow\left[\mathring{\alpha}_{i,t};\mathring{\beta}_{i,t}\right]\sim\N\left(\theta^{\oracle}_{i,t},\Sigma^{\oracle}_{i,t}\right).$}
	\State{$
		p_{i,t}^{\oracle}\leftarrow\argmax_{p\in\left[p_{\min},p_{\max}\right]} ~ p \cdot \left\langle\mathring{\alpha}_{i,t}, \vx_{i,t}\right\rangle+ p^2 \cdot \left\langle\mathring{\beta}_{i,t},\vx_{i,t}\right\rangle.$}
	\State{Observe demand $D_{i,t}\left(p^{\oracle}_{i,t},\vx_i\right),$ and compute the posterior $\N\left(\theta^{\oracle}_{i,t+1},\Sigma^{\oracle}_{i,t+1}\right).$}
		\State{$t\leftarrow t+1$}
	\EndWhile
	\end{algorithmic}
\end{algorithm}

As evidenced by the large literature on the practical success of Thompson sampling \citep{chapelle2011empirical, RVR14, FLW18}, Algorithm \ref{alg:oracle} is a very attractive choice for implementation in practice. 

Algorithm \ref{alg:oracle} attains a strong performance guarantee under the classical formulation compared to an \textit{oracle} that knows all $N$ product demand parameters $\{\theta_i\}_{i=1}^N$ in advance. In particular, the oracle would offer the expected optimal price in each period $t \in [T]$ in epoch $i \in [N]$, \ie,
\begin{align} \label{eq:oracle-price}
\nonumber p^*_{i,t} &= {\arg\max}_{p  \in [p_{\min}, p_{\max}]} ~p \cdot \E_\vep[ D_{i,t}(p, \vx_{i,t})]  \\
&= {\arg\max}_{p  \in [p_{\min}, p_{\max}]} ~ p \langle \alpha_i, \vx_{i,t} \rangle + p^2 \langle \beta_i, \vx_{i,t} \rangle \,.
\end{align}
The resulting \emph{Bayes regret} \citep{RVR14} of a policy $\pi$ relative to the oracle is:
\begin{align}
\text{Bayes Regret}_{N,T}(\pi)=\E_{\theta, \vx, \vep}\left[\sum_{i=1}^N \sum_{t=1}^T p^*_{i,t} D(p_{i,t}^*, \vx_{i,t})-\sum_{i=1}^N  \sum_{t=1}^Tp^{\pi}_{i,t} D(p_{i,t}^{\pi},\vx_{i,t})\right]  \,,
\end{align} 
where the expectation is taken with respect to the unknown product demand parameters, the observed random feature vectors, and the noise in the realized demand. The following theorem bounds the Bayes regret of the Thompson sampling dynamic pricing algorithm:

\begin{theorem}
	\label{theorem:oracle}
When the prior over the demand parameters is known, Algorithm \ref{alg:oracle} satisfies 
	\begin{align*}
	\textnormal{Bayes Regret}_{N,T}(\pi)=\tilde{O}\left(d^{\frac{3}{2}}N\sqrt{T}\right) \,,
	\end{align*}
\end{theorem}

Theorem \ref{theorem:oracle} follows from a similar argument used for the linear bandit setting presented in \cite{RVR14}, coupled with standard concentration bounds for multivariate normal distributions. The proof is given in Appendix \ref{sec:theorem:oracle} for completeness. Note that the regret scales linearly in $N$, since each epoch is an independent learning problem.

\begin{remark} Prior-independent Thompson sampling \citep{agrawal2013thompson} achieves a Bayes regret of $\tilde{O}(d^2N\sqrt{T})$, which is comparable to the performance of Algorithm \ref{alg:oracle}. However, we document a substantial gap in empirical performance between the two approaches in \S \ref{sec:numerics}, motivating our study of learning the prior.
\end{remark}

\subsection{Meta Oracle and Meta Regret}
\label{sec:oracle}
We cannot directly implement Algorithm \ref{alg:oracle} in our setting, since the prior over the product demand parameters $\N(\theta_*,\Sigma_*)$ is unknown. In this paper, we seek to learn the prior (shared structure) \textit{across} products in order to leverage the superior performance of Thompson sampling with a known prior. Thus, a natural question to ask is: 

\centerline{ \textit{What is the price of not knowing the prior in advance?} }

To answer this question, we first define our performance metric. Since our goal is to converge to the policy given in Algorithm \ref{alg:oracle} (which knows the true prior), we define this policy as our \emph{meta oracle}.\footnote{We use the term meta oracle to distinguish from the oracle in the classical formulation.}
Comparing the revenue of our policy relative to the meta oracle leads naturally to the definition of \emph{meta regret} $\regret_{N,T}$ for a policy $\pi$, \ie,
\begin{align*}
\regret_{N,T}(\pi) =\E_{\theta, \vx, \vep} \left[\sum_{i=1}^N \sum_{t=1}^T p^{\oracle}_{i,t}D(p^{\oracle}_{i,t}, \vx_{i,t}) - \sum_{i=1}^N \sum_{t=1}^T p_{i,t}^{\pi} D(p^{\pi}_{i,t}, \vx_{i,t})\right] \,,
\end{align*}
where the expectation is taken with respect to the unknown product demand parameters, the observed random feature vectors, and the noise in the realized demand.

Note that prior-independent Thompson sampling and UCB treat each epoch independently, and would thus achieve meta regret that grows linearly in $N$. Our goal is to design a policy with meta regret that grows sublinearly in $N$. Recall that Theorem \ref{theorem:oracle} bounds the Bayes regret of Thompson sampling with a known prior as $\tilde{O}\left(N\sqrt{T}\right)$. Thus, if our meta regret (\ie, the performance of our meta-learning policy relative to Algorithm \ref{alg:oracle}) grows sublinearly in $N$, then the price of not knowing the prior $\N(\theta_*,\Sigma_*)$ in advance is negligible in experiment-rich environments (large $N$) compared to the cost of learning the demand parameter for each product (the Bayes regret of Algorithm \ref{alg:oracle}).

The values of the prior mean $\theta_*$ as well as the actual product demand  parameter vectors $\{\theta_i \}_{i=1}^N$ are unknown; we consider two settings --- known and unknown $\Sigma_*$ (covariance of the prior).

\begin{remark}[Choice of meta oracle] \label{rem:oracle-choice}
To the best of our knowledge, the \textit{optimal} prior to use for Thompson sampling remains a difficult, open problem. Existing theory shows (in limited settings) that priors that fail to place sufficient mass on the true parameter fare poorly: the closest setting to ours is the linear bandit construction in Proposition 3.1 of \cite{hamidi2020general}, which shows that prior-dependent Thompson sampling with a mis-specified prior can achieve regret that scales exponentially in $d$; Theorem 1 of \cite{liu2015prior} and Theorem 2 of \cite{honda2014optimality} also provide illustrative constructions with the same insight. In the other extreme, many empirical evaluations suggest that overly conservative priors (such as prior-independent approaches) also fare poorly relative to using the true prior (see, \eg, Section 6 of \cite{bastani2020mostly}, the discussions in \cite{chapelle2011empirical}, or our numerical results in Section \ref{sec:numerics}). As a result, we choose Thompson Sampling with the true prior as our meta oracle. However, one can choose alternative meta oracles --- \eg, one that ``widens" the true prior to place more weight on parameters that may induce higher regret --- implementing such a meta oracle would still likely require learning the true prior, which is our primary contribution.
\end{remark}

\paragraph{Non-anticipating Policies:} We restrict ourselves to the family of non-anticipating policies $\Pi: \pi$ = $\{\pi_{i,t}\}$ that form a sequence of random functions $\pi_{i,t}$ that depend only on price and demand observations collected until time $t$ in epoch $i$ (including all times $t \in [T]$ from prior epochs), and feature vector observations up to time $t+1$ in epoch $i$. In particular, let $\cH_{0,0} = (\vx_{1,1})$, and $\cH_{i,t} = (p_{1,1},p_{1,2}, \cdots, p_{i,t}, D_{1,1}, D_{1,2}, \cdots, D_{i,t}, \vx_{1,1}, \vx_{1,2}, \cdots, \vx_{i, t+1})$ denote the history of prices and corresponding demand realizations from prior epochs and time periods, as well as the observed feature vectors up to the next time period; let $\mathcal{F}_{i,t}$ denote the $\sigma$-field generated by $\cH_{i,t}$. Then, we impose that $\pi_{i,t+1}$ is $\mathcal{F}_{i,t}$ measurable. 

\section{\texttt{Meta-DP} Algorithm} \label{sec:algorithm}

We begin with the case where the prior's covariance matrix $\Sigma_*$ is known, and describe the Meta Dynamic Pricing (\texttt{Meta-DP}) algorithm for this setting. We will consider the case of unknown $\Sigma_*$ in the next section.

\subsection{Overview}

The \mpdp~begins by using initial product epochs as an exploration phase to initialize our estimate of the prior mean $\theta_*$. These exploration epochs use the prior-independent Thompson sampling algorithm to ensure no more than $\tilde{O}(d^2 \sqrt{T})$ meta regret for each epoch. After this initial exploration period, our algorithm sequentially updates the estimated prior and leverages this estimate in each subsequent epoch. The key technical challenge is that the estimated prior has finite-sample estimation error, resulting in a Thompson sampling instance with a mis-specified prior. We introduce a prior alignment proof technique to show that, \textit{despite} prior mis-specification, our \mpdp~still achieves meta regret that grows sublinearly in $N$.

\subsection{Algorithm}

The \mpdp~is presented in Algorithm \ref{alg:main}. We first define some additional notation, and then describe the algorithm in detail. 

\paragraph{Additional Notation:} Throughout the rest of the paper, we use
$m_{i,t}=\begin{pmatrix}\vx^{\top}_{i,t}&~p_{i,t}\vx^{\top}_{i,t}
\end{pmatrix}^{\top}$
to denote the price and feature information and $V_{i,t}=\sum_{\tau=1}^tm_{i,t}m^{\top}_{i,t}$ to denote the Fisher information matrix of round $t$ in epoch $i$ for all $i\in[N]$ and $t\in[T].$

\paragraph{Algorithm Description:} The first $N_0$ epochs are treated as exploration epochs, where we define
\begin{align}
\label{eq:N_0}
N_0 ~=~ 4c^2_2d\T^2_e\log_e(4dN^2T)\log_e(2NT)~=~ \tilde{O}(d)\,,
\end{align} 
where $\T_e=\max\left\{6\log_{e/2}(dNT)/c_1,2\lambda_e/c_0\right\}=\tilde{O}(1)$ ($\T_e$ is a high probability upper bound on all $\T_i$'s, see Lemma \ref{lemma:ini} in Appendix \ref{sec:theorem:oracle}), and the constant is given by
\begin{align*} 
c_2=\frac{32\sqrt{x^2_{\max}(1+p^2_{\max})(\sigma^2\lambda^{-1}_e+5\overline{\lambda})}}{\lambda_e\underline{\lambda}\sigma^2} \,.
\end{align*}
As described in the overview, the \mpdp~proceeds in two phases. In particular, we distinguish the following two cases for each epoch $i$:
\begin{enumerate}
	\item \textbf{Epoch} $\mathbf{i \leq N_0}:$ the \mpdp~runs the prior-independent Thompson sampling algorithm \citep{agrawal2013thompson,AbeilleL17} TS$(\N(0,\Psi I_{2d}),\lambda_e),$ where 
	\[ \Psi=p_{\max}\sigma\sqrt{2d\log_e(T(1+x^2_{\max}p^2_{\max}(1+p^2_{\max})T))}+\sqrt{20\overline{\lambda}d\log_e(2T)} \,.\]
	This is simply Algorithm \ref{alg:oracle} with a conservative prior (variance is a function of the horizon $T$).
	\item \textbf{Epoch} $\mathbf{i > N_0}:$ the \mpdp~first computes the OLS estimate of the true parameter for each previous epoch $j <i$. It then average these parameter estimates to form an estimator $\hat{\theta}_i$ of the prior mean $\theta_*,$ \ie,
	\begin{align}
	\label{eq:ts_mean_update}
	\hat{\theta}_{i}=\frac{\sum_{j=1}^{i-1}V_{j,T}^{-1}\left(\sum_{t=1}^{T}D_{j,t}(p_{j,t},x_{j,t})m_{j,t}\right)}{i-1} \,.
	\end{align}
Then, the \mpdp~runs Thompson Sampling (Algorithm \ref{alg:oracle}) with the estimated prior $\N(\hat{\theta}_i,\Sigma_*)$, \ie, TS$(\N(\hat{\theta}_i,\Sigma_*),\lambda_e)$. Specifically, after some random initialization steps (these steps are identical to our meta oracle), our \mpdp~(1) samples the unknown product demand parameters $\mathring{\theta}_{i,t}=\left[\mathring{\alpha}_{i,t};\mathring{\beta}_{i,t}\right]$ from its posterior $\N\left(\theta^{\md}_{i,t},\Sigma^{\md}_{i,t}\right)$, and (2) solves and offers the resulting optimal price based on the demand function given by the sampled parameters
		\begin{align}
		p_{i,t} = \argmax_{p\in\left[p_{\min},p_{\max}\right]} ~ p \cdot \left\langle\mathring{\alpha}_{i,t}, \vx_{i,t}\right\rangle+ p^2 \cdot \left\langle\mathring{\beta}_{i,t},\vx_{i,t}\right\rangle \,.
		\end{align}
		Upon observing the actual realized demand $D_{i,t}\left(p_{i,t},\vx_{i,t}\right)$, the algorithm computes the posterior $\N\left(\theta^{\md}_{i,t+1},\Sigma^{\md}_{i,t+1}\right)$ for round $t+1$.

\end{enumerate}
\begin{algorithm}[h]
\SingleSpacedXI
	\caption{Meta-Dynamic Pricing Algorithm}
	\label{alg:main}
	\begin{algorithmic}[1]
		\State \textbf{Input:} The prior covariance matrix $\Sigma_*,$ the total number of epochs $N,$ the length of each epoch $T,$ the noise parameter $\sigma,$ and the set of feasible prices $[p_{\min},p_{\max}].$
		\State \textbf{Initialization:} $N_0\text{ as defined in Eq. (\ref{eq:N_0})}.$
		\For{each epoch $i=1,\ldots,N$}
		\If{$i\leq N_0$}
		\State{Run TS$\left(\N\left(0,\Psi\right),\lambda_e\right).$}
		\Else
		\State{Update $\hat{\theta}_{i}$ according to Eq. \eqref{eq:ts_mean_update}, and run TS$\left(\N\left(\hat{\theta}_i,\Sigma_*\right),\lambda_e\right).$}
		\EndIf
		\EndFor
	\end{algorithmic}
\end{algorithm}

We now state our main result upper bounding the meta regret of our \mpdp~(Algorithm \ref{alg:main}). The proof is provided in Section \ref{sec:meta_regret} and Appendix \ref{sec:theorem:main_regret}.

\begin{theorem}
	\label{theorem:main_regret}
	The meta regret of the proposed \mpdp~satisfies
	\begin{align*}
	\regret_{N,T}(\mpdp)=\begin{cases}
	\tilde{O}(d^2N\sqrt{T})&\text{ when }N<N_0\\
	\tilde{O}(d^2\sqrt{NT})&\text{ otherwise}
	\end{cases}~=\tilde{O}\left(d^2\sqrt{NT}+d^3\sqrt{T}\right).
	\end{align*}
\end{theorem}

It is worthwhile to compare the bound in Theorem \ref{theorem:main_regret} to the $\tilde{O}(d^2 N\sqrt{T})$ meta regret bound for prior-independent Thompson Sampling (Lemma \ref{lemma:ind-ts} in Appendix \ref{sec:theorem:main_regret}). When $N \lesssim \tilde{O}(d)$, our bound matches that of prior-independent Thompson Sampling, since we simply treat all our epochs as exploration epochs. In the large $N$ regime, our meta regret scales as  $\tilde{O}(d^2\sqrt{NT})$. Thus, our approach of learning the prior is particularly valuable in experiment-rich settings ($N \gg d$). Combining the two regimes yields a bound that is sublinear in both $N$ and $T$.

Theorem \ref{theorem:main_regret} is somewhat surprising in the context of a growing theoretical literature that suggests that a mis-specified prior can result in very poor regret for prior-dependent Thompson Sampling \citep[see, \eg,][]{honda2014optimality,liu2015prior, hamidi2020general}. Indeed, one may expect that the mis-specification induced by using the prior $\N(\hat{\theta}_i, \Sigma_*)$ instead of $\N(\theta_*, \Sigma_*)$ can be substantial, since the ratio between these two probability density functions is unbounded when $\hat{\theta}_i \neq \theta_*$. Yet, using our prior alignment proof strategy (described in the next subsection), we establish that Thompson Sampling is remarkably robust to mis-specification of the prior \textit{mean}, lending theoretical support to previous empirical observations \citep{bastani2020mostly}. 

\subsection{``Prior Alignment" Proof Strategy} \label{sec:meta_regret}

Since we only have a logarithmic number (in $N$ and $T$) of exploration epochs, the meta regret accrued from these epochs is $\tilde{O}(d^2N_0\sqrt{T})$ (see Lemma \ref{lemma:ind-ts} in Appendix \ref{sec:theorem:main_regret}).

In each non-exploration epoch $i > N_0$, the meta oracle starts with the true prior $\N(\theta_*, \Sigma_*)$ while our algorithm $\texttt{Meta-DP}$ starts with the estimated prior $\N(\hat{\theta}_i, \Sigma_*)$. The following lemma (whose proof is in Appendix \ref{sec:lemma:deviation}) bounds the error of the estimated prior mean with high probability:
\begin{lemma}
	\label{lemma:deviation}
	For any fixed $i\geq2$ and $\delta\in[0,2/e],$ with probability at least $1-\delta-2/(N^2T^2)$,
	\begin{align*}
	\left\|\hat{\theta}_i-\theta_*\right\|\leq8\sqrt{\frac{2(\sigma^2/\lambda_e+5\overline{\lambda})d\log_e(4d/\delta)}{i}} \,.
	\end{align*}
\end{lemma}	

Thus, the key challenge in proving Theorem \ref{theorem:main_regret} is bounding the difference in regret incurred by using a Thompson Sampling algorithm with a boundedly mis-specified prior.
We introduce a new ``prior alignment" proof technique to address this challenge. At a high level, we show that after the $\T_i$ exploration time steps, the distributions of the meta oracle's (random) posterior estimate $\theta^{\oracle}_{i,\T_i+1}$ and $\texttt{Meta-DP}$'s (random) posterior estimate $\theta^{\md}_{i,\T_i+1}$ are close. More specifically, there is a continuum of realizations of the stochastic noise (in the observed demands) such that $\texttt{Meta-DP}$ achieves the \textit{same} posterior estimate $\theta^{\md}_{i,\T_i+1} = \theta^{\oracle}_{i,\T_i+1}$ despite starting with a different prior; when such a match occurs, the expected regret moving forward from time $\T_i+1, \cdots, T$ is the same for both policies. Using this approach, the regret of our $\texttt{Meta-DP}$ algorithm can be expressed as a weighted distribution of the regret of the meta oracle (which we bounded in Theorem \ref{theorem:oracle}).

More specifically, the following lemma (whose proof is in Appendix \ref{sec:theorem:main_regret}) establishes the difference in Bayesian posteriors between the meta oracle and our \mpdp. Note that only the means of the posterior differ but the variance is the same.

\begin{lemma}\label{lemma:main1}
	Conditioned on $\theta_i$ and $x_{i,1},\ldots,x_{i,\T_i},$ the posteriors of the meta oracle and our algorithm \mpdp~satisfy
\begin{align*}
\theta^{\oracle}_{i,\T_i+1}-\theta^{\md}_{i,\T_i+1} &= \left(\Sigma_*^{-1}+\sigma\sum_{t=1}^{\T_i}m_{i,t}m_{i,t}^{\top}\right)^{-1}\left(\Sigma_*^{-1}\left(\theta_*-\hat{\theta}_i\right)+\sigma\sum_{t=1}^{\T_i}m_{i,t}\left(\vep_{i,t}^{\oracle}-\vep^{\md}_{i,t}\right)\right) \,, \\
\Sigma^{\oracle}_{i,\T_i+1} &= \Sigma^{\md}_{i,\T_i+1} \,.
\end{align*}
\end{lemma}

Now, consider any non-exploration epoch $i\geq N_0+1$. If upon completion of all exploration steps at time $\T_i+1$, we have that the posteriors of the meta oracle and our \mpdp~coincide --- \ie, $(\theta^{\md}_{i,\T_i+1},\Sigma^{\md}_{i,\T_i+1})=(\theta^{\oracle}_{i,\T_i+1},\Sigma^{\oracle}_{i,\T_i+1})$ --- then both policies would achieve the \textit{same} expected revenue over the time periods $\T_i+1, \cdots, T$. By Lemma \ref{lemma:main1}, we know that $\Sigma^{\oracle}_{i,\T_i+1} = \Sigma^{\md}_{i,\T_i+1}$ always, so all that remains is establishing when $\theta^{\oracle}_{i,\T_i+1} =  \theta^{\md}_{i,\T_i+1}$.

Since the two algorithms begin with different priors but encounter the same covariates $\{x_{i,t}\}_{t=1}^T$ and take the same decisions in $t \in \{1, \cdots, \T_i\}$, their posteriors can only align at time $\T_i+1$ due to the stochasticity in the observations $\vep_{i,t}$.
For convenience, denote the noise terms from $t\in \{1, \cdots, \T_i\}$ of the meta oracle and the \mpdp~respectively as
\begin{align}
\chi^{\oracle}_i &= \begin{pmatrix}\vep^{\oracle}_{i,1}&\ldots&\vep^{\oracle}_{i,\T_i}\end{pmatrix}^{\top} \,, \label{eq:chi-oracle}\\
\chi^{\md}_i &= \begin{pmatrix}\vep^{\md}_{i,1}&\ldots&\vep^{\md}_{i,\T_i}\end{pmatrix}^{\top} \,. \label{eq:chi-md}
\end{align}
Furthermore, let $M_i=\begin{pmatrix}
m_{i,1}&\ldots&m_{i,\T_i}
\end{pmatrix}\in\R^{2d\times \T_i}$. Lemma \ref{lemma:main1} indicates that if
\begin{align}\label{eq:main_key}
\chi^{\md}_{i}-\chi^{\oracle}_{i}=\frac{1}{\sigma}(M_i^{\top}M_i)^{-1}M_i^{\top}\Sigma_*^{-1}\left(\theta_*-\hat{\theta}_i\right),
\end{align}
then the posteriors of both algorithms align with $\theta^{\oracle}_{i,\T_i+1} =  \theta^{\md}_{i,\T_i+1}$. Thus for every realization of the meta oracle's noise terms $\chi^{\oracle}_{i}$ and the prior mean estimation error $\theta_*-\hat{\theta}_i$, there exists a well-defined and feasible choice of \mpdp's error $\chi^{\md}_{i}$ that allows the two posteriors to coincide. Furthermore, by Lemma \ref{lemma:deviation}, $\|\theta_*-\hat{\theta}_i\|$ is bounded as a function of $\sqrt{1/i}$ with high probability, ensuring that the difference in noise terms $\chi^{\md}_{i}-\chi^{\oracle}_{i}$ needed to achieve alignment is small for later epochs (as $i$ grows large). With this observation, we can perform a change of measure over our noise terms and integrate over the resulting distributions, yielding the desired bound on the meta regret. The proof is provided in Appendix \ref{sec:theorem:main_regret}.

\begin{remark}
Our prior alignment approach may be of general interest for analyzing the regret of mis-specified Thompson Sampling instances. \cite{RVR14} propose a related but different approach in Section 3.1 of their paper. Specifically, they relate the regret of implementing $TS(\N(\hat{\theta}_{i},\Sigma_*), \lambda_e)$ in an environment with true prior $\N(\theta_*,\Sigma_*)$ to the regret of $TS(\N(\hat{\theta}_{i},\Sigma_*), \lambda_e)$ in an environment with a \textit{different} true prior $\N(\hat{\theta}_{i},\Sigma_*)$. In contrast, we wish to compare the regret of implementing $TS(\N(\hat{\theta}_{i},\Sigma_*), \lambda_e)$ (\texttt{Meta-DP}, Algorithm \ref{alg:main}) and $TS(\N(\theta_*,\Sigma_*), \lambda_e)$ (meta oracle, Algorithm \ref{alg:oracle}) in the \textit{same} environment with true prior $\N(\theta_*,\Sigma_*)$. We cannot adopt their approach since one must additionally quantify the difference in regret between TS algorithms learning in environments with different true priors; while this regret difference clearly scales sublinearly in $T$, we require a bound that limits to $0$ as the difference in priors $\| \hat{\theta}_i - \theta_*\| \rightarrow 0$ (as $i \rightarrow \infty$). This requirement is because even a constant nonzero difference in regret between the meta oracle and our \mpdp~would result in $O(N)$ meta regret over $N$ epochs. To our knowledge, it is an open problem to derive such a bound. Our ``prior alignment" sidesteps this issue by directly relating $TS(\N(\hat{\theta}_{i},\Sigma_*), \lambda_e)$ and $TS(\N(\theta_*,\Sigma_*), \lambda_e)$ in an environment with true prior $\N(\theta_*,\Sigma_*)$.
\end{remark}

\section{\texttt{Meta-DP++} Algorithm}\label{sec:extension}
In this section, we consider the setting where the prior covariance matrix $\Sigma_*$ is also unknown. We propose the \covmpdp, which builds on top of the \mpdp~and additionally estimates the unknown prior covariance $\Sigma_*.$

\subsection{Overview}

The \covmpdp~also begins by using initial product epochs as an exploration phase to initialize our estimate of the prior mean $\theta_*$ and covariance $\Sigma_*$. After this initial exploration period, our algorithm sequentially updates the estimated prior and leverages this estimate in each subsequent epoch. Once again, the estimated prior has finite-sample estimation error, resulting in a Thompson sampling instance with a mis-specified prior. The key challenge compared to the previous section is that we can no longer exactly ``align" our algorithm's posterior with that of the meta oracle when $\Sigma_*$ is also estimated. We leverage importance sampling arguments from off-policy evaluation to bound the additional meta regret accrued due to this mismatch. Importantly, to ensure that our importance weights remain well-behaved, we \textit{widen} the estimated covariance via a correction term that scales as the finite-sample estimation error of estimating $\hat{\Sigma}_*$.

\subsection{Algorithm}

The \covmpdp~is presented in Algorithm \ref{alg:extension}. We first define some additional notation, and then describe the algorithm in detail.

\paragraph{Additional Notation:} As with the \mpdp, at the beginning of each epoch $i \in [N]$, we update our estimate $\hat{\theta}_i$ of the prior mean $\theta_*$ according to Eq. \eqref{eq:ts_mean_update}. To estimate $\Sigma_*$,  we need unbiased and \textit{independent} estimates for the unknown true demand parameter realizations $\theta_i$ across epochs.\footnote{When estimating the prior covariance, we cannot use an estimator of $\theta_i$ that uses all $T$ observations from epoch $i$ (as we do when estimating the prior mean). This is because the use of the learned prior from past epochs renders observations from later epochs non-independent. We avoid this issue by restricting our estimator of $\theta_i$ to observations from the initialization periods in each epoch, $t \in [\T_i]$.} We use the initialization steps $t\in[\T_i]$ to produce an estimate $\dot{\theta}_{i}$ for $\theta_i,$ \ie, 
\begin{align}
\nonumber\dot{\theta}_i=V_{i,\T_i}^{-1}\left(\sum_{t=1}^{\T_i}D_{i,t}(p_{i,t},x_{i,t})m_{i,t}\right).
\end{align}  

\paragraph{Algorithm Description:} The first $N_1$ epochs are treated as exploration epochs, where we employ the prior-independent Thompson Sampling algorithm. We define
\begin{align}
 N_1&~=~\max\left\{N_0,~256c^2_3d^3\T^2_e\log^3_e(4dN^2T),~c^2_4d^{4}T^2\log^{3}_e(2N^2T)\right\} ~=~ \tilde{O}(d^4T^2) \label{eq:N_1} \,,
\end{align}
and the constants are given by
\begin{align*}
c_3=& \frac{16\sqrt{\sigma^2\lambda^{-1}_e+5\overline{\lambda}}}{\sigma\lambda_e\underline{\lambda}}+\frac{256(\overline{\lambda}\lambda^2_e+16\sigma^2)}{\lambda_e^2\underline{\lambda}^2}\left(\frac{8 p_{\max}x_{\max}\sqrt{(1+p^2_{\max})}}{\lambda_e}+\frac{S}{\sigma\lambda_e}\right)\,, \quad c_4=\frac{10^4\sigma(\overline{\lambda}\lambda^2_e+16\sigma^2)}{\lambda_e^2\underline{\lambda}^2} \,.
\end{align*}
Note that we now require $\tilde{O}(\min\{N,d^4T^2\})$ exploration epochs, whereas we only required $\tilde{O}\left(d^2\right)$ exploration epochs for the \mpdp. 

As described in the overview, the \covmpdp~proceeds in two phases: 
\begin{enumerate}
\item \textbf{Epoch $\mathbf{i \leq N_1}$:} the \covmpdp~runs the prior-independent Thompson sampling algorithm \citep{agrawal2013thompson,AbeilleL17} TS$(\N(0,\Psi I_{2d}),\lambda_e),$ where 
\[\Psi=p_{\max}\sigma\sqrt{2d\log_e(T(1+x^2_{\max}p^2_{\max}(1+p^2_{\max})T))}+\sqrt{20\overline{\lambda}d\log_e(2T)} \,.\]
This is simply Algorithm \ref{alg:oracle} with a conservative prior (variance is a function of the horizon $T$).

\item \textbf{Epoch $\mathbf{i > N_1}$:} the \covmpdp~computes an estimator $\hat{\theta}_i$ of the prior mean $\theta_*$ using Eq. \eqref{eq:ts_mean_update} (same as \mpdp), and an estimator $\hat{\Sigma}_i$ of the prior covariance $\Sigma_*$ as
\begin{align}\label{eq:ts_cov_update}
\hat{\Sigma}_i=\frac{1}{i-2}\sum_{j=1}^{i-1}\left(\dot\theta_{j}-\frac{\sum_{k=1}^{i-1}\dot\theta_k}{i-1}\right)\left(\dot\theta_{j}-\frac{\sum_{k=1}^{i-1}\dot\theta_k}{i-1}\right)^{\top}-\frac{\sigma^2\sum_{j=1}^{i-1}\E\left[V^{-1}_{j,\T_j}\right]}{i-1} \,.
\end{align}
The second term $\sigma^2\sum_{j=1}^{i-1}\E\left[V^{-1}_{j,\T_j}\right]/(i-1)$ accounts for the estimation error in $\{\dot{\theta}_j\}_{j=1}^{i-1}$.

As noted earlier, we then \textit{widen} our estimator to account for finite-sample estimation error:
\begin{align}\label{eq:widening}
\hat{\Sigma}^{w}_i ~=~ \hat{\Sigma}_i+\frac{128(\overline{\lambda}\lambda^2_e+16\sigma^2d)}{\lambda_e^2}\sqrt{\frac{5d\log_e(2N^2T)}{i}}\cdot I_{2d} \,,
\end{align}
where $I_{2d}$ is the $(2d)$-dimensional identity matrix. 

Then, the \covmpdp~runs Thompson Sampling (Algorithm \ref{alg:oracle}) with the estimated prior $\N(\hat{\theta}_i, \hat{\Sigma}^{w}_i)$, \ie, TS$(\N(\hat{\theta}_i, \hat{\Sigma}^{w}_i), \lambda_e)$. Specifically, after some random initialization steps (these steps are identical to our meta oracle), our \covmpdp~(1) samples the unknown product demand parameters $\mathring{\theta}_{i,t}=\left[\mathring{\alpha}_{i,t};\mathring{\beta}_{i,t}\right]$ from the posterior $\N\left(\theta^{\mts}_{i,t},\Sigma^{\mts}_{i,t}\right)$, and (2) solves and offers the resulting optimal price based on the demand function given by the sampled parameters
\begin{align}
p_{i,t} = \argmax_{p\in\left[p_{\min},p_{\max}\right]} ~ p \cdot \left\langle\mathring{\alpha}_{i,t}, \vx_{i,t}\right\rangle+ p^2 \cdot \left\langle\mathring{\beta}_{i,t},\vx_{i,t}\right\rangle \,.
\end{align}
Upon observing the actual realized demand $D_{i,t}\left(p_{i,t},\vx_{i,t}\right)$, the algorithm computes the posterior $\N\left(\theta^{\mts}_{i,t+1},\Sigma^{\mts}_{i,t+1}\right)$ for round $t+1$.

\end{enumerate}

\begin{algorithm}[!ht]
\SingleSpacedXI
	\caption{Meta-Dynamic Pricing++ Algorithm}
	\label{alg:extension}
	\begin{algorithmic}[1]
		\State \textbf{Input:} The total number of products $N,$ the length of each epoch $T,$ the noise parameter $\sigma,$ and the set of feasible prices $[p_{\min},p_{\max}].$
		\For{epoch $i=1,\ldots,N$}
		\If{$i\leq N_1$}
		\State{Run TS$\left(\N\left(0,\Psi\right),\lambda_e\right).$}
		\Else
		\State{Update $\hat{\theta}_{i}$ and $\hat{\Sigma}_i$ according to Eqs. \eqref{eq:ts_mean_update} and \eqref{eq:ts_cov_update} respectively.}
		\State{Compute widened prior mean estimate $\hat{\Sigma}^{w}_i$ according to Eq. \eqref{eq:widening}.}
		\State{Run TS$\left(\N\left(\hat{\theta}_i,\hat{\Sigma}^{w}_i\right),\lambda_e\right).$}
		\EndIf
		\EndFor
	\end{algorithmic}
\end{algorithm}

We now state our main result upper bounding the meta regret of our \covmpdp~(Algorithm \ref{alg:extension}). The proof is provided in Section \ref{sec:proof++} and Appendix \ref{sec:theorem:cov_regret}.

\begin{theorem}
	\label{theorem:cov_regret}
	The meta regret of the proposed \covmpdp~satisfies
	\begin{align*}
	\regret_{N,T}(\covmpdp) &= \tilde{O}\left(\min\left\{d^2 N T^{\frac{1}{2}},~ d^4 N^{\frac{1}{2}}T^{\frac{3}{2}} \right\}\right) = \tilde{O}\left(d^3(NT)^{\frac{5}{6}}\right) \,.
	\end{align*}
\end{theorem}

It is worthwhile to compare the bound in Theorem \ref{theorem:cov_regret} to the $\tilde{O}(d^2 N\sqrt{T})$ meta regret bound for prior-independent Thompson Sampling (Lemma \ref{lemma:ind-ts} in Appendix \ref{sec:theorem:main_regret}). When $N \lesssim \tilde{O}(d^4 T^2)$, our bound matches that of prior-independent Thompson Sampling, since we simply treat all our epochs as exploration epochs. In the large $N$ regime, our meta regret scales as  $\tilde{O}(d^4 N^{\frac{1}{2}}T^{\frac{3}{2}})$. Thus, our approach of learning the prior is particularly valuable in settings with many short-horizon experiments ($N \gg T$). For instance, as discussed in Example \ref{ex:ruelala}, sellers like Rue La La host many events, offering new items with short selling seasons. Combining the two regimes yields a bound that is sublinear in both $N$ and $T$.

\subsection{Proof Strategy} \label{sec:proof++}

The number of exploration epochs $N_1$ is logarithmic number in $N$ but quadratic in $T$. This motivates the analysis of two cases: (i) when the number of epochs $N < N_1 = \tilde{O}(d^4 T^2)$, the meta regret guarantees given by existing prior-independent approaches is already good; (ii) when we transition to an experiment rich environment with $N > N_1$, the meta regret accrued from these epochs is small since their cardinality scales logarithmically in $N$ (see argument in Appendix \ref{sec:theorem:cov_regret}). We now focus on the latter case where $N$ is large.

Once again, following the proof strategy employed for \mpdp, we employ ``prior alignment" to match the means of the meta oracle's (random) posterior estimate and \texttt{Meta-DP++}'s (random) posterior estimates. However, since $\Sigma_*$ was known in the previous section, matching the posterior means $\theta^{\md}_{i,\T_i+1} = \theta^{\oracle}_{i,\T_i+1}$ implied equality of the \textit{entire distribution} of the posterior (see Lemma \ref{lemma:main1}). This equivalence allowed us to exactly equate the expected regret (after alignment) for the meta oracle and our \mpdp.

However, when $\Sigma_*$ is unknown, matching the posterior means $\theta^{\mts}_{i,\T_i+1} = \theta^{\oracle}_{i,\T_i+1}$ no longer implies that the posterior distributions are equal. Furthermore, since the Bayesian update for the covariance matrix does not depend on the noise terms (it depends only on the observed covariates and chosen prices), we cannot use any alignment strategy based on $\chi^{\oracle}_i$ and $\chi^{\mts}_i$ to get exact equivalence of the posterior distributions.
Thus, the key added challenge in proving Theorem \ref{theorem:cov_regret} is bounding the difference in regret between our \covmpdp~and the meta oracle \textit{after} alignment of the means of their posteriors at time $t=\T_i$.

Specifically, in each non-exploration epoch $i > N_1$, the meta oracle starts with the true prior $\N(\theta_*, \Sigma_*)$ while our algorithm \texttt{Meta-DP++} starts with the (widened) estimated prior $\N(\hat{\theta}_i, \hat{\Sigma}^{w}_i)$. Lemma \ref{lemma:deviation} from the previous section already provides a bound on $\|\hat{\theta}_i - \theta_*\|$, and the following lemma (whose proof is in Appendix \ref{sec:lemma:deviation_cov}) bounds the error of the estimated covariance $\|\hat{\Sigma}_i - \Sigma_*\|$ (and thus the error of our widened covariance $\|\hat{\Sigma}^w_i - \Sigma_*\|$) with high probability:
\begin{lemma}
	\label{lemma:deviation_cov}
	For any fixed $i\geq3$ and $\delta\in[0,2/e]$, with probability at least $1-2\delta-2/(N^2T^2)$, 
	\begin{align*}
	\left\|\hat{\Sigma}_i-\Sigma_*\right\|_{op}\leq \frac{128(\overline{\lambda}\lambda^2_e+16\sigma^2d)}{\lambda_e^2}\left(\sqrt{\frac{5d\log_e(2/\delta)}{i}}\vee\frac{5d\log_e(2/\delta)}{i}\right) \,.
	\end{align*} 
\end{lemma}

At time $t=\T_i + 1$, we use a change of measure to ``align" our \covmpdp's prior $\N(\theta^{\mts}_{i,\T_i+1},\Sigma^{\mts}_{i,\T_i+1})$ to $\N(\theta^{\oracle}_{i,\T_i+1},\Sigma^{\mts}_{i,\T_i+1})$. Combining Lemma \ref{lemma:deviation_cov} and the fact that both policies offer the same prices in the random exploration periods, we know that $\Sigma^{\oracle}_{i,\T_i+1}$ and $\Sigma^{\mts}_{i,\T_i+1}$ are close with high probability for later epochs. However, it remains to bound the regret difference between the meta oracle's policy, which employs the prior $\N(\theta^{\oracle}_{i,\T_i+1},\Sigma^{\oracle}_{i,\T_i+1})$, and our \covmpdp, which employs the prior $\N(\theta^{\oracle}_{i,\T_i+1},\Sigma^{\mts}_{i,\T_i+1})$. We leverage importance sampling arguments from off-policy evaluation \citep{PrecupSS00, MurphyVRC01} to bound this remaining term. Prior widening is instrumental in this last step, ensuring that our importance weights do not diverge.

\begin{remark}
While our \mpdp~does not require prior widening, we widen our prior for our \covmpdp~as described above. This allows us to shave off some extra factors of the dimension $d$ in our analysis, by ensuring that the importance weights are well-behaved post-alignment. This is consistent with recent work by \cite{hamidi2020general}, who show that Thompson sampling can in general incur a worst-case regret that scales exponentially in $d$, unless it uses a widened posterior variance at each step. Furthermore, we observe (often significantly) improved empirical performance on both synthetic and real datasets by employing our \covmpdp~compared to its non-widened analog (see Section \ref{sec:numerics}).
\end{remark}

\subsection{Additional Remarks} \label{ssec:remarks}

\paragraph{Hierarchical Model:} An alternative heuristic to leverage shared structure is to use hierarchical Thompson Sampling, maintaining a posterior on the shared prior and updating it after each epoch. In Appendix \ref{app:hierarchical}, we compare the \mpdp~to a hierarchical approach; while the hierarchical algorithm outperforms prior-independent Thompson Sampling by leveraging shared structure, we find that it still significantly underperforms compared to the \mpdp~for moderate to large values of $N$ due to excessive exploration.

\paragraph{Knowledge of $N,T$:} Our formulation assumes knowledge of $N$ and $T$. However, this assumption can easily be removed using the well-known ``doubling trick". In particular, we can initially fix any values $N_0$ and $T_0$, and iteratively double the length of the respective horizons; we refer the interested reader to \cite{CBL06} for details. For the \mpdp, we would simply continue to update the estimated prior mean; for the \covmpdp, we would need to also follow the prior widening schedule. It is easy to see that our regret bounds are preserved up to logarithmic terms under such an approach.

\paragraph{Overlapping Epochs:} We model epochs as fully sequential for simplicity; if epochs overlap, we would need to additionally model a customer arrival process for each epoch. Our algorithms straightforwardly generalize to a setting where arrivals are randomly distributed across overlapping epochs. In particular, both the \mpdp~and the \covmpdp~can be modified to only use samples from the \textit{initialization period} $t\in[\T_i]$ in each epoch for estimating the prior mean (note that our estimation of the prior covariance already only uses samples from initialization periods) without affecting the meta regret bounds and analysis. Therefore, when epochs overlap, we will update our estimate of the prior as soon as we see $\tilde{O}(1)$ customer responses for any product.

\section{Numerical Experiments} \label{sec:numerics}

We now validate our theoretical results by empirically comparing the performance of our proposed algorithms against prior-independent Thompson Sampling \citep{agrawal2013thompson}. As discussed earlier, this approach ignores learning shared structure (the prior) across products, and achieves $\tilde{O}(d^2 N\sqrt{T})$ meta regret (see Lemma \ref{lemma:ind-ts} in Appendix \ref{sec:theorem:main_regret}). When the prior covariance is unknown, we illustrate the benefits of prior widening by additionally comparing against a version of the \covmpdp~that greedily uses the estimated covariance matrix  (\ie, $\Sigma_i=\hat{\Sigma}_i$).

In addition to meta regret, we present results on Bayes regret (relative to the classical oracle) to illustrate that our transfer learning approach significantly increases performance under the standard metric. We perform numerical experiments on both synthetic data as well as a real dataset on auto loans provided by the Columbia University Center for Pricing and Revenue Management.

A number of additional numerical results are presented in Appendix \ref{app:add_numerics}, including comparison to a hierarchical Thompson Sampling heuristic (\ref{app:hierarchical}), examining the estimation error of the prior as a function of $N$ (\ref{sec:add_numerical_err}), as well as results under a revenue metric (\ref{sec:add_numerical}).

\subsection{Synthetic Data}

We begin with the case where the prior covariance $\Sigma_*$ is known. 

\paragraph{Parameters:} We consider $N=700$ products, each with a selling horizon of $T=300$ periods. We set the feature dimension $d=5,$ the prior mean $\theta_*=[1.2\times\mathbf{1}_d;-0.3\times\mathbf{1}_d]^{\top},$ and the prior covariance $\Sigma_*=0.2\times I_{2d}.$ In each epoch $i \in [N]$ and each round $t \in [T]$, each entry of the observed feature vector $\vx_{i,t}$ is drawn i.i.d. from the uniform distribution over $[0,1/\sqrt{d}]^d$; note that this ensures the $\ell_2$ norm of each feature vector is upper bounded by $1.$ For each product $i \in [N]$, we randomly draw a demand parameter $\theta_i$ i.i.d. from the true prior $\N\left(\theta_*,\Sigma_*\right).$ The allowable prices lie in $(0,5]$. Finally, the noise distribution is the standard normal distribution, \ie, $\sigma=1.$

\begin{figure}[h]
	\centering
	\includegraphics[width=17cm,height=6.5cm]{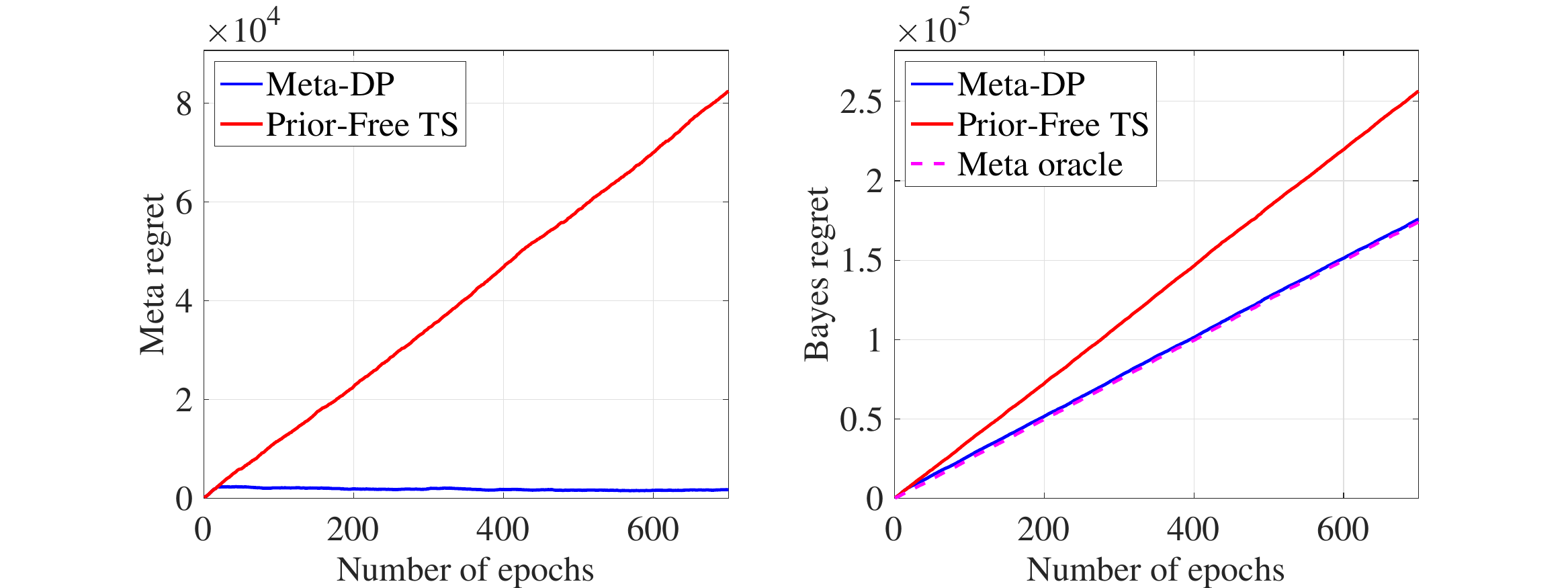}
	\caption{Cumulative meta regret and Bayes regret for \texttt{Meta-DP} and prior-independent Thompson Sampling.}
	\label{fig:d5}
\end{figure}

\paragraph{Results:} We plot the cumulative meta regret and Bayes regret of each algorithm, averaged over 20 random trials, as a function of the number of epochs $N$ (recall that each epoch lasts for $T$ periods). The results are shown in Figure \ref{fig:d5}. Both algorithms are identical during the initial exploration epochs.

As expected, the prior-independent approach achieves meta regret that scales linearly in $N$, since each epoch is treated independently. In contrast, the left panel of Figure \ref{fig:d5} shows that \texttt{Meta-DP} achieves nearly zero meta regret after the exploration epochs as it has learned the prior.

The right panel of Figure \ref{fig:d5} examines Bayes regret; note that even the meta-oracle achieves $O(N)$ Bayes regret (Theorem \ref{theorem:oracle}). However, the \textit{slope} of \texttt{Meta-DP} closely matches that of the meta-oracle after the initial exploration epochs, \ie, we do not accrue additional regret (relative to the meta oracle) as $N$ grows large. In contrast, the slope of prior-independent Thompson Sampling is significantly larger, resulting in additional regret continually accruing as $N$ grows large. In particular, when $N=700,$ the Bayes regret of prior independent Thompson Sampling is 39\% larger than that of \texttt{Meta-DP} and 48\% larger than that of the meta oracle. Thus, our approach of learning shared structure is particularly valuable in experiment-rich environments.

\begin{figure}[h]
	\subfigure[$d=1$]{\label{fig:d1}\includegraphics[width=17cm,height=6.5cm]{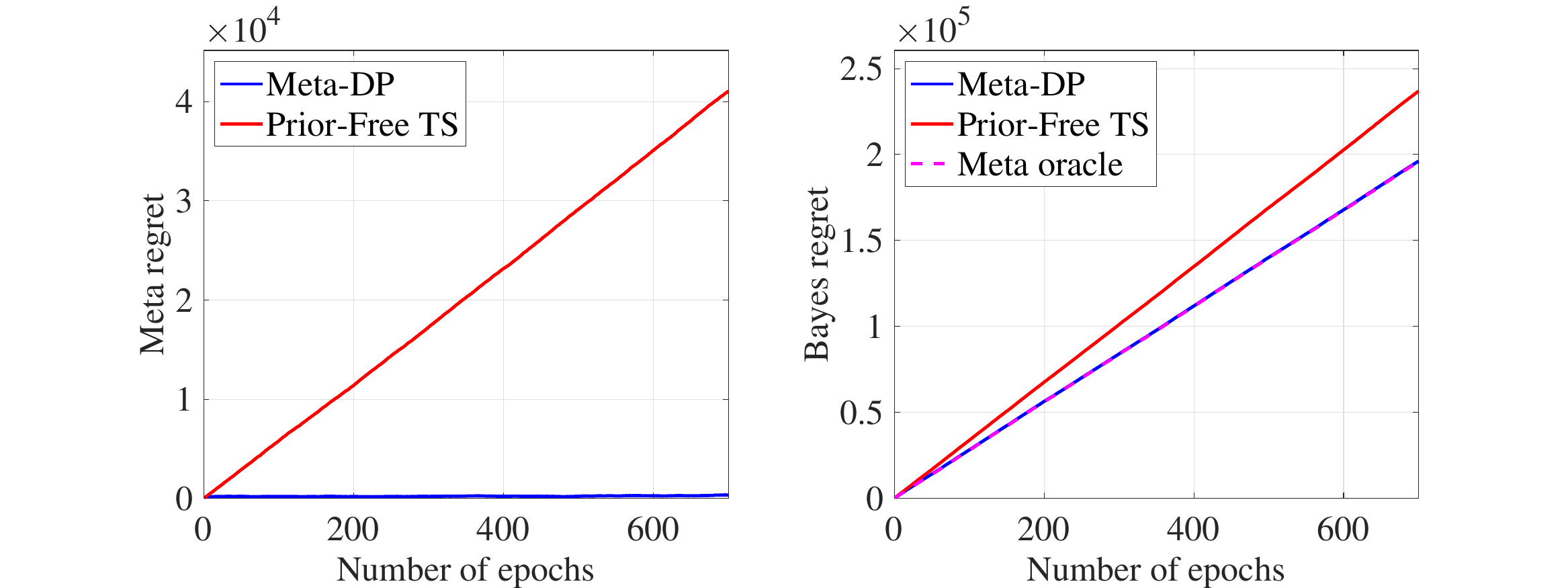}}
	\subfigure[$d=10$]{\label{fig:d10}\includegraphics[width=17cm,height=6.5cm]{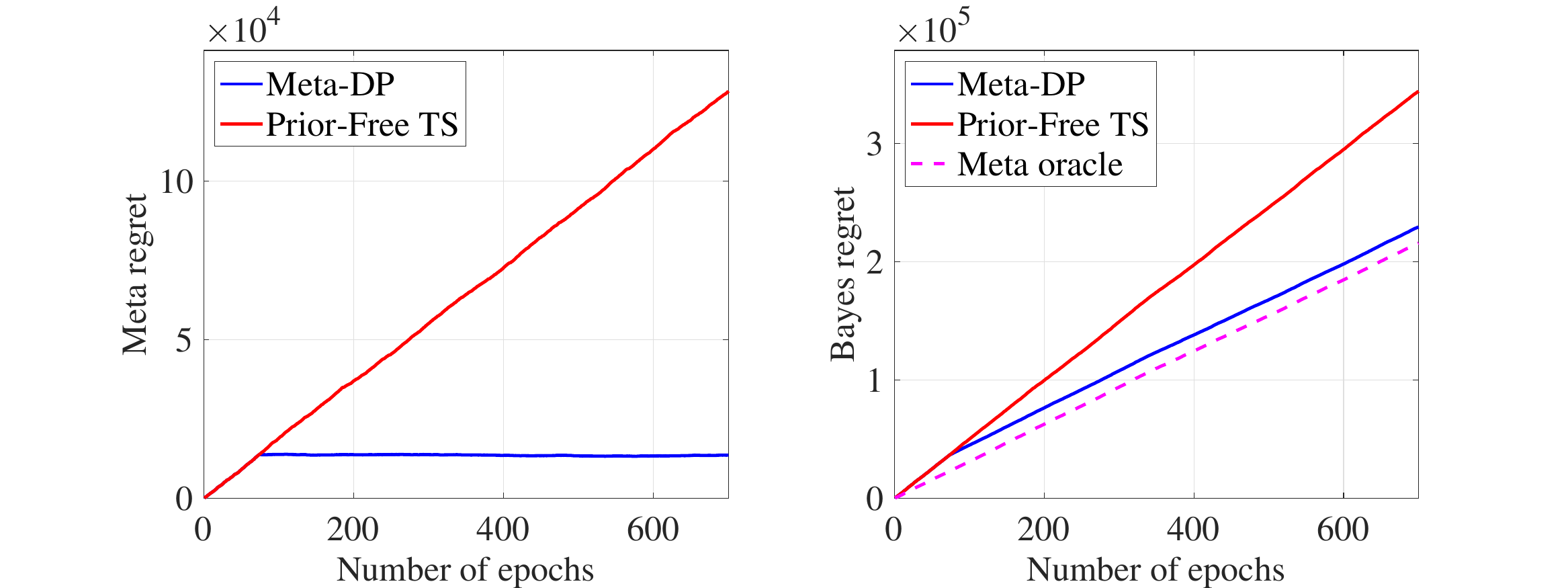}}
	\caption{Cumulative meta regret and Bayes regret for \texttt{Meta-DP} and prior-independent Thompson Sampling for different values of the feature dimension $d$.}
	\label{fig:d_var}
\end{figure}

\paragraph{Varying the feature dimension $d$:} We now explore how our results vary as we change the dimension of the observed features. Our previous results considered $d=5$. We now additionally consider:
\begin{enumerate}
\item \textit{No features, $d=1$:} We set $\vx_{i,t}=1$ for all $i\in[N]$ and $t\in[T].$
\item \textit{Many features, $d=10$:} Each entry of the observed feature vector $\vx_{i,t}$ is again drawn i.i.d. from the uniform distribution over $[0,1/\sqrt{d}]^d$ for all $i\in[N]$ and $t\in[T].$
\end{enumerate}
The results for both cases, averaged over 20 random trials, are shown in Figures \ref{fig:d1} and \ref{fig:d10} respectively. Again, we see that \texttt{Meta-DP} substantially outperforms prior-independent Thompson sampling algorithm in both meta regret and Bayes regret, regardless of the choice of feature dimension $d$. Note that we require more exploration epochs when $d$ is larger (recall that $N_0$ scales as $d$).

\begin{figure}[h]
	\centering
	\includegraphics[width=17cm,height=6.5cm]{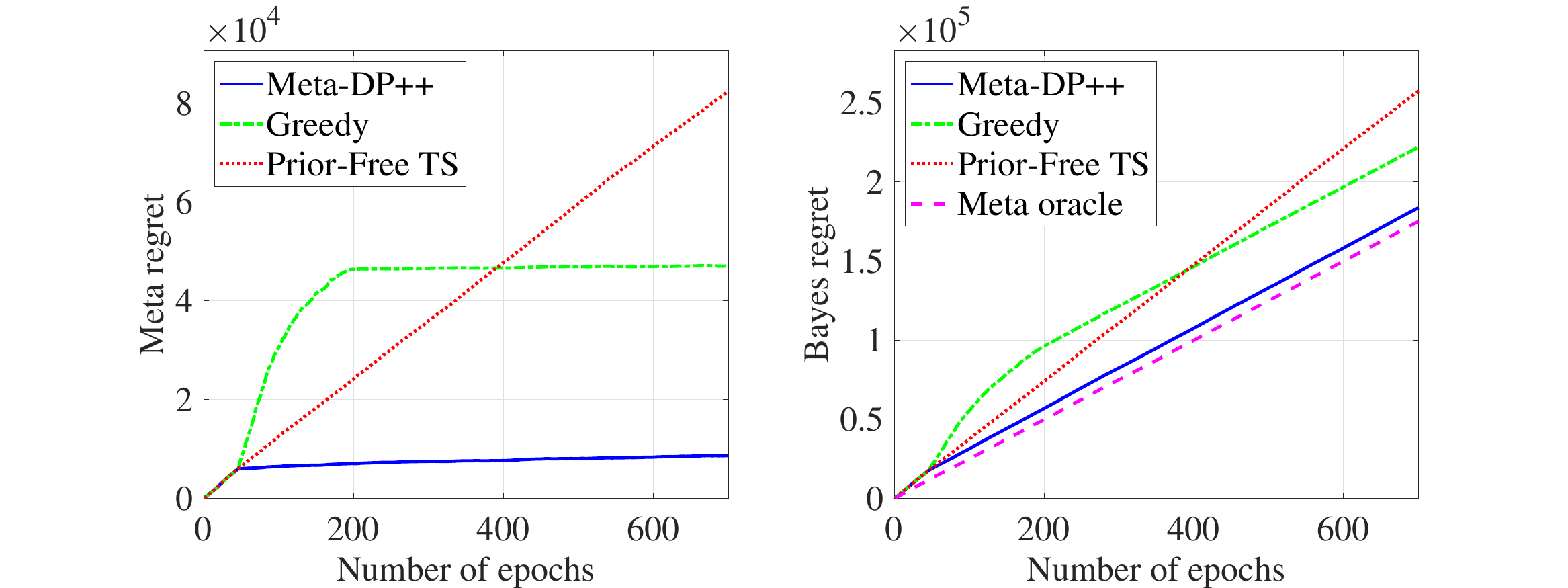}
	\caption{Cumulative meta regret and Bayes regret for \texttt{Meta-DP++} and benchmark algorithms.}
	\label{fig:pp}
\end{figure}

\paragraph{Unknown prior covariance $\Sigma_*$:} We now shift our attention to the \covmpdp, and follow the same setup described earlier. To quantify the benefit of prior widening, we additionally consider a version of the \covmpdp~that \textit{greedily} uses the estimated covariance matrix, \ie, $\Sigma_i=\hat{\Sigma}_i.$ The results, averaged over 20 random trials, are shown in Figure \ref{fig:pp}. We see that the \covmpdp~significantly outperforms both the prior-independent Thompson sampling algorithm as well as the non-widened greedy benchmark in meta regret (left panel) and Bayes regret (right panel). Interestingly, the greedy approach performs significantly worse in earlier epochs after the initial exploration epochs (when it relies on a prior that is likely to be significantly mis-specified); in later epochs, the greedy approach's slope begins to match that of \texttt{Meta-DP++} as it starts learning the true prior. Thus, prior widening appears critical to ensure good performance on \textit{each} pricing problem --- particularly earlier ones, where we should be careful not to over-rely on a prior is likely to be significantly mis-specified. The overall success of \texttt{Meta-DP++} suggests that the price of not knowing the prior in advance is negligible in experiment-rich environments (large $N$).

\subsection{Real Data on Online Auto-Lending}
\label{sec:auto_lending}
We now turn to the on-line auto lending dataset. This dataset was first studied by \cite{PSR15}, and subsequently used to evaluate dynamic pricing algorithms by \cite{ban2017personalized}. We will follow a similar set of modeling assumptions.

The dataset records all auto loan applications received by a major online lender in the United States from July 2002 through November 2004. It contains $208,085$ loan applications. For each application, we observe some loan-specific features (\eg, date of application, the term and amount of loan requested, and the borrower's personal
information), the lender's pricing decision (\ie, the monthly payment required of the borrower), and the resulting demand (\ie, whether or not this offer was accepted by the borrower). We refer the interested reader to Columbia University Center for Pricing and Revenue Management \citep{CRPM} for a detailed description of the dataset.

\paragraph{Algorithms:} We consider the setting where both the prior mean and prior covariance are unknown. Thus, we compare the performance of \covmpdp~against that of prior-independent Thompson Sampling, the ILSX algorithm proposed in \cite{ban2017personalized}, and the greedy version of \texttt{Meta-DP++} that does not employ prior widening.

\paragraph{Products:} We first define a set of related products. We segment loans by the borrower's state (there are 50 states), the term class of the loan (0-36, 37-48, 49-60, or over 60 months), and the car type (new, used, or refinanced). The expected demand and loan decisions offered for each type of loan is likely different based on these attributes. We consider loans that share all three attributes as a single ``product" offered by the online lender. We thus obtain a total of $N=589$ unique products. The number of applicants in the data for each loan type determines $T$ for each product; importantly, note that $T$ is not identical across products.

\begin{remark}
We use three categorical features (state, term of loan, and car type) to define $N=589$ products. In contrast, the ILSX algorithm \citep{ban2017personalized} sets $N=1$ and encodes this information as product features; this results in a feature vector of dimension $\Theta(d+N)$, since each possible value of the categorical feature will be represented as 1-hot encoding. The resulting meta regret of ILSX will therefore still grow superlinearly in $N$ (unlike our proposed algorithms). Moreover, their demand model is less expressive compared to ours since it does not allow for different price elasticities by state/term/car type (see our earlier Remark \ref{rem:prod-features} for discussion).
\end{remark}

\begin{remark}
Following our model, we simulate each epoch sequentially. In reality, customers will likely arrive randomly for each loan type at different points of time. We note that the \mpdp~only uses the initial sample from each epoch for estimating the prior mean, and thus, in principle, it can be adapted to a setting where arrivals are randomly distributed across overlapping epochs as well (see discussion in \S \ref{ssec:remarks}).
\end{remark}

\paragraph{Features:} We use the feature selection results from \cite{ban2017personalized}, which yields the following features: FICO score, the loan amount approved, prime rate, and the competitor's rate.
\begin{figure}[!ht]
	\centering
	\subfigure[Cumulative meta regret and Bayes regret for \texttt{Meta-DP++} and benchmark algorithms]{\includegraphics[width=17cm,height=6.5cm]{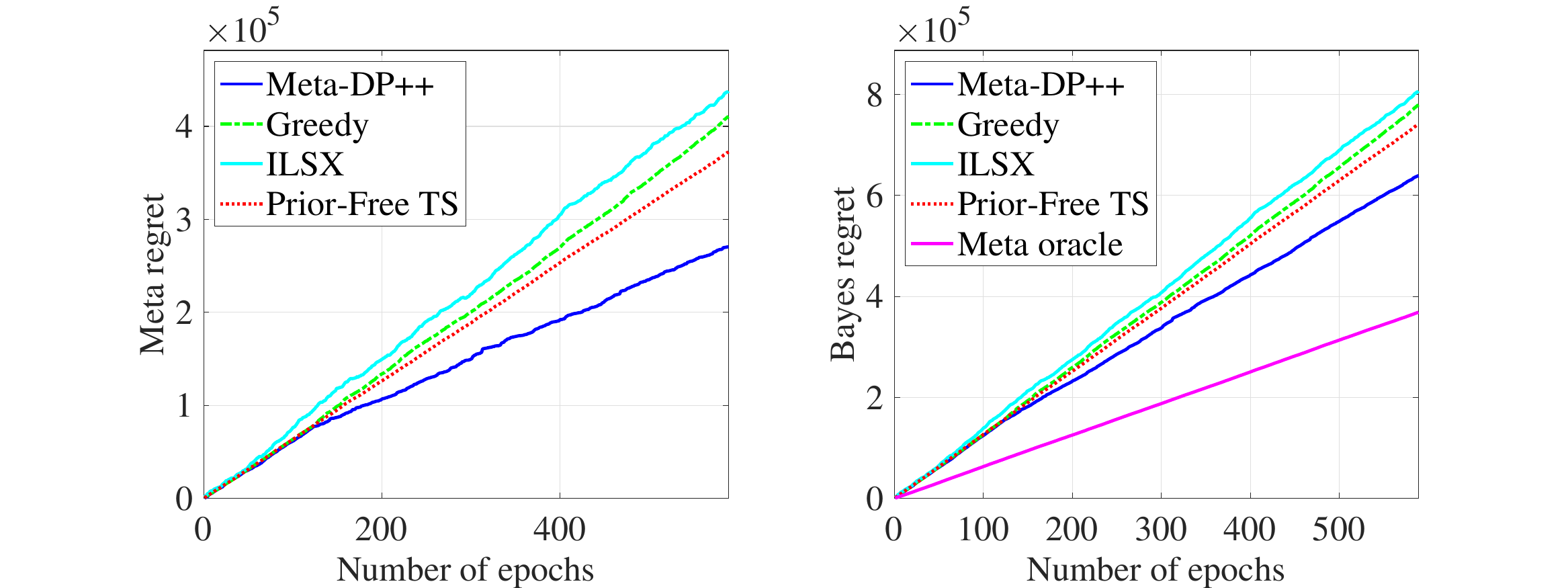}}
	\caption{Computational results on a real dataset on online auto loans.}
	\label{fig:auto}
\end{figure}

\paragraph{Setup:} Following the approach of \cite{PSR15} and \cite{ban2017personalized}, we impute the price of a loan as the net present value of future payments (a function of the monthly payment, customer rate, and term approved; we refer the reader to the cited references for details). The allowable price range in our experiment is $[0,30]$.

We note that, although we use a linear demand model, our responses are binary (\ie, whether a customer accepts the loan). This approach is common in the literature \citep[see, e.g.,][]{LCLS10}. \cite{BZ15} provide theoretical justification for this approach by showing that we may still converge to the optimal price despite the demand model being misspecified.

Finally, unlike our model and analysis, the true distribution over loan demand parameters across products may not be a multivariate Gaussian. We use the entire dataset to estimate each product's demand parameter, and then fit a multivariate Gaussian prior over the empirical distribution of product demand parameters --- our meta oracle uses this prior. However, our regret is evaluated with respect to the true data (\ie, our meta oracle may perform poorly in Bayes regret if the prior is far from a multivariate Gaussian). Thus, this experiment can provide a check on whether our algorithms (which seek to mimic the meta oracle) are robust to model misspecification of the prior.

\paragraph{Results:} We average our results over 100 random permutations of the data. The results are shown in Figure \ref{fig:auto}. We first note that, despite potential misspecification of the prior's model class, the meta oracle (prior-dependent Thompson Sampling) achieves much better Bayes regret (right panel) than all algorithms. This implies that the (potentially mis-specified) shared prior across products is informative, and thus leveraging shared structure may be valuable. Then, by design, our \covmpdp~learns this shared structure, incurring meta regret that grows sublinearly in $N$ (left panel). Consistent with our results on synthetic data, we see that the \covmpdp~significantly outperforms the benchmark algorithms; this is true even though the multivariate Gaussian prior that we estimate may not be the true prior. This result suggests that our proposed algorithms may be robust to model misspecification of the prior.

\section{Discussion \& Conclusions} \label{sec:conc}

Firms are increasingly performing experimentation. This provides an opportunity for decision-makers to learn not just \textit{within} experiments, but also \textit{across} experiments. In this paper, we consider the multi-product dynamic pricing setting where a decision-maker must learn a sequence of related unknown parameters through experimentation; we capture the relationship across these unknown parameters by imposing that they arise from a shared distribution (the prior). We propose meta-learning policies that efficiently learn both the shared distribution across experiments and the individual unknown parameters within experiments.

Our meta-learning approach can easily be adapted beyond dynamic pricing applications to classical multi-armed and contextual bandit problems as well. For instance, consider clinical trials, which were the original motivation for bandit problems \citep{thompson1933likelihood, lai1985asymptotically}. Many have argued the benefits of Bayesian clinical trials, which allow for the use of historical information and for synthesizing results of past relevant trials, \eg, past clinical trials on the same disease may indicate that patients with certain biomarkers or concomitant medications are less likely to benefit from standard therapy. Such information can be encoded in a Bayesian prior to potentially allow for more informative clinical trials and improved treatment allocations to patients within the trial \citep[see, e.g.,][]{berry2006bayesian, anderer2019adaptive}. Our meta-learning approach can inform how such priors are constructed. Importantly, prior widening gracefully transitions from an uninformative to an informative prior as we accrue data from more related clinical trials.

Our prior widening technique is inspired by the emerging literature studying prior misspecification in Thompson sampling. In general, adopting a more conservative prior allows Thompson sampling to still achieve the optimal theoretical guarantee, while a less conservative prior may cause failure to converge \citep{honda2014optimality,liu2015prior}. However, the use of a conservative prior often results in poor empirical performance, and can erode the benefit of using Thompson sampling over UCB and other prior-free approaches \citep[see, \eg,][]{RVR14,bastani2020mostly}. We take the view that a successful implementation of Thompson sampling \textit{requires} learning an appropriate prior, and propose meta-learning policies to achieve this goal across a sequence of learning problems.

\ACKNOWLEDGMENT{The authors gratefully acknowledge Columbia University Center for Pricing and Revenue Management for providing us the dataset on auto loans. We are also grateful to Amit Peleg, Jackie Baek, Omar Besbes, Dan Russo, various seminar participants and an anonymous review team for valuable feedback on earlier drafts.}


{\SingleSpacedXI
\bibliographystyle{ormsv080} 
\bibliography{refs} 
}

%
%
%
\newpage
\begin{APPENDIX}{}
We begin by defining some helpful notation. First, let
\[ \rev\left(\theta,\hat{\theta},\Sigma,t\right)=\E\left[\sum_{s=1}^tp_{i,s}D_{i,s}(p_{i,s},x_{i,s})\right] \,,\]
be the expected total revenue over $t$ time steps obtained by running TS$(\N(\hat{\theta},\Sigma),0)$ --- the Thompson sampling algorithm in Algorithm \ref{alg:oracle} with the (possibly incorrect) prior $\N\left(\hat{\theta},\Sigma\right)$ and exploration parameter $\lambda_e=0$ --- in an epoch with true parameter $\theta$. Second, let 
\[\rev_*\left(\theta,t\right)=\E\left[\sum_{s=1}^t p_{i,s}^* D_{i,s}(p_{i,s}^*,x_{i,s})\right] \,,\]
be the expected total revenue over $t$ time steps obtained by the oracle --- recall $p^*_{i,s}$ is the oracle price defined in Eq. \eqref{eq:oracle-price} --- in an epoch with true parameter $\theta.$

All norms $\|\cdot\|$ refer to the $\ell_2$ norm unless stated otherwise.

\section{Meta oracle Regret Analysis}\label{sec:theorem:oracle}

We first state the following lemma, whose proof is provided in Section \ref{sec:lemma:ini}. 
\begin{lemma}\label{lemma:ini}
For any epoch $i\in[N],$ the length of the random exploration periods $\T_i$ is upper bounded by 
\begin{align}\label{eq:t_e}
\T_e=\max\left\{6\log_{e/2}(dNT)/c_1,2\lambda_e/c_0\right\}
\end{align}
with probability at least $1-2/(N^3T^2)$. The constants are given by
\[c_0=\frac{\lambda_0}{3}\left[\frac{p^2_{\max}+p^2_{\min}+2}{2}-\sqrt{\left(\frac{p^2_{\max}+p^2_{\min}+2}{2}\right)^2-(p_{\max}-p_{\min})^2}\right] , \quad c_1=\frac{c_0}{(1+p^2_{\max})x^2_{\max}} \,.\]
\end{lemma}
In other words, we incur at most logarithmic regret due to the initial random exploration in Algorithm \ref{alg:oracle}.

\proof{Proof of Theorem \ref{theorem:oracle}} The proof proceeds in three steps. We first show that the regret incurred in the initial random exploration steps is negligible. We then map the remaining regret to a linear bandit formulation, and bound the resulting terms.

First, define the event
\begin{align} \label{event-A}
\mathcal{A}=\{\T_i\leq\T_e~\forall i\in[N]\} \,.
\end{align}
By Lemma \ref{lemma:ini}, $\Pr(\neg \mathcal{A}) \leq 2/(NT)^2$. We can decompose the regret from Algorithm \ref{alg:oracle} into exploration and non-exploration periods, conditioned on whether or not $\mathcal{A}$ holds:
\begin{align}
\nonumber& \underset{\theta_i\sim\N\left(\theta_*,\Sigma_*\right)}{\E}\left[\rev_*\left(\theta_i,T\right)-\sum_{t=1}^{\T_i}p^{\oracle}_{i,t}D_{i,t}(p^{\oracle}_{i,t},x_{i,t})-\rev\left(\theta_i,\theta^{\oracle}_{i,\T_i},\Sigma^{\oracle}_{i,\T_i},T-\T_i\right)\right]\\
\nonumber&= \E\left[\rev_*\left(\theta_i,T\right)-\sum_{t=1}^{\T_i}p^{\oracle}_{i,t}D_{i,t}(p^{\oracle}_{i,t},x_{i,t})-\rev\left(\theta_i,\theta^{\oracle}_{i,\T_i},\Sigma^{\oracle}_{i,\T_i},T-\T_i\right)\middle|\neg \mathcal{A}\right]\Pr(\neg \mathcal{A})\\
&\nonumber \quad+ {\E}\left[\rev_*\left(\theta_i,\T_i\right)-\sum_{t=1}^{\T_i}p^{\oracle}_{i,t}D_{i,t}(p^{\oracle}_{i,t},x_{i,t})\middle| \mathcal{A}\right] +{\E}\left[\left(\rev_*\left(\theta_i,T-\T_i\right)-\rev\left(\theta_i,\theta^{\oracle}_{i,\T_i},\Sigma^{\oracle}_{i,\T_i},T-\T_i\right)\right)\middle|\mathcal{A}\right]\\
\nonumber&\leq {\E}\left[\frac{2p_{\max}x_{\max}\sqrt{1+p^2_{\max}}\|\theta_i\|}{2N^2T}\right]+ {\E}\left[2p_{\max}x_{\max}\sqrt{1+p^2_{\max}}\T_e\|\theta_i\|\right]\\
\label{eq:oracle6}&\quad +{\E}\left[\left(\rev_*\left(\theta_i,T-\T_i\right)-\rev\left(\theta_i,\theta^{\oracle}_{i,\T_i},\Sigma^{\oracle}_{i,\T_i},T-\T_i\right)\right)\middle|\mathcal{A}\right] \,,
\end{align}
where we have used the facts that $\Pr(\neg \mathcal{A}) \leq 2/(NT)^2$, the worst-case regret achievable in a single time period is $2p_{\max}x_{\max}\sqrt{1+p^2_{\max}}\|\theta_i\|$, and $\T_i \leq \T_e$ on the event $\mathcal{A}$.

The first two terms in Eq. \eqref{eq:oracle6} are $O(1/(N^2T)) + O(\log(dNT)) = \tilde{O}(1)$. To analyze the third term in Eq. \eqref{eq:oracle6}, we construct a mapping between the dynamic pricing and linear bandit problems, in order to leverage existing results on TS and UCB for linear bandits \citep{RVR14,AYPS11}. In particular, we can map the Bayes regret of an epoch
\[ \underset{\theta_i\sim\N\left(\theta_*,\Sigma_*\right)}{\E}\left[\left(\rev_*\left(\theta_i,T-\T_i\right)-\rev\left(\theta_i,\theta^{\oracle}_{i,\T_i},\Sigma^{\oracle}_{i,\T_i},T-\T_i\right)\right)\middle|E\right] \,,\]
to the Bayes regret of the Thompson sampling algorithm \citep{RVR14} for a linear bandit instance as follows. Let the unknown parameter $\theta=\begin{pmatrix}\alpha^{\top}&~\beta^{\top}\end{pmatrix}^{\top}$ be drawn from the prior $\N\left(\theta_*,\Sigma_*\right)$. Take the decision set to be $A_t=\{(p\vx_{i,t};p^2\vx_{i,t}):p\in[p_{\min},p_{\max}]\}$, where $x_{i,t}$ is the feature vector drawn i.i.d from the feature distribution. Note that the magnitude of the $\ell_2$-norm of an action is at most $p_{\max}\sqrt{1+p^2_{\max}}x_{\max}$ and the noise terms are conditionally $(p_{\max}\sigma)$-subgaussian.

Using this mapping, by Theorem 3 of \cite{AYPS11} and Lemma \ref{aux_thm1} in Appendix \ref{sec:aux_results}, the Bayes regret of an epoch is upper bounded as 
\begin{align}
\nonumber&\underset{\theta_i\sim\N\left(\theta_*,\Sigma_*\right)}{\E}\left[\left(\rev_*\left(\theta_i,T-\T_i\right)-\rev\left(\theta_i,\theta^{\oracle}_{i,\T_i},\Sigma^{\oracle}_{i,\T_i},T-\T_i\right)\right)\middle|E\right]\\
\label{eq:oracle1}&={\E}\left[\widetilde{O}\left(\|\theta\|\sqrt{dT}\left(\|\theta\|+\sqrt{d}\right)\right)\right] ~=~ {\E}\left[\widetilde{O}\left(\|\theta\|^2\sqrt{dT}+\|\theta\|d\sqrt{T}\right)\right].
\end{align}
where Eq. \eqref{eq:oracle1} follows from the facts that (i) the upper bound on the regret of a linear bandit instance scales linearly with the maximum absolute value of the rewards and, (ii) the absolute value of the expected reward (revenue) for each round is upper bounded as
\begin{align}
	\max_{p\in\left[p_{\min},p_{\max}\right]}\left|\langle m,\theta\rangle\right| ~\leq~ \max_{p\in\left[p_{\min},p_{\max}\right]}\|m\|\|\theta\| ~=~ p_{\max}\sqrt{1+p^2_{\max}}x_{\max}\|\theta\| ~=~ O\left(\|\theta\|\right) \,.
\end{align}

To complete the proof, we must bound $\underset{\theta\sim\N\left(\theta_*,\Sigma_*\right)}{\E}\left[\|\theta\|^2\right]$. By the ``trace trick", we have
\begin{align}
	\nonumber\underset{\theta\sim\N\left(\theta_*,\Sigma_*\right)}{\E}\left[\|\theta\|^2\right] &= \tr\left(\underset{\theta\sim\N\left(\theta_*,\Sigma_*\right)}{\E}\left[\theta\theta^{\top}\right]\right)\\
	\nonumber &= \tr\left(\underset{\theta\sim\N\left(\theta_*,\Sigma_*\right)}{\E}\left[\left(\theta-\theta_*\right)\left(\theta-\theta_*\right)^{\top}+\theta_*\theta^{\top}+\theta\theta_*^{\top}-\theta_*\theta_*^{\top}\right]\right)\\
 \nonumber &= \tr\left(\Sigma_*+\theta_*\underset{\theta\sim\N\left(\theta_*,\Sigma_*\right)}{\E}\left[\theta^{\top}\right]+\underset{\theta\sim\N\left(\theta_*,\Sigma_*\right)}{\E}\left[\theta\right]\theta_*^{\top}-\theta_*\theta_*^{\top}\right)\\
	\nonumber&= \tr\left(\Sigma_*+2\theta_*\theta_*^{\top}-\theta_*\theta_*^{\top}\right)\\
	\nonumber &= \tr\left(\Sigma_*\right)+\tr\left(\|\theta_*\|^2\right)\\
	\label{eq:oracle4} &\leq d\overline{\lambda} +S^2 ~=~ O(d)\,,
\end{align} 
where we have used the definition of the covariance matrix $\Sigma_*=\underset{\theta\sim\N\left(\theta_*,\Sigma_*\right)}{\E}\left[\left(\theta-\theta_*\right)\left(\theta-\theta_*\right)^{\top}+\theta_*\theta^{\top}\right],$ and the last step follows from Assumptions \ref{assumption:x_norm} and \ref{assumption:Sigma}. Moreover, by Cauchy-Schwarz inequality, we have
\begin{align}
	\label{eq:oracle5}\underset{\theta\sim\N\left(\theta_*,\Sigma_*\right)}{\E}\left[\|\theta\|\right] ~\leq~ \sqrt{{\E}\left[\|\theta\|^2\right]} ~\leq~ \sqrt{d\overline{\lambda}+S^2} ~=~ O\left(\sqrt{d}\right).
\end{align}
Substituting Eqs. \eqref{eq:oracle4} and \eqref{eq:oracle5} into Eq.  \eqref{eq:oracle1}, we obtain that the third term of Eq. \eqref{eq:oracle6} is $\tilde O(d^{3/2}T^{1/2})$. Noting that the first and second terms of Eq. \eqref{eq:oracle6} contribute $\tilde{O}(1)$ regret, we can bound the total regret of each epoch as $\tilde O(d^{3/2}T^{1/2})$.

Since each epoch is mutually independent, the Bayes regret of Algorithm \ref{alg:oracle} over all $N$ epochs is simply $N\times\tilde O(d^{3/2}T^{1/2})=\tilde O(d^{3/2}NT^{1/2})$.
\Halmos
\endproof

\subsection{Proof of Lemma \ref{lemma:ini}}\label{sec:lemma:ini}

Recall that $V_{i,t}=\sum_{s=1}^{t}\begin{pmatrix}
x^{\top}_{i,s}&~p_{i,s}x_{i,s}^{\top}
\end{pmatrix}^{\top}\begin{pmatrix}
x^{\top}_{i,s}&~p_{i,s}x_{i,s}^{\top}
\end{pmatrix}$ is the Fisher information matrix of epoch $i$ after time step $t$. Lemma \ref{lemma:ini} states that $\lambda_{\min}(V_{i,\T_e}) \geq \lambda_e$ with high probability. Since $V_{i,t}$ is a random matrix, we will apply the following matrix Chernoff inequality to lower bound its minimum eigenvalue (Note that $\lambda_{\max}\left(\begin{pmatrix}
x^{\top}_{i,s}&~p_{i,s}x_{i,s}^{\top}
\end{pmatrix}^{\top}\begin{pmatrix}
x^{\top}_{i,s}&~p_{i,s}x_{i,s}^{\top}
\end{pmatrix}\right)\leq(1+p^2_{\max})x^2_{\max}$).

\begin{lemma}[Theorem 3.1 of \citealp{T11}]
	\label{lemma:matrix_chernoff}
	 For any $\zeta\in[0,1)$, any real number $u,$ and any $t\leq\T_i$
	\begin{align*}
	\Pr\left(\lambda_{\min}(V_{i,t}) ~\geq~ (1-\zeta)u \text{ and } \lambda_{\min}\left(\E\left[V_{i,t}\right])~\geq~u\right)\right) ~\geq ~ 1-d\left(\frac{\exp(-\zeta)}{(1-\zeta)^{1-\zeta}}\right)^{c_1u/c_0} \,.
	\end{align*}
\end{lemma}

The above lemma states that the probability that $\lambda_{\min}(V_{i,t})$ is much less than 	$\lambda_{\min}\left(\E\left[V_{i,t}\right]\right)$ is small. To apply the above result, we must first lower bound the minimum eigenvalue of $\E\left[V_{i,t}\right]$:

\begin{lemma}
	\label{lemma:min_eig}
	For all $t\leq\T_i,$ the minimum eigenvalue of $\E\left[V_{i,t}\right]$ is lower bounded as
	\begin{align*}
	\lambda_{\min}\left(\E\left[V_{i,t}\right]\right)\geq c_0t.
	\end{align*}	
\end{lemma} 
\proof{Proof of Lemma \ref{lemma:min_eig}}
	From linearity of expectation, we have 
	\begin{align*}
	\E\left[V_{i,t}\right] &= \sum_{\tau\text{ even, } \tau\leq t}\E\left[\begin{pmatrix}
	x_{i,\tau}\\p_{i,t}x_{i,\tau}
	\end{pmatrix}\begin{pmatrix}
	x^{\top}_{i,\tau}&~p_{i,\tau}x^{\top}_{i,\tau}
	\end{pmatrix}\right]+\sum_{\tau\text{ odd, } \tau\leq i}\E\left[\begin{pmatrix}
	x_{i,\tau}\\p_{i,t}x_{i,\tau}
	\end{pmatrix}\begin{pmatrix}
	x^{\top}_{i,\tau}&~p_{i,\tau}x^{\top}_{i,\tau}
	\end{pmatrix}\right]\\
	& \geq \frac{t}{3}\left(\begin{pmatrix}
	\E[\vx_{i,1}\vx_{i,1}^{\top}]&p_{\min}\E[\vx_{i,1}\vx_{i,1}^{\top}]\\
	p_{\min}\E[\vx_{i,1}\vx_{i,1}^{\top}]&p^2_{\min}\E[\vx_{i,1}\vx_{i,1}^{\top}]
	\end{pmatrix}+\begin{pmatrix}
	\E[\vx_{i,1}\vx_{i,1}^{\top}]&p_{\max}\E[\vx_{i,1}\vx_{i,1}^{\top}]\\
	p_{\max}\E[\vx_{i,1}\vx_{i,1}^{\top}]&p^2_{\max}\E[\vx_{i,1}\vx_{i,1}^{\top}]
	\end{pmatrix}\right)\\
	&= \frac{t}{3}\begin{pmatrix}
	2\E[\vx_{i,1}\vx_{i,1}^{\top}]&\left(p_{\min}+p_{\max}\right)\E[\vx_{i,1}\vx_{i,1}^{\top}]\\
	\left(p_{\min}+p_{\max}\right)\E[\vx_{i,1}\vx_{i,1}^{\top}]&\left(p^2_{\min}+p^2_{\max}\right)\E[\vx_{i,1}\vx_{i,1}^{\top}]
	\end{pmatrix}\\
	&= \frac{t}{3}\begin{pmatrix}
	2&\left(p_{\min}+p_{\max}\right)\\
	\left(p_{\min}+p_{\max}\right)&\left(p^2_{\min}+p^2_{\max}\right)
	\end{pmatrix}\otimes\E[\vx_{i,1}\vx_{i,1}^{\top}] \,.
	\end{align*}
We can compute the minimum eigenvalue of $\begin{pmatrix}
	2&\left(p_{\min}+p_{\max}\right)\\
	\left(p_{\min}+p_{\max}\right)&\left(p^2_{\min}+p^2_{\max}\right)
	\end{pmatrix}$ to be 
\[\frac{p^2_{\max}+p^2_{\min}+2}{2}-\sqrt{\left(\frac{p^2_{\max}+p^2_{\min}+2}{2}\right)^2-(p_{\max}-p_{\min})^2} \,.\]
Note that the eigenvalues of a symmetric positive semi-definite matrix coincide with its singular values. Thus, we can apply Lemma \ref{lemma:kronecker} to obtain that the minimum eigenvalue of $\E\left[V_{i,t}\right]$ is at least
\begin{align*}
\lambda_{\min} \left( \E\left[V_{i,t}\right] \right) &\geq \frac{t}{3} \cdot \lambda_{\min} \begin{pmatrix}
	2&\left(p_{\min}+p_{\max}\right)\\
	\left(p_{\min}+p_{\max}\right)&\left(p^2_{\min}+p^2_{\max}\right)
	\end{pmatrix} \cdot  \lambda_{\min}\left(\E[\vx_{i,1}\vx_{i,1}^{\top}]\right)\\
	& \geq \frac{t\lambda_0}{3}\left[\frac{p^2_{\max}+p^2_{\min}+2}{2}-\sqrt{\left(\frac{p^2_{\max}+p^2_{\min}+2}{2}\right)^2-(p_{\max}-p_{\min})^2}\right] \,,
\end{align*}
where we have used Assumption \ref{assumption:pos-def}.
	\Halmos
\endproof

\proof{Proof of Lemma \ref{lemma:ini}} Taking $\zeta=1/2$ in Lemma \ref{lemma:matrix_chernoff} and substituting the result from Lemma \ref{lemma:min_eig}, we have
\begin{align*}
\Pr\left(\lambda_{\min}(V_{i,t})\geq\frac{c_0t}{2}\right)\geq1-2d\left(\frac{e}{2}\right)^{-\frac{c_1t}{2}}.
\end{align*}
Setting $t=\T_e=\max\left\{6\log_{e/2}(dNT)/c_1,\max 2\lambda_e/c_0\right\},$ this implies
\begin{align*}
\Pr\left(\lambda_{\min}(V_{i,\T_e}) ~\geq~ \lambda_e\right) ~\geq~ 1-\frac{2}{N^3T^2},
\end{align*}
and we can conclude the proof.
\Halmos
\endproof

\section{Convergence of Prior Mean Estimate}\label{sec:lemma:deviation}

Lemma \ref{lemma:deviation} shows that, after observing $i$ epochs of length $T$, our estimate $\hat{\theta}_i$ of the unknown prior mean $\theta_*$ is close with high probability. To prove Lemma \ref{lemma:deviation}, we first focus on the case where the event $\mathcal{A}$ defined in Eq. \eqref{event-A} holds. We will show that at the end of each epoch, our estimated parameter vector $\dot{\theta}_i$ is probably close to the true parameter vector $\theta_i$ (Lemma \ref{lemma:ols}), which implies that the average of our estimated parameters from each epoch $\frac{1}{i}\sum_{j=1}^i \dot{\theta}_j$ is probably close to the average of the true parameters from each epoch $\frac{1}{i}\sum_{j=1}^i \theta_j$ (Lemma \ref{lemma:ols1}). Next, we will show that the latter term $\frac{1}{i}\sum_{j=1}^i \theta_j$ is a good approximation of $\theta_*$ (Lemma \ref{lemma:ols2}). Combining these steps via a triangle inequality and accounting for the probability $\mathcal{A}$ does not hold yields the result in Lemma \ref{lemma:deviation}.

We first state two useful lemmas from the literature regarding the concentration of OLS estimates and the matrix Hoeffding bound.

\begin{lemma}
	\label{lemma:ols}
	When the event $\mathcal{A}$ holds, for any epoch $i\in[N]$ and $\delta\in[0,2/e]$, conditional on $F_i=\sigma(\dot{\theta}_1,\ldots,\dot{\theta}_{i-1})$, we have
	\begin{align} \textstyle
	\nonumber\Pr\left(\left\|\dot{\theta}_i-\theta_i\right\|\geq 2\sigma\sqrt{\frac{2d\log_e (2/\delta)}{\lambda_{e}}} ~\middle|~F_i\right)\leq\delta,
	\end{align}
\end{lemma}
\proof{Proof of Lemma \ref{lemma:ols}} When $\mathcal{A}$ holds, the random exploration periods are completed before $T$ time steps, guaranteeing that $\lambda_{\min}(V_{i,T})\geq\lambda_e$. Thus, this result follows immediately from Theorem 4.1 of \cite{ZM18}, where we note that $d+\log_e(2/\delta)\leq 2d\log_e(2/\delta)$ for $\delta<2/e$.
\Halmos
\endproof

\begin{lemma}[\citealp{JinNJ19}]\label{lemma:matrix_hoeffding}
	Let random vectors $X_1,\ldots,X_n\in\R^{d},$ satisfy that for all $i\in[n]$ and $u\in\R,$
	\[\E[X_i|\sigma(X_1,\ldots,X_{i-1})]=0,\quad\Pr\left(\|X_i\|\geq u\middle|\sigma(X_1,\ldots,X_{i-1})\right)\leq2\exp\left(-\frac{u^2}{2\sigma_i^2}\right) \,,\]
	then for any $\delta>0,$
	\[ \Pr\left(\left\|\sum_{i\in[n]}X_i\right\|\leq4\sqrt{\sum_{i\in[n]}\sigma_i^2\log_e(2d/\delta)}\right)\geq1-\delta \,.\]
\end{lemma}

We now show that the average of our estimated parameters from each epoch is close to the average of the true parameters from each epoch with high probability.

\begin{lemma}\label{lemma:ols1}
	When the event $\mathcal{A}$ holds, for any $i \geq 2$, the following holds with probability at least $1-\delta$:
	\begin{align*}
	\left\|\frac{1}{i} \sum_{j=1}^{i}\left(\dot{\theta}_j-\theta_j\right)\right\|\leq8\sigma\sqrt{\frac{d\log_e(4d/\delta)}{\lambda_ei}} \,.
	\end{align*}
\end{lemma}

\proof{Proof of Lemma \ref{lemma:ols1}}
By Lemma \ref{lemma:ols}, we have for any $u\in\R$,
\[ \Pr(\|\dot{\theta}_i-\theta_i\|\geq u \mid F_i) ~\leq~ 2\exp(-\lambda_eu^2/8d\sigma^2) \,.\]
Furthermore, since the OLS estimator is unbiased, $\E[\dot{\theta}_i|F_i]=\theta_i$. Thus, we can apply the matrix Hoeffding inequality (Lemma \ref{lemma:matrix_hoeffding}) to obtain
\begin{align*}
\Pr\left(\left\|\frac{1}{i-1}\sum_{j=1}^{i-1}(\dot{\theta}_i-\theta_i)\right\|~\leq~ 8\sqrt{\frac{\sigma^2d\log_e(4d/\delta)}{\lambda_e(i-1)}}\right) ~\geq~ 1-\delta.
\end{align*}
Noting that $i\leq 2(i-1)$ for all $i\in\{2,\ldots,N\}$ concludes the proof.
\Halmos
\endproof

\begin{lemma}\label{lemma:ols2}
	When the event $\mathcal{A}$ holds, for any $i \geq 2$, the following holds with probability at least $1-\delta$:
	\begin{align*}
	\left\|\frac{1}{i}\sum_{j=1}^{i}\theta_j-\theta_*\right\| ~\leq~ 8\sqrt{\frac{5\overline{\lambda}d\log_e(4d/\delta)}{i}} \,.
	\end{align*}
\end{lemma}

\proof{Proof of Lemma \ref{lemma:ols2}} We first show a concentration inequality for the quantity $\|\theta_j-\theta_*\|$ similar to that of Lemma \ref{lemma:ols}. Note that for any unit vector $s\in\R^{2d},$ $u^{\top}(\theta_i-\theta_*)$ is a zero-mean normal random variable with variance at most $\overline{\lambda}.$ Therefore, for any $u\in\R,$
\begin{align}\label{eq:ols2_1}
\Pr\left(|s^{\top}(\theta_j-\theta_*)|\geq u\right)\leq 2\exp\left(-\frac{u^2}{2\overline{\lambda}}\right) \,.
\end{align}
Consider $W,$ a $(1/2)$-cover of the unit ball in $\R^{2d}.$ We know that $|W|\leq 4^{2d}.$ Let $s(\theta_j)=\theta_j-\theta_*/\|\theta_j-\theta_*\|,$ then there exists $w_{s(\theta_j)}\in W,$ such that $\|w_{s(\theta_j)}-s(\theta_j)\|\leq1/2$ by definition of $W.$ Hence,
\begin{align*}
\|\theta_j-\theta_*\|=\langle s(\theta_j),\theta_j-\theta_*\rangle=\langle s(\theta_j)-w_{s(\theta_j)},\theta_j-\theta_*\rangle+\langle w_{s(\theta_j)},\theta_j-\theta_*\rangle\leq\frac{\|\theta_j-\theta_*\|}{2}+\langle w_{s(\theta_j)},\theta_j-\theta_*\rangle \,.
\end{align*}
Rearranging the terms yields
\begin{align*}
\|\theta_j-\theta_*\| \leq 2\langle w_{s(\theta_j)},\theta_j-\theta_*\rangle \,.
\end{align*}
Applying an union bound to all possible $w\in W$ with inequality \eqref{eq:ols2_1}, we have for any $u\in\R,$
\begin{align*}
\Pr(\|\theta_j-\theta_*\|\geq u) &\leq\Pr(\exists w\in W:~\langle w,\theta_j-\theta_*\rangle\geq u/2) \\
&\leq2\cdot4^{2d}\exp\left(-\frac{u^2}{2\overline{\lambda}}\right) \\
&\leq\exp\left(5d-\frac{u^2}{2\overline{\lambda}}\right) \,.
\end{align*}
If $u^2\leq10\overline{\lambda}d,$ we have
\[ \Pr(\|\theta_j-\theta_*\|\geq u) ~\leq~1 ~\leq~ 2\exp\left(-\frac{u^2}{20\overline{\lambda}d}\right) \,; \]
else if $u^2=10\overline{\lambda}d+v$ for some $v\geq0,$ we have
\begin{align*}
\Pr(\|\theta_j-\theta_*\|\geq u) &\leq \exp\left(-\frac{v}{2\overline{\lambda}}\right)\\
&\leq 2\exp\left(-\frac{u^2}{20\overline{\lambda}d}\right) \,.
\end{align*}
Thus, for any $u\in\R,$ we can write
\begin{align}\label{eq:theta_dev}
\Pr(\|\theta_j-\theta_*\|\geq u)\leq2\exp\left(-\frac{u^2}{20\overline{\lambda}d}\right) \,.
\end{align}
Applying Lemma \ref{lemma:matrix_hoeffding}, we have
\begin{align*}
\Pr\left(\left\|\frac{\sum_{j=1}^{i-1}\theta_j}{i-1}-\theta_*\right\|\leq4\sqrt{\frac{10\overline{\lambda}d\log_e(4d/\delta)}{i-1}}\right)\geq1-\delta.
\end{align*}
The proof can be concluded by the observation $i\leq 2(i-1)$ for all $i\in\in\{2,\ldots,N\}.$
\Halmos
\endproof

We can now combine Lemmas \ref{lemma:ini}, \ref{lemma:ols1} and \ref{lemma:ols2} to prove Lemma \ref{lemma:deviation}.

\proof{Proof of Lemma \ref{lemma:deviation}}
When the event $\mathcal{A}$ holds, we can use the triangle inequality and a union bound over Lemmas \ref{lemma:ols1} and \ref{lemma:ols2} to obtain
\begin{align*}
\left\|\hat{\theta}_i-\theta_*\right\| &= \left\|\frac{\sum_{j=1}^{i-1}\dot{\theta}_j}{i-1}-\frac{\sum_{j=1}^{i-1}\theta_j}{i-1}+\frac{\sum_{j=1}^{i-1}\theta_j}{i-1}-\theta_*\right\| \\
&\leq \left\|\frac{1}{i-1}\sum_{j=1}^{i-1}\left(\dot{\theta}_j-\theta_j\right)\right\|+\left\|\frac{1}{i-1}\sum_{j=1}^{i-1}\theta_j-\theta_*\right\|\\
&\leq 8\sqrt{\frac{2(\sigma^2/\lambda_e+5\overline{\lambda})d\log_e(4dN/\delta)}{i}} \,,
\end{align*}
with probability at least $1-2\delta,$ where we have use the fact that $\sqrt{a}+\sqrt{b}\leq \sqrt{2(a+b)}$. By Lemma \ref{lemma:ini}, the event $\mathcal{A}$ does not hold with probability at most $2/(N^2T^2)$. Thus, a second union bound yields the result.
\Halmos
\endproof

\section{\texttt{Meta-DP} Regret Analysis}
 \label{sec:theorem:main_regret}

Appendix \ref{app:meta-int} provides the proof of Lemma \ref{lemma:main1} and the statement of an intermediate Lemma \ref{lemma:ind-ts}. Appendix \ref{app:meta-proof} provides the proof of Theorem \ref{theorem:main_regret}, following the proof strategy outlined in Section \ref{sec:meta_regret}. 

\subsection{Intermediate Lemmas} \label{app:meta-int}

Recall that for any $t \in \{\T_i+1, \cdots, T\}$, the meta oracle maintains and samples from its posterior $\N\left(\theta^{\oracle}_{i,t},\Sigma^{\oracle}_{i,t}\right)$ (see Algorithm \ref{alg:oracle}), while our \mpdp~maintains and samples parameters from its posterior $\N\left(\theta^{\md}_{i,t},\Sigma^{\md}_{i,t}\right)$ (see Algorithm \ref{alg:main}). Lemma \ref{lemma:main1} in Section \ref{sec:meta_regret} established the difference in Bayesian posteriors between the meta oracle and our \mpdp. The proof follows from the standard update rules for Bayesian linear regression and is given below.

\proof{Proof of Lemma \ref{lemma:main1}}

Using the posterior update rule for Bayesian linear regression \citep{Bishop06}, the posterior of the oracle at $t = \T_i+1$  is
\begin{align*}
\theta^{\oracle}_{i,\T_i+1}=&\left(\Sigma_*^{-1}+\sigma\sum_{t=1}^{\T_i}m_{i,t}m_{i,t}^{\top}\right)^{-1}\left(\Sigma_*^{-1}\theta_*+\sigma\sum_{t=1}^{\T_i}m_{i,t}D_{i,t}\right)\\
=&\left(\Sigma_*^{-1}+\sigma\sum_{t=1}^{\T_i}m_{i,t}m_{i,t}^{\top}\right)^{-1}\left(\Sigma_*^{-1}\theta_*+\sigma\sum_{t=1}^{\T_i}m_{i,t}m^{\top}_{i,t}\theta_i+\sigma\sum_{t=1}^{\T_i}m_{i,t}\vep^{\oracle}_{i,t}\right),\\
\Sigma^{\oracle}_{i,\T_i+1}=&\left(\Sigma_*^{-1}+\sigma\sum_{t=1}^{\T_i}m_{i,t}m_{i,t}^{\top}\right)^{-1}.
\end{align*}
Similarly, the posterior of the \mpdp~at $t=\T_i+1$ is
\begin{align*}
\theta^{\md}_{i,\T_i+1}=&\left(\Sigma_*^{-1}+\sigma\sum_{t=1}^{t_e}m_{i,t}m_{i,t}^{\top}\right)^{-1}\left(\Sigma_*^{-1}\hat{\theta}_i+\sigma\sum_{t=1}^{\T_i}m_{i,t}D_{i,t}\right)\\
=&\left(\Sigma_*^{-1}+\sigma\sum_{t=1}^{\T_i}m_{i,t}m_{i,t}^{\top}\right)^{-1}\left(\Sigma_*^{-1}\hat{\theta}_i+\sigma\sum_{t=1}^{\T_i}m_{i,t}m^{\top}_{i,t}\theta_i+\sigma\sum_{t=1}^{\T_i}m_{i,t}\vep^{\md}_{i,t}\right),\\
\Sigma^{\md}_{i,\T_i+1}=&\left(\Sigma_*^{-1}+\sigma\sum_{t=1}^{\T_i}m_{i,t}m_{i,t}^{\top}\right)^{-1}.
\end{align*}
The result follows directly.
\Halmos
\endproof

We also note that the prior-independent Thompson sampling algorithm employed in the exploration epochs satisfies a meta regret guarantee:
\begin{lemma}
	\label{lemma:ind-ts}
	The meta regret of the prior-independent Thompson sampling algorithm in a single epoch is $\widetilde{O}(d^{2}T^{1/2}).$
\end{lemma}
The proof can be easily adapted from the literature \citep[see, \eg,][]{agrawal2013thompson,AbeilleL17}, and is thus omitted. We note that our normalization implies $\E[\|\theta\|]=\Theta(d^{1/2})$). Lemma \ref{lemma:ind-ts} ensures that we accrue at most $\widetilde{O}(d^2N_0\sqrt{T})$ regret in the $N_0$ exploration epochs; from Eq. \eqref{eq:N_0}, we know that $N_0$ grows merely poly-logarithmically in $N$ and $T$.

\subsection{Proof of Theorem \ref{theorem:main_regret}} \label{app:meta-proof}

Consider any non-exploration epoch $i\geq N_0+1$. If upon completion of all exploration steps at time $\T_i+1$, we have that the posteriors of the meta oracle and our \mpdp~coincide --- \ie, $(\theta^{\md}_{i,\T_i+1},\Sigma^{\md}_{i,\T_i+1})=(\theta^{\oracle}_{i,\T_i+1},\Sigma^{\oracle}_{i,\T_i+1})$ --- then both policies would achieve the same expected revenue over the time periods $\T_i+1, \cdots, T$, \ie, we would have
\[\rev\left(\theta_i,\theta^{\md}_{i,\T_i+1},\Sigma^{\md}_{i,\T_i+1},T-\T_i\right)=\rev\left(\theta_i,\theta^{\oracle}_{i,\T_i+1},\Sigma^{\oracle}_{i,\T_i+1},T-\T_i\right)\,.\]
By Lemma \ref{lemma:main1}, we know that $\Sigma^{\oracle}_{i,\T_i+1} = \Sigma^{\md}_{i,\T_i+1}$ always, so all that remains is establishing when $\theta^{\oracle}_{i,\T_i+1} =  \theta^{\md}_{i,\T_i+1}$.

Since the two algorithms begin with different priors but encounter the same covariates $\{x_{i,t}\}_{t=1}^T$ and take the same decisions in $t \in \{1, \cdots, \T_i\}$, their posteriors can only align at time $\T_i+1$ due to the stochasticity in the observations $\vep_{i,t}$.
As shown in Eq. \eqref{eq:main_key} in Section \ref{sec:meta_regret}, alignment occurs with $\theta^{\oracle}_{i,\T_i+1} =  \theta^{\md}_{i,\T_i+1}$ if
\begin{align*}
\chi^{\md}_{i}-\chi^{\oracle}_{i}=\frac{1}{\sigma}(M_i^{\top}M_i)^{-1}M_i^{\top}\Sigma_*^{-1}\left(\theta_*-\hat{\theta}_i\right)\,,
\end{align*}
where we recall $\chi^{\oracle}, \chi^{\md}_{i}$ were defined in Eqs. \eqref{eq:chi-oracle}-\eqref{eq:chi-md}.
 
Now, we start by defining the clean event
\begin{align} \label{event-E}
\mathcal{E} = \left\{ \left\|\hat{\theta}_i-\theta_*\right\| \leq 8\sqrt{\frac{2(\sigma^2/\lambda_e+5\overline{\lambda})d\log_e(4dN^2T)}{i}}\,, \quad \T_i\leq \T_e \qquad \forall i\geq N_0+1 \right\} \,,
\end{align}
which stipulates that for every epoch $i$ after the initial $N_0$ exploration epochs, (i) our estimated prior mean $\hat{\theta}_i$ is close to the unknown prior mean $\theta_*$ (which holds with high probability by Lemma \ref{lemma:deviation}), (ii) and the event $\mathcal{A}$ defined in Eq. \eqref{event-A} holds, ensuring that the number of exploration periods per epoch is small (which holds with high probability by Lemma \ref{lemma:ini}). Since $\mathcal{E}$ holds with high probability, we first focus on analyzing the meta regret conditioned on $\mathcal{E}$.

Denote the meta regret of epoch $i$ conditioned on the event $\mathcal{E}$ defined in Eq. \eqref{event-E} as $\regret_{N,T}(i) \mid \mathcal{E}$. The next lemma bounds the meta regret for any epoch $i \geq N_0$ under the event $\mathcal{E}$.

\begin{lemma}\label{lemma:main2}
	The meta regret of an epoch $i\geq N_0+1$ satisfies
\[ \regret_{N,T}(i) \mid \mathcal{E} ~=~ \tilde{O}\left(d^2\sqrt{\frac{T}{i}}+\frac{\sqrt{d}}{N}\right) \,. \]
\end{lemma}

\proof{Proof of Lemma \ref{lemma:main2}}
As noted earlier, during the exploration periods $1 \leq t  \leq \T_i$, the meta oracle and our \mpdp~encounter the same covariates $\{x_{i,t}\}_{t=1}^T$ and offer the same prices; thus, by construction, they achieve the same expected revenue and the resulting meta regret is $0$. Then, we can write
\begin{align}
\nonumber\regret_{N,T}(i) \mid \mathcal{E} &= \E_{\theta_i,\hat{\theta}_i,\chi^{\oracle}_i,\chi_i^{\md}}\left[\rev\left(\theta_i,\theta^{\oracle}_{i,\T_i+1},\Sigma^{\oracle}_{i,\T_i+1},T-\T_i\right)-\rev\left(\theta_i,\theta^{\md}_{i,\T_i+1},\Sigma^{\md}_{i,\T_i+1},T-\T_i\right)\mid \mathcal{E} \right]\\
\nonumber
&= \E_{\theta_i,\hat{\theta}_i,\chi_i^{\md}}\left[\rev_*\left(\theta_i,T-\T_i\right)-\rev\left(\theta_i,\theta^{\md}_{i,\T_i+1},\Sigma^{\md}_{i,\T_i+1},T-\T_i\right)\mid \mathcal{E}\right]\\
\label{eq:main7} &\quad-\E_{\theta_i,\hat{\theta}_i,\chi^{\oracle}_i}\left[\rev_*\left(\theta_i,T-\T_i\right)-\rev\left(\theta_i,\theta^{\oracle}_{i,\T_i+1},\Sigma^{\oracle}_{i,\T_i+1},T-\T_i\right) \mid \mathcal{E}\right] \,.
\end{align}

We will use our prior alignment technique to express the first term in Eq. \eqref{eq:main7} in terms of the second term in Eq. \eqref{eq:main7}; in other words, we will use a change of measure suggested by Eq. \eqref{eq:main_key} to express the true regret of our \mpdp~as a function of the true regret of the meta oracle.

We start by expanding the first term of Eq. \eqref{eq:main7} as
\begin{align*}
& \E_{\chi_i^{\md}}\left[\rev_*\left(\theta_i,T-\T_i\right)-\rev\left(\theta_i,\theta^{\md}_{i,\T_i+1},\Sigma^{\md}_{i,\T_i+1},T-\T_i\right)\mid \mathcal{E}\right]\\
&= \int_{\chi_i^{\md}}\frac{\exp\left(-\left\|\chi_i^{\md}\right\|^2/2\sigma^2\right)}{(2\pi\sigma^2)^{t_e/2}}\left(\rev_*\left(\theta_i,T-\T_i\right)-\rev\left(\theta_i,\theta^{\md}_{i,\T_i+1},\Sigma^{\md}_{i,\T_i+1},T-\T_i\right)\right)d\chi_i^{\md} \mid \mathcal{E} \,.
\end{align*}
Given a realization of $\chi^{\md}_{i},$ we denote $\chi^{\oracle}_i(\chi^{\md}_{i})$ (with some abuse of notation) as the corresponding realization of $\chi^{\oracle}_i$ that satisfies Eq. \eqref{eq:main_key}. Note that this is a unique one-to-one mapping. We then perform a change of measure to continue:
\begin{align}
\nonumber&\int_{\chi_i^{\md}}\frac{\exp\left(-\left\|\chi_i^{\md}\right\|^2/2\sigma^2\right)}{\exp\left(-\left\|\chi_i^{\oracle}(\chi_i^{\md})\right\|^2/2\sigma^2\right)}\frac{\exp\left(-\left\|\chi_i^{\oracle}(\chi_i^{\md})\right\|^2/2\sigma^2\right)}{(2\pi\sigma^2)^{\T_i/2}}\\
\nonumber&\times\left(\rev_*\left(\theta_i,T-\T_i\right)-\rev\left(\theta_i,\theta^{\md}_{i,\T_i+1},\Sigma^{\md}_{i,\T_i+1},T-\T_i\right)d\chi_i^{\md}\right) \mid \mathcal{E}\\
\nonumber=&\int_{\left\|\chi_i^{\md}\right\|\leq4\sigma\sqrt{\T_i\log_e(2NT)}}\frac{\exp\left(-\left\|\chi_i^{\md}\right\|^2/2\sigma^2\right)}{\exp\left(-\left\|\chi_i^{\oracle}(\chi_i^{\md})\right\|^2/2\sigma^2\right)}\frac{\exp\left(-\left\|\chi_i^{\oracle}(\chi_i^{\md})\right\|^2/2\sigma^2\right)}{(2\pi\sigma^2)^{\T_i/2}}\\
\nonumber&\qquad\times\left(\rev_*\left(\theta_i,T-\T_i\right)-\rev\left(\theta_i,\theta^{\md}_{i,\T_i+1},\Sigma^{\md}_{i,\T_i+1},T-\T_i\right)d\chi_i^{\md}\right) \mid \mathcal{E}\\
\nonumber&\quad+\int_{\left\|\chi_i^{\md}\right\|\geq4\sigma\sqrt{\T_i\log_e(2NT)}}\frac{\exp\left(-\left\|\chi_i^{\md}\right\|^2/2\sigma^2\right)}{\exp\left(-\left\|\chi_i^{\oracle}(\chi_i^{\md})\right\|^2/2\sigma^2\right)}\frac{\exp\left(-\left\|\chi_i^{\oracle}(\chi_i^{\md})\right\|^2/2\sigma^2\right)}{(2\pi\sigma^2)^{\T_i/2}}\\
\nonumber&\qquad\times\left(\rev_*\left(\theta_i,T-\T_i\right)-\rev\left(\theta_i,\theta^{\md}_{i,\T_i+1},\Sigma^{\md}_{i,\T_i+1},T-\T_i\right)d\chi_i^{\md}\right) \mid \mathcal{E}\\
\nonumber\leq&\max_{\left\|\chi_i^{\md}\right\|\leq4\sigma\sqrt{\T_i\log_e(2NT)}}\exp\left(\frac{\left\|\chi_i^{\oracle}(\chi_i^{\md})\right\|^2-\left\|\chi_i^{\md}\right\|^2}{2\sigma^2}\right)\int_{\left\|\chi_i^{\md}\right\|\leq4\sigma\sqrt{\T_i\log_e(2NT)}}\frac{\exp\left(-\left\|\chi_i^{\oracle}(\chi_i^{\md})\right\|^2/2\sigma^2\right)}{(2\pi\sigma^2)^{\T_i/2}}\\
\nonumber&\qquad\times\left(\rev_*\left(\theta_i,T-\T_i\right)-\rev\left(\theta_i,\theta^{\md}_{i,\T_i+1},\Sigma^{\md}_{i,\T_i+1},T-\T_i\right)d\chi_i^{\md}\right) \mid \mathcal{E}\\
\nonumber&\quad+\int_{\left\|\chi_i^{\md}\right\|\geq4\sigma\sqrt{\T_i\log_e(2NT)}}\frac{\exp\left(-\left\|\chi_i^{\md}\right\|^2/2\sigma^2\right)}{\exp\left(-\left\|\chi_i^{\oracle}(\chi_i^{\md})\right\|^2/2\sigma^2\right)}\frac{\exp\left(-\left\|\chi_i^{\oracle}(\chi_i^{\md})\right\|^2/2\sigma^2\right)}{(2\pi\sigma^2)^{\T_i/2}}\\
\nonumber&\qquad\times\left(\rev_*\left(\theta_i,T-\T_i\right)-\rev\left(\theta_i,\theta^{\md}_{i,\T_i+1},\Sigma^{\md}_{i,\T_i+1},T-\T_i\right)d\chi_i^{\md}\right) \mid \mathcal{E}\\
\nonumber\leq&\max_{\left\|\chi_i^{\md}\right\|\leq4\sigma\sqrt{\T_i\log_e(2NT)}}\exp\left(\frac{\left\|\chi_i^{\oracle}(\chi_i^{\md})\right\|^2-\left\|\chi_i^{\md}\right\|^2}{2\sigma^2}\right)\int_{\chi_i^{\md}}\frac{\exp\left(-\left\|\chi_i^{\oracle}(\chi_i^{\md})\right\|^2/2\sigma^2\right)}{(2\pi\sigma^2)^{\T_i/2}}\\
\nonumber&\qquad\times\left(\rev_*\left(\theta_i,T-\T_i\right)-\rev\left(\theta_i,\theta^{\md}_{i,\T_i+1},\Sigma^{\md}_{i,\T_i+1},T-\T_i\right)d\chi_i^{\md}\right) \mid \mathcal{E}\\
\nonumber&\quad+\int_{\left\|\chi_i^{\md}\right\|\geq4\sigma\sqrt{\T_i\log_e(2NT)}}\frac{\exp\left(-\left\|\chi_i^{\md}\right\|^2/2\sigma^2\right)}{\exp\left(-\left\|\chi_i^{\oracle}(\chi_i^{\md})\right\|^2/2\sigma^2\right)}\frac{\exp\left(-\left\|\chi_i^{\oracle}(\chi_i^{\md})\right\|^2/2\sigma^2\right)}{(2\pi\sigma^2)^{\T_i/2}}\\
\label{eq:main10}&\qquad\times\left(\rev_*\left(\theta_i,T-\T_i\right)-\rev\left(\theta_i,\theta^{\md}_{i,\T_i+1},\Sigma^{\md}_{i,\T_i+1},T-\T_i\right)d\chi_i^{\md}\right) \mid \mathcal{E}\\
\nonumber\leq&\max_{\left\|\chi_i^{\md}\right\|\leq4\sigma\sqrt{\T_i\log_e(2NT)}}\exp\left(\frac{\left\|\chi_i^{\oracle}(\chi_i^{\md})\right\|^2-\left\|\chi_i^{\md}\right\|^2}{2\sigma^2}\right)\E_{\chi_i^{\oracle}}\left[\rev_*\left(\theta_i,T-\T_i\right)-\rev\left(\theta_i,\theta^{\oracle}_{i,\T_i+1},\Sigma^{\oracle}_{i,\T_i+1},T-\T_i\right) \mid \mathcal{E}\right]\\
\nonumber&+\E_{\chi_i^{\md}}\left[\rev_*\left(\theta_i,T-\T_i\right)-\rev\left(\theta_i,\theta^{\md}_{i,\T_i+1},\Sigma^{\md}_{i,\T_i+1},T-\T_i\right)\mid \mathcal{E} ,\left\|\chi_i^{\md}\right\|\geq4\sigma\sqrt{\T_i\log_e(2NT)}\right]\\
\label{eq:main1}&\quad\times\Pr\left(\left\|\chi_i^{\md}\right\|\geq4\sigma\sqrt{\T_i\log_e(2NT)}\right) \,.
\end{align}
Here, inequality \eqref{eq:main10} follows from the fact that $\rev_*(\theta_i,T-T_i)\geq\rev(\theta_i,\theta,\Sigma,T-T_i)$ for any choice of $\theta$ and $\Sigma.$ Thus, we have expressed the true regret of our \mpdp~as the sum of a term that is proportional to the true regret of the meta oracle, and an additional term that depends on the tail probability of $\chi_i^{\md}$. To obtain our desired bound, we will argue that (i) the coefficient of the first term decays to $1$ as the epoch number $i$ grows large, ensuring that our meta regret goes to $0$ for later epochs, and (ii) the second term is negligible with high probability since $\chi_i^{\md}$ is a subgaussian random variable.

We start by characterizing the coefficient of the first term in Eq. \eqref{eq:main1}:
\begin{align}
\nonumber&\max_{\left\|\chi_i^{\md}\right\|\leq4\sigma\sqrt{\T_i\log_e(2NT)}}\exp\left(\frac{\left\|\chi_i^{\oracle}(\chi_i^{\md})\right\|^2-\left\|\chi_i^{\md}\right\|^2}{2\sigma^2}\right)\\
\nonumber=&\max_{\left\|\chi_i^{\md}\right\|\leq4\sigma\sqrt{\T_i\log_e(2NT)}}\exp\left(\frac{\left\|n^{\md}_{i}-\frac{1}{\sigma}(M_i^{\top}M_i)^{-1}M_i^{\top}\Sigma_*^{-1}\left(\theta_*-\hat{\theta}_i\right)\right\|^2-\left\|\chi_i^{\md}\right\|^2}{2\sigma^2}\right)\\
\nonumber=&\max_{\left\|\chi_i^{\md}\right\|\leq4\sigma\sqrt{\T_i\log_e(2NT)}}\exp\left(\frac{\left(\chi_i^{\md}\right)^{\top}(M_i^{\top}M_i)^{-1}M_i^{\top}\Sigma_*^{-1}\left(\theta_*-\hat{\theta}_i\right)}{\sigma^3}+\frac{\left\|(M_i^{\top}M_i)^{-1}M_i^{\top}\Sigma_*^{-1}\left(\theta_*-\hat{\theta}_i\right)\right\|^2}{2\sigma^4}\right)\\
\nonumber\leq&\max_{\left\|\chi_i^{\md}\right\|\leq4\sigma\sqrt{\T_i\log_e(2NT)}}\exp\left(\frac{\left\|\chi_i^{\md}\right\|\left\|(M_i^{\top}M_i)^{-1}M_i^{\top}\Sigma_*^{-1}\left(\theta_*-\hat{\theta}_i\right)\right\|}{\sigma^3}+\frac{\left\|(M_i^{\top}M_i)^{-1}M_i^{\top}\Sigma_*^{-1}\left(\theta_*-\hat{\theta}_i\right)\right\|^2}{2\sigma^4}\right)\\
\label{eq:main2}=&\exp\left(\frac{4\sqrt{\T_i\log_e(2NT)}\left\|(M_i^{\top}M_i)^{-1}M_i^{\top}\Sigma_*^{-1}\left(\theta_*-\hat{\theta}_i\right)\right\|}{\sigma^2}+\frac{\left\|(M_i^{\top}M_i)^{-1}M_i^{\top}\Sigma_*^{-1}\left(\theta_*-\hat{\theta}_i\right)\right\|^2}{2\sigma^4}\right) \,.
\end{align}
Note that 
\begin{align}
\nonumber4\left\|(M_i^{\top}M_i)^{-1}M_i^{\top}\Sigma_*^{-1}\left(\theta_*-\hat{\theta}_i\right)\right\|\leq&\lambda_{\max}\left((M_i^{\top}M_i)^{-1}\right)\sqrt{\lambda_{\max}(M_iM_i^{\top})}\lambda_{\max}(\Sigma_*^{-1})\left\|\hat{\theta}_i-\theta_*\right\|\\
\nonumber\leq&32\sqrt{\frac{\T_ix^2_{\max}(1+p^2_{\max})(\sigma^2\lambda^{-1}_e+5\overline{\lambda})d\log_e(4dN^2T)}{\lambda_e^{2}\underline{\lambda}^{2}i}}\\
\label{eq:main8}\leq&c_2\sigma^2\sqrt{\frac{d\T_i\log_e(4dN^2T)}{i}} \,.
\end{align}
Furthermore, by the definition of $N_0$ in Eq. \eqref{eq:N_0}, we have for all $i\geq N_0+1,$
\begin{align}\label{eq:main3}
\frac{4\sqrt{\T_i\log_e(2NT)}\left\|(M_i^{\top}M_i)^{-1}M_i^{\top}\Sigma_*^{-1}\left(\theta_*-\hat{\theta}_i\right)\right\|}{\sigma^2}+\frac{\left\|(M_i^{\top}M_i)^{-1}M_i^{\top}\Sigma_*^{-1}\left(\theta_*-\hat{\theta}_i\right)\right\|^2}{2\sigma^4}\leq 1 \,.
\end{align}
Combining Eqs. \eqref{eq:main2} and \eqref{eq:main3}, and applying Lemma \ref{lemma:exp_algebra} in Appendix \ref{sec:aux_results} yields
\begin{align}\label{eq:main4}
\nonumber&\max_{\left\|\chi_i^{\md}\right\|\leq4\sigma\sqrt{\T_i\log_e(2NT)}}\exp\left(\frac{\left\|\chi_i^{\oracle}(\chi_i^{\md})\right\|^2-\left\|\chi_i^{\md}\right\|^2}{2\sigma^2}\right)\\
\nonumber&\leq 1+\frac{8\sqrt{\T_i\log_e(2NT)}\left\|(M_i^{\top}M_i)^{-1}M_i^{\top}\Sigma_*^{-1}\left(\theta_*-\hat{\theta}_i\right)\right\|}{\sigma^2}+\frac{\left\|(M_i^{\top}M_i)^{-1}M_i^{\top}\Sigma_*^{-1}\left(\theta_*-\hat{\theta}_i\right)\right\|^2}{\sigma^4}\\
\nonumber&\leq 1+\frac{16\sqrt{\T_i\log_e(2NT)}\left\|(M_i^{\top}M_i)^{-1}M_i^{\top}\Sigma_*^{-1}\left(\theta_*-\hat{\theta}_i\right)\right\|}{\sigma^2}\\
&\leq 1+4c_2\T_i\sqrt{\frac{d\log_e(4dN^2T)\log_e(2NT)}{i}} \,,
\end{align}
where we have used Eq. \eqref{eq:main8} in the last step. Plugging this into Eq. \eqref{eq:main1}, we can now bound
\begin{align}
\nonumber&\E_{\chi_i^{\md}}\left[\rev_*\left(\theta_i,T-\T_i\right)-\rev\left(\theta_i,\theta^{\md}_{i,\T_i+1},\Sigma^{\md}_{i,\T_i+1},T-\T_i\right)\mid \mathcal{E}\right]\\
\nonumber\leq&\left( 1+4c_2\T_i\sqrt{\frac{d\log_e(4dN^2T)\log_e(2NT)}{i}}\right)\E_{\chi_i^{\oracle}}\left[\rev_*\left(\theta_i,T-\T_i\right)-\rev\left(\theta_i,\theta^{\oracle}_{i,\T_i+1},\Sigma^{\oracle}_{i,\T_i+1},T-\T_i\right) \mid \mathcal{E}\right]\\
\nonumber&+\E_{\chi_i^{\md}}\left[\rev_*\left(\theta_i,T-\T_i\right)-\rev\left(\theta_i,\theta^{\md}_{i,\T_i+1},\Sigma^{\md}_{i,\T_i+1},T-\T_i\right)\mid \mathcal{E} ,\left\|\chi_i^{\md}\right\|\geq4\sigma\sqrt{\T_i\log_e(2NT)}\right]\\
\label{eq:main9}&\quad\times\Pr\left(\left\|\chi_i^{\md}\right\|\geq4\sigma\sqrt{\T_i\log_e(2NT)}\right) \,.
\end{align}
As desired, this establishes that the coefficient of our first term decays to $1$ as $i$ grows large. Thus, our meta regret from the first term approaches $0$ for large $i$. We now show that the second term in Eq. \eqref{eq:main9} is negligible with high probability. Similar to the proof of Lemma \ref{lemma:ols2}, for any $u\in\R,$ we can write
$\Pr\left(\left\|\chi^{\md}_i\right\|\geq u\right)\leq2\exp\left(-{u^2}/{(10\sigma^2\T_i)}\right),$ which implies
\begin{align}\label{eq:main5}
\Pr\left(\left\|\chi^{\md}_i\right\|\geq 4\sigma\sqrt{\T_i\log_e(2NT)}\right)\leq\frac{1}{NT} \,.
\end{align}
Moreover, noting that the worst-case regret achievable in a single time period is $2p_{\max}x_{\max}\sqrt{1+p^2_{\max}}\|\theta_i\|$, and $\T_i \leq \T_e$ on the event $\mathcal{E}$, we can bound
\begin{align}\label{eq:main6}
\nonumber&\E_{\chi_i^{\md}}\left[\rev_*\left(\theta_i,T-\T_i\right)-\rev\left(\theta_i,\theta^{\md}_{i,\T_e+1},\Sigma^{\md}_{i,\T_i+1},T-\T_i\right)\mid \mathcal{E},\left\|\chi_i^{\md}\right\|\geq4\sigma\sqrt{\T_i\log_e(2NT)}\right]\\
\nonumber&\leq 2(T-\T_i)p_{\max}x_{\max}\sqrt{1+p^2_{\max}}\E[\|\theta_i\|]\\
&=O(\sqrt{d} T) \,,
\end{align}
where we recall from Eq. \eqref{eq:oracle5} that $\E[\|\theta_i\|] = O(\sqrt{d})$.
Substituting Eqs. \eqref{eq:main5} and \eqref{eq:main6}, into Eq. \eqref{eq:main9}, we obtain 
\[ \left(1+4c_2\T_i\sqrt{\frac{d\log_e(4dN^2T)\log_e(2NT)}{i}}\right)\E_{\chi_i^{\oracle}}\left[\rev_*\left(\theta_i,T-\T_i\right)-\rev\left(\theta_i,\theta^{\oracle}_{i,\T_i+1},\Sigma^{\oracle}_{i,\T_i+1},T-\T_i\right) \mid \mathcal{E}\right]+O\left(\frac{\sqrt{\sqrt{d}}}{N}\right) \,.\]
Substituting the above into Eq. \eqref{eq:main7}, we can bound the meta regret of epoch $i$ as
\begin{align*}
&\regret_{N,T}(i) \mid \mathcal{E}\\
&\leq \left(4c_2\T_i\sqrt{\frac{d\log_e(4dN^2T)\log_e(2NT)}{i}}\right)\E_{\chi_i^{\oracle}}\left[\rev_*\left(\theta_i,T-\T_i\right)-\rev\left(\theta_i,\theta^{\oracle}_{i,\T_i+1},\Sigma^{\oracle}_{i,\T_i+1},T-\T_i\right) \mid \mathcal{E}\right]+O\left(\frac{\sqrt{d}}{N}\right)\\
&= \tilde{O}\left(d^2\sqrt{\frac{T}{i}}+\frac{\sqrt{d}}{N}\right) \,.
\end{align*}
Here, we have used the fact that the meta oracle's true regret is bounded (Theorem \ref{theorem:oracle}), \ie,
\[ \E_{\chi_i^{\oracle}}\left[\rev_*\left(\theta_i,T-\T_i\right)-\rev\left(\theta_i,\theta^{\oracle}_{i,\T_i+1},\Sigma^{\oracle}_{i,\T_i+1},T-\T_i\right) \mid \mathcal{E}\right] \leq \tilde{O} (d^{3/2}\sqrt{T}) \,. \]
\Halmos
\endproof

The remaining proof of Theorem \ref{theorem:main_regret} follows straightforwardly.

\proof{Proof of Theorem \ref{theorem:main_regret}} The meta regret can then be decomposed as follows:
\begin{align*}
\regret_{N,T} &=\left(\regret_{N,T} \mid \mathcal{E} \right) \Pr( \mathcal{E})+\left(\regret_{N,T} \mid \neg \mathcal{E}\right) \Pr(\neg \mathcal{E}) \\
&\leq \left(\regret_{N,T} \mid \mathcal{E} \right)+\left(\regret_{N,T} \mid \neg \mathcal{E} \right)\Pr(\neg \mathcal{E}) \,.
\end{align*}
Recall that the event $\mathcal{E}$ is composed of two events: $\mathcal{A}$ (bounded by Lemma \ref{lemma:ini}) and a bound on $\|\hat{\theta}_i - \theta_*\|$ (bounded by Lemma \ref{lemma:deviation}). Applying a union bound over the epochs $i\geq N_0+1$ to Lemma \ref{lemma:deviation} (setting $\delta=1/(N^2T)$), and applying Lemma \ref{lemma:ini} yields a bound
\[ \Pr\left( \mathcal{E} \right) \geq 1-1/(NT)-4/(NT^2) \geq 1-5/(NT) \,.\]
Recall that when the event $\mathcal{E}$ is violated, the meta regret is $O(NT),$ so we can bound $\left(\regret_{N,T} \mid \neg \mathcal{E} \right)\Pr(\neg \mathcal{E}) \leq O(NT \times 1/(NT)) =O(1)$.
Therefore, the overall meta regret is simply
\begin{align*}
\regret_{N,T}\leq\left(\regret_{N,T} \mid \mathcal{E} \right)+O(1) \,.
\end{align*}
When $N>N_0,$ applying our result in Lemma \ref{lemma:main2} yields
\begin{align*}
\sum_{i=1}^{N_0}\left(\regret_{N,T}(i) \mid \mathcal{E}\right)+\sum_{i=N_0+1}^N\left(\regret_{N,T}(i) \mid \mathcal{E} \right)+O(1)
~&\leq N_0\tilde{O}(d^2\sqrt{T})+\sum_{i=N_0+1}^N\tilde{O}\left(d^2\sqrt{\frac{T}{i}}+\frac{\sqrt{d}}{N}\right) +O(1)\\
&\leq \sum_{i=1}^N\tilde{O}\left(d^2\sqrt{\frac{T}{i}}+\frac{\sqrt{d}}{N}\right) +\tilde{O}(d^3\sqrt{T})\\
&= \tilde{O}\left(d^2(NT)^{\frac{1}{2}}+d^3\sqrt{T}\right),
\end{align*}
where we have use the fact that $\sum_{i=1}^N1/\sqrt{i}\leq 2\sqrt{N}$ in the last step.
\Halmos
\endproof

\section{Convergence of Prior Covariance Estimate}\label{sec:lemma:deviation_cov}

Lemma \ref{lemma:deviation_cov} shows that, after observing $i$ epochs of length $T$, our estimator $\hat{\Sigma}_i$ is close to $\Sigma_*$ with high probability. To prove Lemma \ref{lemma:deviation_cov}, we first focus on the case where the event $\mathcal{A}$ defined in Eq. \eqref{event-A} holds. For ease of notation, denote the average of the estimated parameters from each epoch as
\[ \bar{\theta}_i= \frac{1}{i-1}\sum_{k=1}^{i-1}\dot\theta_k \,. \]
Then, recall from the definition in Eq. \eqref{eq:ts_cov_update} that \[ \hat{\Sigma}_i=\frac{1}{i-2}\sum_{j=1}^{i-1}\left(\dot\theta_{j}-\bar{\theta}_i \right)\left(\dot\theta_{j}-\bar{\theta}_i\right)^{\top}-\frac{\sigma^2}{i-1}\sum_{j=1}^{i-1}\E\left[V^{-1}_{j,\T_j}\right] \,. \]
Then, we can expand
\begin{align}
	\nonumber\left\|\hat{\Sigma}_i-\Sigma_*\right\|_{op}&= \left\|\frac{1}{i-2}\sum_{j=1}^{i-1}\left(\dot\theta_{j}-\bar{\theta}_i\right)\left(\dot\theta_{j}-\bar{\theta}_i\right)^{\top}-\frac{\sigma^2\sum_{j=1}^{i-1}\E\left[V^{-1}_{j,\T_j}\right]}{i-1} - \Sigma_*\right\|_{op}\\
	\nonumber&= \left\|\frac{1}{i-2}\sum_{j=1}^{i-1}\left(\dot\theta_{j}-\theta_*\right)\left(\dot\theta_{j}-\theta_*\right)^{\top}-\frac{i-1}{i-2}\left(\theta_*-\bar{\theta}_i\right)\left(\theta_*-\bar{\theta}_i\right)^{\top}-\frac{\sigma^2\sum_{j=1}^{i-1}\left[V^{-1}_{j,\T_j}\right]}{i-1} -\Sigma_*\right\|_{op}\\
	\nonumber&= \left\|\frac{1}{i-2}\sum_{j=1}^{i-1}\left(\dot\theta_{j}-\theta_*\right)\left(\dot\theta_{j}-\theta_*\right)^{\top}-\frac{i-1}{i-2}\Sigma_*-\frac{\sigma^2\sum_{j=1}^{i-1}\E\left[V^{-1}_{j,\T_j}\right]}{i-2}\right.\\
	\nonumber&\quad\left.-\frac{i-1}{i-2}\left(\theta_*-\bar{\theta}_i\right)\left(\theta_*-\bar{\theta}_i\right)^{\top}+\frac{1}{i-2}\Sigma_*+\frac{\sigma^2\sum_{j=1}^{i-1}\E\left[V^{-1}_{j,\T_j}\right]}{(i-1)(i-2)}\right\|_{op}\\
	\nonumber &\leq \frac{i-1}{i-2}\left\|\frac{1}{i-1}\sum_{j=1}^{i-1}\left(\dot\theta_{j}-\theta_*\right)\left(\dot\theta_{j}-\theta_*\right)^{\top}-\Sigma_*-\frac{\sigma^2\sum_{j=1}^{i-1}\E\left[V^{-1}_{j,\T_j}\right]}{i-1}\right\|_{op}\\
	\label{eq:cov2}&\quad+ \frac{i-1}{i-2}\left\|\left(\theta_*-\bar{\theta}_i\right)\left(\theta_*-\bar{\theta}_i\right)^{\top}-\frac{1}{i-1}\Sigma_*-\frac{\sigma^2\sum_{j=1}^{i-1}\E[\left[V^{-1}_{j,\T_j}\right]}{(i-1)^2}\right\|_{op}.
\end{align}

We proceed by showing that each of the two terms is a subgaussian random variable, and therefore satisfies standard concentration results. The following lemma first establishes that both terms have expectation zero, \ie, $\hat{\Sigma}_i$ is an unbiased estimator of the true prior covariance matrix $\Sigma_*$.

\begin{lemma}\label{lemma:unbiased1}
When the event $\mathcal{A}$ holds, for any epoch $i \geq 3$,
\begin{align*}
\E\left[\frac{1}{i-1}\sum_{j=1}^{i-1}\left(\dot\theta_{j}-\theta_*\right)\left(\dot\theta_{j}-\theta_*\right)^{\top}\right] &= \Sigma_*+\frac{\sigma^2\sum_{j=1}^{i-1}\E\left[V^{-1}_{j,\T_j}\right]}{i-1} \,, \\
\E\left[\left(\theta_*-\bar{\theta}_i\right)\left(\theta_*-\bar{\theta}_i\right)^{\top}\right] &= \frac{1}{i-1}\Sigma_*+\frac{\sigma^2\sum_{j=1}^{i-1}\E\left[V^{-1}_{j,\T_j}\right]}{(i-1)^2} \,.
\end{align*}
	\end{lemma}

\proof{Proof of Lemma \ref{lemma:unbiased1}}
When $\mathcal{A}$ holds, the random exploration time steps are completed before $T$ time steps. Denote 
\begin{align} \label{def-Delta}
 \Delta_j=V^{-1}_{j,\T_j}\left(\sum_{t=1}^{\T_j}\vep_{j,t}m_{j,t}\right) = \dot\theta_j - \theta_j\,.
\end{align}
Then noting that $\E[\theta_j]=\theta_*$, $\E[\Delta_j]=0$, and $\E[\Delta_j\Delta_j^{\top}]=\sigma^2\E\left[V^{-1}_{j,\T_j}\right],$ we can write
\begin{align*}
\E\left[(\dot{\theta}_j-\theta_*)(\dot{\theta}_j-\theta_*)^{\top}\right] &= \E\left[(\theta_j+\Delta_j)(\theta_j+\Delta_j)^{\top}-\theta_*\theta^{\top}_*\right] \\
&=\E\left[\theta_j\theta_j^{\top}-\theta_*\theta^{\top}_*\right]+\E\left[\Delta_j\Delta_j^{\top}\right] \\
&=\Sigma_*+\sigma^2\E\left[V^{-1}_{j,\T_j}\right] \,.
\end{align*}
Summing over $j$ and dividing by $(i-1)$ on both sides yields the first statement. For the second statement, we can write
\begin{align*}
\E\left[(\bar{\theta}_i-\theta_*)(\bar{\theta}_i-\theta_*)^{\top}\right] &= \E\left[\bar{\theta}_i\bar{\theta}_i^{\top}-\theta_*\theta^{\top}_*\right]\\
&= \E\left[\left(\frac{\sum_{k=1}^{i-1}\dot\theta_k}{i-1}\right)\left(\frac{\sum_{k=1}^{i-1}\dot\theta_k}{i-1}\right)^{\top}-\theta_*\theta^{\top}_*\right]\\
&= \E\left[\frac{\sum_{k=1}^{i-1}\theta_k\theta_k^{\top}+\sum_{k=1}^{i-1}\Delta_k\Delta_k^{\top}+\sum_{1\leq j_1< j_2\leq i-1}\theta_{j_1}\theta^{\top}_{j_2}}{(i-1)^2}-\theta_*\theta^{\top}_*\right]\\
&= \E\left[\frac{\sum_{k=1}^{i-1}\theta_k\theta_k^{\top}+\sum_{k=1}^{i-1}\Delta_k\Delta_k^{\top}}{(i-1)^2}-\frac{1}{i-1}\theta_*\theta^{\top}_*\right]\\
&= \frac{1}{i-1}\Sigma_*+\frac{\sigma^2\sum_{k=1}^{i-1}\E\left[V^{-1}_{j,\T_j}\right]}{(i-1)^2} \,.
\end{align*}
\Halmos
\endproof

Having established that both terms in Eq. \eqref{eq:cov2} have expectation zero, the following lemma shows that these terms are subgaussian and therefore concentrate with high probability. 

\begin{lemma}\label{lemma:cov1}
When the event $\mathcal{A}$ holds, for any $\delta\in[0,1]$, the following holds with probability at least $1-2\delta$:
	\begin{align*}
	\left\|\frac{\sum_{j=1}^{i-1}\left(\dot\theta_{j}-\theta_*\right)\left(\dot\theta_{j}-\theta_*\right)^{\top}}{i-1}-\Sigma_*-\frac{\sigma^2\sum_{j=1}^{i-1}\E\left[V^{-1}_{j,\T_j}\right]}{i-1}\right\|_{op} &\leq \frac{16(\overline{\lambda}\lambda^2_e+16\sigma^2d)}{\lambda_e^2}\left(\sqrt{\frac{5d+2\log_e(2/\delta)}{i-1}}\vee\frac{5d+2\log_e(2/\delta)}{i-1}\right) \,, \\
\left\|\left(\theta_*-\bar{\theta}_i\right)\left(\theta_*-\bar{\theta}_i\right)^{\top}-\frac{1}{i-1}\Sigma_*-\frac{\sigma^2\sum_{j=1}^{i-1}\E\left[V^{-1}_{j,\T_j}\right]}{(i-1)^2}\right\|_{op} &\leq \frac{16(\overline{\lambda}\lambda^2_e+16\sigma^2d)(5d+2\log_e(2/\delta))}{\lambda_e^2(i-1)} \,.
\end{align*}
\end{lemma}

\proof{Proof of Lemma \ref{lemma:cov1}}
First, since the OLS estimator is unbiased, we have that $\E\left[\dot\theta_j-\theta_*\right]=0$ for all $j$, and consequently, $\E\left[\bar{\theta}_i-\theta_*\right]=0.$ Recall also our definition of $\Delta_j$ from Eq. \eqref{def-Delta}. Then, for any $v\in\R^{2d}$ such that $\|v\|=1,$ we can write for all $u\in\R,$
\begin{align*}
\E\left[\exp(u\langle v,\dot\theta_j-\theta_*\rangle)\right]=&\E\left[\exp(u\langle v,\theta_j-\theta_*\rangle)\exp(u\langle v,\Delta_j\rangle)\right]\\
&= \E\left[\exp(u\langle v,\theta_j-\theta_*\rangle)\right]\E\left[\exp(u\langle v,\Delta_j\rangle)\right]\\
&= \exp\left(\frac{u^2v^{\top}\Sigma_*v}{2}\right)\E\left[\exp(u\langle v,\Delta_j\rangle)\right]\\
&\leq \exp\left(u^2\left(\frac{\overline{\lambda}}{2}+\frac{8\sigma^2d}{\lambda_e^2}\right)\right) \,,
\end{align*}
where we have re-used Lemmas \ref{lemma:ols} (from Appendix \ref{sec:lemma:deviation}) and \ref{lemma:mgf} (from Appendix \ref{sec:aux_results}) in the last step. Similarly,
\begin{align*}
\E\left[\exp(u\langle v,\bar\theta-\theta_*\rangle)\right]\leq\exp\left(\frac{u^2}{i-1}\left(\frac{\overline{\lambda}}{2}+\frac{8\sigma^2d}{\lambda_e^2}\right)\right).
\end{align*}
By definition, along with Lemma \ref{lemma:unbiased1}, this implies that $\dot\theta_j-\theta_*$ is a $\left(\sqrt{(\overline{\lambda}\lambda_e^2+16\sigma^2d)/2\lambda_e^2}\right)$-subgaussian vector and, similarly $\bar\theta-\theta_*$ is a $\left(\sqrt{(\overline{\lambda}\lambda_e^2+16\sigma^2d)/[\lambda_e^2(i-1)]}\right)$-subgaussian vector. Applying concentration results for subgaussian random variables (see Lemma \ref{lemma:cov} from Appendix \ref{sec:aux_results}), we have with probability at least $1-\delta,$
\begin{align*}&\left\|\frac{\sum_{j=1}^{i-1}\left(\dot\theta_{j}-\theta_*\right)\left(\dot\theta_{j}-\theta_*\right)^{\top}}{i-1}-\Sigma_*-\frac{\sigma^2\sum_{j=1}^{i-1}\E\left[V^{-1}_{j,\T_j}\right]}{i-1}\right\|_{op}\leq&\frac{16(\overline{\lambda}\lambda^2_e+16\sigma^2d)}{\lambda_e^2}\left(\sqrt{\frac{5d+2\log_e(2/\delta)}{i-1}}\vee\frac{5d+2\log_e(2/\delta)}{i-1}\right) \,.
\end{align*}
Similarly, with probability at least $1-\delta,$
\begin{align*}
\left\|\left(\theta_*-\bar{\theta}_i\right)\left(\theta_*-\bar{\theta}_i\right)^{\top}-\frac{1}{i-1}\Sigma_*-\frac{\sigma^2\sum_{j=1}^{i-1}\E\left[V^{-1}_{j,\T_j}\right]}{(i-1)^2}\right\|_{op}\leq\frac{16(\overline{\lambda}\lambda^2_e+16\sigma^2d)(5d+2\log_e(2/\delta))}{\lambda_e^2(i-1)} \,.
\end{align*}
Combining these with a union bound yields the result.
\Halmos
\endproof

The proof of Lemma \ref{lemma:deviation_cov} directly follows as shown below.

\proof{Proof of Lemma \ref{lemma:deviation_cov}}
When the event $\mathcal{A}$ holds, we can apply Lemma \ref{lemma:cov1} to Eq. \eqref{eq:cov2}. It is helpful to note that $(i-1)/(i-2)\leq2$ and $1/(i-1)\leq2/i$ for all $i\geq 3,$ and $5d+2\log_e(2/\delta)\leq10d\log_e(2/\delta)$ for all $\delta\in[0,2/e]$. By Lemma \ref{lemma:ini}, the event $\mathcal{A}$ does not hold with probability at most $2/(N^2T^2)$. Thus, a second union bound yields the result.
\Halmos
\endproof

\section{\texttt{Meta-DP++} Regret Analysis}\label{sec:theorem:cov_regret}

As discussed in Section \ref{sec:proof++}, we consider two cases; we first focus on the more substantive case where $N > N_1$.

We define a new clean event
\begin{align}
\J = \bigg\{ \nonumber~\forall i\geq N_1 \,, \qquad
&\T_i\leq \T_e\,, \quad \left\|\hat{\theta}_i-\theta_*\right\|\leq8\sqrt{\frac{(\sigma^2/\lambda_e+5\overline{\lambda})d\log_e(4dN^2T)}{i}} \,, \\
\nonumber&\left\|\hat{\Sigma}_i-\Sigma_*\right\|_{op}\leq\frac{128(\overline{\lambda}\lambda^2_e+16\sigma^2d)}{\lambda_e^2}\left(\sqrt{\frac{5d\log_e(2N^2T)}{i}}\vee\frac{5d\log_e(2N^2T)}{i}\right) \,,\\
\label{event:J}&\|\theta_i\|\leq S+5\sigma\sqrt{d\log_e(2N^2T)} \bigg\} \,,
\end{align}
which stipulates that for every epoch after the initial $N_1$ exploration epochs, (i) the event $\mathcal{A}$ defined in Eq. \eqref{event-A} holds, ensuring that the number of exploration periods per epoch is small, (ii) our estimated prior mean $\hat{\theta}_i$ is close to the unknown prior mean $\theta_*$, (iii) our estimated prior covariance $\hat{\Sigma}_i$ is close to the unknown prior covariance $\Sigma_*$, and (iv) the true parameter for epoch i $\theta_i \sim \N(\theta_*, \Sigma_*)$ is not too large in the $\ell_2$-norm. These events all hold with high probability based on Lemma \ref{lemma:ini}, \ref{lemma:deviation}, and \ref{lemma:deviation_cov}, and by the properties of multivariate Gaussians respectively; therefore the event $\J$ holds with high probability.

Denote the meta regret of epoch $i$ conditioned on the event $\mathcal{J}$ defined in Eq. \eqref{event:J} as $\regret_{N,T}(i) \mid \J$. As noted earlier, during the exploration periods $1 \leq t \leq \T_i$, the meta oracle and our \covmpdp~encounter the same covariates $\{x_{i,t} \}_{t=1}^T$ and offer the same prices; thus, by construction, they achieve the same expected revenue and the resulting meta regret is 0. Then, as in the proof of Theorem \ref{theorem:main_regret}, we can write
\begin{align}
\nonumber\regret_{N,T}(i) \mid \J=&\E_{\theta_i,\hat{\theta}_i,\chi^{\oracle}_i,\chi_i^{\mts}}\left[\rev\left(\theta_i,\theta^{\oracle}_{i,\T_i+1},\Sigma^{\oracle}_{i,\T_i+1},T-\T_i\right)-\rev\left(\theta_i,\theta^{\mts}_{i,\T_i+1},\Sigma^{\mts}_{i,\T_i+1},T-\T_i\right)\mid\J\right]\\
\nonumber=&\E_{\theta_i,\hat{\theta}_i,\chi_i^{\mts}}\left[\rev_*\left(\theta_i,T-\T_i\right)-\rev\left(\theta_i,\theta^{\mts}_{i,\T_i+1},\Sigma^{\mts}_{i,\T_i+1},T-\T_i\right)\mid \J\right]\\
\label{eq:cov4}&\quad-\E_{\theta_i,\hat{\theta}_i,\chi^{\oracle}_i}\left[\rev_*\left(\theta_i,T-\T_i\right)-\rev\left(\theta_i,\theta^{\oracle}_{i,\T_i+1},\Sigma^{\oracle}_{i,\T_i+1},T-\T_i\right)\mid \J\right] \,.
\end{align}
Appendix \ref{ssec:intermediate++} states two intermediate lemmas and Appendix \ref{ssec:cov-proof} provides the proof of Theorem \ref{theorem:cov_regret}.

\subsection{Intermediate Lemmas} \label{ssec:intermediate++}

First, as we did for the proof of Theorem \ref{theorem:main_regret}, we characterize the meta regret accrued by aligning the mean of the meta oracle's posterior $\theta^{\oracle}_{i,\T_i+1}$ and the mean of our  \covmpdp~$\theta^{\mts}_{i,\T_i+1}$.

\begin{lemma}\label{lemma:align} For an epoch $i\geq N_1,$
\begin{align*}
&\E_{\theta_i,\hat{\theta}_i,\chi_i^{\mts}}\left[\rev_*\left(\theta_i,T-\T_i\right)-\rev\left(\theta_i,\theta^{\mts}_{i,\T_i+1},\Sigma^{\mts}_{i,\T_i+1},T-\T_i\right)\mid\J\right]\\
\leq&\left(1+\frac{16c_3d^{3/2}\T_i\log^{3/2}_e(4dN^2T)}{\sqrt{i}}\right)\E_{\theta_i,\hat{\theta}_i,\chi_i^{\oracle}}\left[\rev_*\left(\theta_i,T-\T_i\right)-\rev\left(\theta_i,\theta^{\oracle}_{i,\T_i+1},\Sigma^{\mts}_{i,\T_i+1},T-\T_i\right)\middle|\J\right]+O\left(\frac{1}{N}\right).
\end{align*}
\end{lemma}

\proof{Proof of Lemma \ref{lemma:align}}
By the posterior update rule of Bayesian linear regression \citep{Bishop06}, we have
\begin{align*}
\theta^{\oracle}_{i,\T_i+1}=&\left(\Sigma_*^{-1}+\sigma\sum_{t=1}^{\T_i}m_{i,t}m_{i,t}^{\top}\right)^{-1}\left(\Sigma_*^{-1}\theta_*+\sigma\sum_{t=1}^{\T_i}m_{i,t}m^{\top}_{i,t}\theta_i+\sigma\sum_{t=1}^{\T_i}m_{i,t}\vep^{\oracle}_{i,t}\right) \,,\\
\theta^{\mts}_{i,\T_i+1}=&\left(\left(\hat{\Sigma}^{w}_i\right)^{-1}+\sigma\sum_{t=1}^{\T_i}m_{i,t}m_{i,t}^{\top}\right)^{-1}\left(\left(\hat{\Sigma}^{w}_i\right)^{-1}\hat{\theta}_i+\sigma\sum_{t=1}^{\T_i}m_{i,t}m^{\top}_{i,t}\theta_i+\sigma\sum_{t=1}^{\T_i}m_{i,t}\vep^{\mts}_{i,t}\right) \,.
\end{align*}
Denoting $M_i=\begin{pmatrix}
	m_{i,1}&\ldots&m_{i,\T_i}
\end{pmatrix}\in\R^{2d\times \T_i},$ we observe that prior alignment is achieved with $\theta^{\mts}_{i,\T_i+1}=\theta^{\oracle}_{i,\T_i+1}$ when the following holds:
\begin{align}\label{eq:cov5}
\chi^{\oracle}_i-\chi^{\mts}_i=\underbrace{\frac{1}{\sigma}(M_i^{\top}M_i)^{-1}\left[\left(\hat{\Sigma}^{w}_i\right)^{-1}\hat{\theta}_i-\Sigma^{-1}_*\theta_*+\left(\Sigma^{-1}_*-\left(\hat{\Sigma}^{w}_i\right)^{-1}\right)\left(\left(\hat{\Sigma}^{w}_i\right)^{-1}\hat{\theta}_i+\sigma M_iM^{\top}_i\theta_i+M_i\chi_i^{\mts}\right)\right]}_{\Delta_n} \,.
\end{align}
We denote the RHS of the above equation as $\Delta_n$ for ease of exposition. While this expression is more complicated than Eq. \eqref{eq:main_key}, it still induces a mapping between $\chi^{\oracle}_i$ and $\chi^{\mts}_i$. We then proceed similarly to the proof of Lemma \ref{lemma:main2}. We start by expanding
\begin{align*}
&\E_{\chi_i^{\mts}}\left[\rev_*\left(\theta_i,T-\T_i\right)-\rev\left(\theta_i,\theta^{\mts}_{i,\T_i+1},\Sigma^{\mts}_{i,\T_i+1},T-\T_i\right)\middle|\J\right]\\
=&\int_{\chi_i^{\mts}}\frac{\exp\left(-\left\|\chi_i^{\mts}\right\|^2/2\sigma^2\right)}{(2\pi\sigma^2)^{\T_i/2}}\left(\rev_*\left(\theta_i,T-\T_i\right)-\rev\left(\theta_i,\theta^{\mts}_{i,\T_i+1},\Sigma^{\mts}_{i,\T_i+1},T-\T_i\right)\right)d\chi_i^{\mts}|\J.
\end{align*}
Given a realization of $\chi^{\mts}_{i},$ we denote $\chi^{\oracle}_i(\chi^{\mts}_{i})$ (with some abuse of notation) as the corresponding realization of $\chi^{\oracle}_i$ that satisfies Eq. \eqref{eq:cov5}. It is easy to see that this is a unique one-to-one mapping. We then perform a change of measure (similar to Eq. \eqref{eq:main1}) to continue:
\begin{align}
\nonumber&\int_{\chi_i^{\mts}}\frac{\exp\left(-\left\|\chi_i^{\mts}\right\|^2/2\sigma^2\right)}{\exp\left(-\left\|\chi_i^{\oracle}(\chi_i^{\mts})\right\|^2/2\sigma^2\right)}\frac{\exp\left(-\left\|\chi_i^{\oracle}(\chi_i^{\mts})\right\|^2/2\sigma^2\right)}{(2\pi\sigma^2)^{\T_i/2}}\\
\nonumber&\times\left(\rev_*\left(\theta_i,T-\T_i\right)-\rev\left(\theta_i,\theta^{\mts}_{i,\T_i+1},\Sigma^{\mts}_{i,\T_i+1},T-\T_i\right)d\chi_i^{\mts}\right)|\J\\
\nonumber\leq&\max_{\left\|\chi_i^{\mts}\right\|\leq4\sigma\sqrt{\T_i\log_e(2NT)}}\exp\left(\frac{\left\|\chi_i^{\oracle}(\chi_i^{\mts})\right\|^2-\left\|\chi_i^{\mts}\right\|^2}{2\sigma^2}\right)\E_{\chi_i^{\oracle}}\left[\rev_*\left(\theta_i,T-\T_i\right)-\rev\left(\theta_i,\theta^{\oracle}_{i,\T_i+1},\Sigma^{\mts}_{i,\T_i+1},T-\T_i\right)\middle|\J\right]\\
\nonumber&+\E_{\chi_i^{\mts}}\left[\rev_*\left(\theta_i,T-\T_i\right)-\rev\left(\theta_i,\theta^{\mts}_{i,\T_i+1},\Sigma^{\mts}_{i,\T_i+1},T-\T_i\right)\middle|\J,\left\|\chi_i^{\mts}\right\|\geq4\sigma\sqrt{\T_i\log_e(2NT)}\right]\\
\nonumber&\quad\times\Pr\left(\left\|\chi_i^{\mts}\right\|\geq4\sigma\sqrt{\T_i\log_e(2NT)}\right)\\
\nonumber\leq&\max_{\left\|\chi_i^{\mts}\right\|\leq4\sigma\sqrt{\T_i\log_e(2NT)}}\exp\left(\frac{\left\|\chi_i^{\oracle}(\chi_i^{\mts})\right\|^2-\left\|\chi_i^{\mts}\right\|^2}{2\sigma^2}\right)\E_{\chi_i^{\oracle}}\left[\rev_*\left(\theta_i,T-\T_i\right)-\rev\left(\theta_i,\theta^{\oracle}_{i,\T_i+1},\Sigma^{\mts}_{i,\T_i+1},T-\T_i\right)\middle|\J\right]\\
\label{eq:cov6}&+\frac{\sqrt{\kappa+S^2}p_{\max}x_{\max}\sqrt{1+p^2_{\max}}}{N} \,,
\end{align}
where the last step follows from Eqs. \eqref{eq:main5} and \eqref{eq:main6}. Thus, we have expressed the true regret of our \covmpdp~as the sum of a term that is proportional to the true regret of a policy that is aligned with the meta oracle (\ie, it employs the prior $\N(\theta^{\mts}_{i,\T_i+1},\Sigma^{\mts}_{i,\T_i+1})$), and an additional term that is small (\ie, scales as $1/N$).

We now characterize the coefficient of the first term in Eq. \eqref{eq:cov6}:
\begin{align}
\nonumber&\max_{\left\|\chi_i^{\mts}\right\|\leq4\sigma\sqrt{\T_i\log_e(2NT)}}\exp\left(\frac{\left\|\chi_i^{\oracle}(\chi_i^{\mts})\right\|^2-\left\|\chi_i^{\mts}\right\|^2}{2\sigma^2}\right)\\
\nonumber&= \max_{\left\|\chi_i^{\mts}\right\|\leq4\sigma\sqrt{\T_i\log_e(2NT)}}\exp\left(\frac{\left\|\chi^{\mts}_{i}+\Delta_n\right\|^2-\left\|\chi_i^{\mts}\right\|^2}{2\sigma^2}\right)\\
\nonumber&= \max_{\left\|\chi_i^{\mts}\right\|\leq4\sigma\sqrt{\T_i\log_e(2NT)}}\exp\left(\frac{\left(\chi_i^{\mts}\right)^{\top}\Delta_n}{\sigma^2}+\frac{\|\Delta_n\|^2}{2\sigma^2}\right)\\
\nonumber&\leq \max_{\left\|\chi_i^{\mts}\right\|\leq4\sigma\sqrt{\T_i\log_e(2NT)}}\exp\left(\frac{\left\|\chi_i^{\mts}\right\|\|\Delta_n\|}{\sigma^2}+\frac{\|\Delta_n\|^2}{2\sigma^2}\right)\\
\label{eq:cov7}&= \max_{\left\|\chi_i^{\mts}\right\|\leq4\sqrt{\T_i\log_e(2NT)}}\exp\left(\frac{4\sqrt{t_e\log_e(2NT)}\|\Delta_n\|}{\sigma}+\frac{\|\Delta_n\|^2}{2\sigma^2}\right) \,.
\end{align}
To continue, we must characterize $\|\Delta_n\|$. Applying the triangle inequality, we have that
\begin{align}
\label{eq:cov8}\|\Delta_n\|\leq&\frac{1}{\sigma\lambda_e}\left\|\left(\hat{\Sigma}^{w}_i\right)^{-1}\hat{\theta}_i-\Sigma^{-1}_*\theta_*\right\|+\frac{1}{\sigma\lambda_e}\left\|\left(\Sigma^{-1}_*-\left(\hat{\Sigma}^{w}_i\right)^{-1}\right)\left(\left(\hat{\Sigma}^{w}_i\right)^{-1}\hat{\theta}_i+\sigma M_iM^{\top}_i\theta_i+M_i\chi_i^{\mts}\right)\right\| \,.
\end{align}
The first term of Eq. \eqref{eq:cov8} satisfies
\begin{align}
\nonumber&\frac{1}{\sigma\lambda_e}\left\|\left(\hat{\Sigma}^{w}_i\right)^{-1}\hat{\theta}_i-\Sigma^{-1}_*\theta_*\right\|\\
\nonumber&=\frac{1}{\sigma\lambda_e}\left\|\Sigma^{-1}_*\left(\hat{\theta}_i-\theta_*\right)+\left(\left(\hat{\Sigma}^{w}_i\right)^{-1}-\Sigma^{-1}_*\right)\left(\hat{\theta}_i-\theta_*\right)+\left(\left(\hat{\Sigma}^{w}_i\right)^{-1}-\Sigma^{-1}_*\right)\theta_*\right\|\\
\nonumber&\leq \frac{1}{\sigma\lambda_e}\left\|\Sigma^{-1}_*\left(\hat{\theta}_i-\theta_*\right)\right\|+\frac{1}{\sigma\lambda_e}\left\|\left(\left(\hat{\Sigma}^{w}_i\right)^{-1}-\Sigma^{-1}_*\right)\left(\hat{\theta}_i-\theta_*\right)\right\|+\frac{1}{\sigma\lambda_e}\left\|\left(\left(\hat{\Sigma}^{w}_i\right)^{-1}-\Sigma^{-1}_*\right)\theta_*\right\|\\
\label{eq:cov9}&\leq 8\sqrt{\frac{(\sigma^2/\lambda_e+5\overline{\lambda})d\log_e(4dN^2T)}{\sigma^2\lambda_e^2i}}\left(\frac{1}{\underline{\lambda}}+\left\|\left(\hat{\Sigma}^{w}_i\right)^{-1}-\Sigma^{-1}_*\right\|_{op}\right)+\frac{S}{\sigma\lambda_e}\left\|\left(\hat{\Sigma}^{w}_i\right)^{-1}-\Sigma^{-1}_*\right\|_{op}.
\end{align}
Next, the second term of Eq. \eqref{eq:cov8} satisfies
\begin{align}
\nonumber & \frac{1}{\sigma\lambda_e}\left\|\left(\Sigma^{-1}_*-\left(\hat{\Sigma}^{w}_i\right)^{-1}\right)\left(\left(\hat{\Sigma}^{w}_i\right)^{-1}\hat{\theta}_i+\sigma M_iM^{\top}_i\theta_i+M_i\chi_i^{\mts}\right)\right\|\\
\nonumber&\leq \frac{\left\|\Sigma_*^{-1}-\left(\hat{\Sigma}^{w}_i\right)^{-1}\right\|_{op}}{\sigma\lambda_e}\left(\left\|\left(\hat{\Sigma}^{w}_i\right)^{-1}\hat{\theta}_i\right\|+\left\|\sigma M_iM^{\top}_i\theta_i\right\|+\left\|M_i\chi_i^{\mts}\right\|\right)\\
\nonumber&\leq \frac{\left\|\Sigma_*^{-1}-\left(\hat{\Sigma}^{w}_i\right)^{-1}\right\|_{op}}{\sigma\lambda_e}\left(\left\|\left(\hat{\Sigma}^{w}_i\right)^{-1}\right\|_{op}\left(S+1\right)+\sigma \T_ix^2_{\max}(p^2_{\max}+p^4_{\max})+4\sigma p_{\max}x_{\max}\sqrt{\T_i(1+p^2_{\max})\log_e(2NT)}\right)\\
\label{eq:cov10}& \leq \frac{\left\|\Sigma_*^{-1}-\left(\hat{\Sigma}^{w}_i\right)^{-1}\right\|_{op}}{\sigma\lambda_e}\left(\left\|\Sigma_*^{-1}\right\|_{op}\left(S+1\right)+\sigma \T_ix^2_{\max}(p^2_{\max}+p^4_{\max})+4\sigma p_{\max}x_{\max}\sqrt{\T_i(1+p^2_{\max})\log_e(2NT)}\right)\\
\label{eq:cov11}&\leq \frac{8 p_{\max}x_{\max}\sqrt{\T_i(1+p^2_{\max})\log_e(2NT)}\left\|\Sigma_*^{-1}-\left(\hat{\Sigma}^{w}_i\right)^{-1}\right\|_{op}}{\lambda_e} \,,
\end{align}
where Eq. \eqref{eq:cov10} follows from the fact that $\|\hat{\Sigma}^w_i\|_{op}\geq\|\Sigma_*\|_{op}$ (on the event $\J$) and because both matrices are positive semi-definite (since they are covariance matrices). Applying Lemma \ref{lemma:op_prod}, we can simplify the term
\begin{align}
\nonumber\left\|\Sigma_*^{-1}-\left(\hat{\Sigma}^{w}_i\right)^{-1}\right\|_{op}&= \left\|\left(\hat{\Sigma}^{w}_i\right)^{-1}(\hat{\Sigma}^w_i-\Sigma_*)\Sigma^{-1}_*\right\|_{op}\\
\nonumber&\leq \left\|\left(\hat{\Sigma}^{w}_i\right)^{-1}\right\|_{op}\left\|\hat{\Sigma}^w_i-\Sigma_*\right\|_{op}\left\|\Sigma^{-1}_*\right\|_{op} \\
\label{eq:cov12}&\leq \frac{256(\overline{\lambda}\lambda^2_e+16\sigma^2d)}{\lambda_e^2\underline{\lambda}^2}\sqrt{\frac{5d\log_e(2N^2T)}{i}}.
\end{align}
Combining Eqs. \eqref{eq:cov8}--\eqref{eq:cov12}, we have
\begin{align*}
\|\Delta_n\|\leq c_3\sigma d\sqrt{\frac{d\T_i\log_e(4dN^2T)\log_e(2N^2T)}{i}} \,.
\end{align*}
Substituting this expression into Eq. \eqref{eq:cov7}, we can bound the coefficient
\begin{align*}
\max_{\left\|\chi_i^{\mts}\right\|\leq4\sigma\sqrt{\T_i\log_e(2NT)}}\exp\left(\frac{\left\|\chi_i^{\oracle}(\chi_i^{\mts})\right\|^2-\left\|\chi_i^{\mts}\right\|^2}{2\sigma^2}\right) &\leq \exp\left(8c_3d\T_i\log_e(2N^2T)\sqrt{\frac{d\log_e(4dN^2T)}{i}}\right)\\
&\leq 1+16c_3d\T_i\log^4_e(4dN^2T)\sqrt{\frac{d}{i}} \,,
\end{align*}
where we used Lemma \ref{lemma:exp_algebra} in the last step. Substituting into Eq. \eqref{eq:cov6} yields the result.
\Halmos
\endproof

We will use Lemma \ref{lemma:align} in the proof of Theorem \ref{theorem:cov_regret} to characterize the meta regret from prior alignment. The next lemma will help us characterize the remaining meta regret due to the difference in the covariance matrices post-alignment.

\begin{lemma}\label{lemma:cov_regret1}
When the event $\J$ holds, we can write
\[ \prod_{t=\T_i+1}^T\max_{\theta:\left\|\theta-\theta^{\oracle}_{i,t}\right\|\leq C}\frac{d\N\left(\theta^{\oracle}_{i,t},\Sigma'^{\mts}_{i,t}\right)}{d\N\left(\theta^{\oracle}_{i,t},\Sigma^{\oracle}_{i,t}\right)} ~\leq~ 1+\frac{2c_4d^{5/2}T\log^{3/2}_e(2N^2T)}{\sqrt{i}} ~\leq~ 3 \,. \]
\end{lemma}

\proof{Proof of Lemma \ref{lemma:cov_regret1}} By the definition of the multivariate normal distribution, we have
\begin{align*}
&\max_{\theta:\left\|\theta-\theta^{\oracle}_{i,t}\right\|\leq C}\frac{d\N\left(\theta^{\oracle}_{i,t},\Sigma'^{\mts}_{i,t}\right)}{d\N\left(\theta^{\oracle}_{i,t},\Sigma^{\oracle}_{i,t}\right)}\\
&= \sqrt{\frac{\det\left(\Sigma_{i,t}^{\oracle}\right)}{\det\left(\Sigma'^{\mts}_{i,t}\right)}}\max_{\theta:\left\|\theta-\theta^{\oracle}_{i,t}\right\|\leq C}\exp\left(\frac{\left(\theta-\theta^{\oracle}_{i,t}\right)^{\top}\left(\Sigma^{\oracle}_{i,t}\right)^{-1}\left(\theta-\theta^{\oracle}_{i,t}\right)}{2}-\frac{\left(\theta-\theta^{\oracle}_{i,t}\right)^{\top}\left(\Sigma'^{\mts}_{i,t}\right)^{-1}\left(\theta-\theta^{\oracle}_{i,t}\right)}{2}\right)\\
&= \sqrt{\frac{\det(\Sigma_{i,t}^{\oracle})}{\det(\Sigma'^{\mts}_{i,t})}}\max_{\theta:\left\|\theta-\theta'^{\mts}_{i,t}\right\|\leq C}\exp\left(\frac{\left(\theta-\theta^{\oracle}_{i,t}\right)^{\top}\left(\Sigma_*^{-1}-\left(\hat{\Sigma}^w_i\right)^{-1}\right)\left(\theta-\theta^{\oracle}_{i,t}\right)}{2}\right)\\
&\leq \sqrt{\frac{\det\left(\left(\hat{\Sigma}^w_i\right)^{-1}+\sum_{\tau=1}^{t-1}w_{i,\tau}w^{\top}_{i,\tau}\right)}{\det\left(\Sigma^{-1}_*+\sum_{\tau=1}^{t-1}w_{i,\tau}w^{\top}_{i,\tau}\right)}}\exp\left(\frac{C^2\left\|\Sigma_*^{-1}-\left(\hat{\Sigma}^w_i\right)^{-1}\right\|_{op}}{2}\right)\\
&\leq \sqrt{\frac{\det\left(\left(\hat{\Sigma}^w_i\right)^{-1}+\sum_{\tau=1}^{t-1}w_{i,\tau}w^{\top}_{i,\tau}\right)}{\det\left(\Sigma^{-1}_*+\sum_{\tau=1}^{t-1}w_{i,\tau}w^{\top}_{i,\tau}\right)}}\exp\left(\frac{128C^2(\overline{\lambda}\lambda^2_e+16\sigma^2d)}{\lambda_e^2\underline{\lambda}^2}\sqrt{\frac{5d\log_e(2N^2T)}{i}}\right) \,,
\end{align*}
where we have used Eq. \eqref{eq:cov12} in the last step. Since our estimated covariance matrix is widened, we know that on the event $\J$, $\Sigma^{-1}_*-\left(\hat{\Sigma}^w_i\right)^{-1}=\Sigma^{-1}_*\left(\hat{\Sigma}^w_i-\Sigma_*\right)\left(\hat{\Sigma}^w_i\right)^{-1}$ is positive semi-definite, and thus it is evident that $\left(\Sigma^{-1}_*+\sum_{\tau=1}^{t-1}w_{i,\tau}w^{\top}_{i,\tau}\right)-\left(\left(\hat{\Sigma}^w_i\right)^{-1}+\sum_{\tau=1}^{t-1}w_{i,\tau}w^{\top}_{i,\tau}\right)$ is also positive semi-definite. Therefore, conditioned on the clean event $\J$,
\[ \sqrt{\frac{\det\left(\left(\hat{\Sigma}^w_i\right)^{-1}+\sum_{\tau=1}^{t-1}w_{i,\tau}w^{\top}_{i,\tau}\right)}{\det\left(\Sigma^{-1}_*+\sum_{\tau=1}^{t-1}w_{i,\tau}w^{\top}_{i,\tau}\right)}} \leq1 \,.\]
The result follows directly.
\Halmos
\endproof

\subsection{Proof of Theorem \ref{theorem:cov_regret}} \label{ssec:cov-proof}

\proof{Proof of Theorem \ref{theorem:cov_regret}}
First, we consider the ``small N" regime, where $N\leq N_1$. In this case, our \covmpdp~simply executes $N$ instances prior-independent Thompson sampling. Then, an immediate consequence of Lemma \ref{lemma:ind-ts} is that the meta regret is bounded by $N \times \widetilde{O}\left(d^2T^{1/2}\right)=\widetilde{O}\left(d^3(NT)^{5/6}\right)$ because $N \leq N_1 =O(d^4T^2).$ Thus, the result already holds in this case.

We now turn our attention to the ``large N" regime, \ie, $N > N_1$. The meta regret can be decomposed as
\begin{align*}
\regret_{N,T}&=\left(\regret_{N,T}| \J\right)\Pr( \J)+\left(\regret_{N,T}|\neg \J\right)\Pr(\neg \J) \\
&\leq\left(\regret_{N,T}|\J\right)+\left(\regret_{N,T}|\neg \J\right)\Pr(\neg \J) \,.
\end{align*}
Recall that the event $\J$ is composed of four events, each of which hold with high probability. Applying a union bound over the epochs $i\geq N_1+1$ to Lemma \ref{lemma:ini}, Lemma \ref{lemma:deviation} (setting $\delta=1/(N^2T)$), Lemma \ref{lemma:deviation_cov} (with $\delta=1/(N^2T)$), and Eq. \eqref{eq:theta_dev} (with $u=5\sigma\sqrt{d\log_e(2N^2T)}$), we obtain that 
\[ \Pr\left(\J\right) \geq 1-4/(NT)-6/(NT^2)\geq1-10/(NT) \,.\]
Recall that when the event $\J$ is violated, the meta regret is $O(NT)$, so we can bound $\left(\regret_{N,T}|\neg \J\right)\Pr(\neg \J) = O(NT \times 1/(NT))=O(1)$. Therefore, the overall meta regret is simply
\begin{align}
\regret_{N,T}\leq\left(\regret_{N,T} \mid \J\right)+O(1) \,.
\end{align}

Thus, it suffices to bound $\regret_{N,T} \mid \J$. As described in Section \ref{sec:proof++}, we consider bounding the meta regret post-alignment ($t = \T_i+1, \cdots , T$), where our \covmpdp~follows the aligned posterior $\N(\theta^{\oracle}_{i,\T_i+1},\Sigma^{\mts}_{i,\T_i+1})$. Let $\N(\theta^{\oracle}_{i,t},\Sigma'^{\mts}_{i,t})$ denote the posterior of our \covmpdp~at time step $t$, if it begins with the prior $\N(\theta^{\oracle}_{i,\T_i+1},{\Sigma}^{\mts}_{i,\T_i+1})$ in time step $\T_i+1,$ but follows the randomness of the oracle. Then, we can write
\begin{align*}
&\E_{\theta_i,\hat{\theta}_i,}\left[\rev_*\left(\theta_i,T-\T_i\right)-\rev\left(\theta_i,\theta^{\oracle}_{i,\T_i+1},\Sigma^{\mts}_{i,\T_i+1},T-\T_i\right)\middle|\J\right]\\
&=\E_{\theta_i,\hat{\theta}_i,}\left[\int_{\theta}\rev_*\left(\theta_i,T-\T_i\right)-\rev\left(\theta_i,\theta,0,1\right)-\rev\left(\theta_i,\theta^{\mts}_{i,\T_i+2},\Sigma^{\mts}_{i,\T_i+2},T-\T_i-1\right)d\N(\theta^{\oracle}_{i,\T_i+1},\Sigma^{\mts}_{i,\T_i+1})\middle|\J\right]\\
&= \E_{\theta_i,\hat{\theta}_i,}\left[\int_{\theta:\|\theta\|\leq C}\rev_*\left(\theta_i,T-\T_i\right)-\rev\left(\theta_i,\theta,0,1\right)-\rev\left(\theta_i,\theta^{\mts}_{i,\T_i+2},\Sigma^{\mts}_{i,\T_i+2},T-\T_i-1\right)d\N(\theta^{\oracle}_{i,\T_i+1},\Sigma^{\mts}_{i,\T_i+1})\middle|\J\right]\\
&\quad+\E_{\theta_i,\hat{\theta}_i,}\left[\int_{\theta:\|\theta\|> C}\rev_*\left(\theta_i,T-\T_i\right)-\rev\left(\theta_i,\theta,0,1\right)-\rev\left(\theta_i,\theta^{\mts}_{i,\T_i+2},\Sigma^{\mts}_{i,\T_i+2},T-\T_i-1\right)d\N(\theta^{\oracle}_{i,\T_i+1},\Sigma^{\mts}_{i,\T_i+1})\middle|E\right]\\
& \leq \E_{\theta_i,\hat{\theta}_i,}\left[\max_{\theta:\left\|\theta-\theta'^{\mts}_{i,t}\right\|\leq C}\frac{d\N(\theta^{\oracle}_{i,\T_i+1},\Sigma'^{\mts}_{i,\T_i+1})}{d\N(\theta^{\oracle}_{i,\T_i+1},\Sigma^{\oracle}_{i,\T_i+1})}\left(\rev_*\left(\theta_i,1\right)-\rev\left(\theta_i,\theta^{\oracle}_{i,\T_i+1},\Sigma^{\oracle}_{i,\T_i+1},1\right)\right)\middle|\J\right]\\
&\quad+ \E_{\theta_i,\hat{\theta}_i,}\left[\max_{\theta:\left\|\theta-\theta^{\oracle}_{i,\T_i+1}\right\|\leq C}\frac{d\N(\theta^{\oracle}_{i,\T_i+1},\Sigma'^{\mts}_{i,\T_i+1})}{d\N(\theta^{\oracle}_{i,\T_i+1},\Sigma^{\oracle}_{i,\T_i+1})}\left(\rev_*\left(\theta_i,T-\T_i-1\right)-\rev\left(\theta_i,\theta^{\oracle}_{i,\T_i+2},\Sigma'^{\mts}_{i,\T_i+2},T-\T_i-1\right)\right)\middle|\J\right]\\
&\quad +\E_{\theta_i,\hat{\theta}_i,}\left[\int_{\theta:\left\|\theta-\theta^{\oracle}_{i,\T_i+1}\right\|> C}\rev_*\left(\theta_i,T-\T_i\right)d\N(\theta^{\oracle}_{i,\T_i+1},\Sigma^{\mts}_{i,\T_i+1})\middle|\J\right],
\end{align*}
where $C=5\sigma\sqrt{d\log_e(NT)}.$ Inductively, we have
\begin{align}
\nonumber&\E_{\theta_i,\hat{\theta}_i,\chi_i^{\oracle}}\left[\rev_*\left(\theta_i,T-\T_i\right)-\rev\left(\theta_i,\theta^{\oracle}_{i,\T_i+1},\Sigma^{\mts}_{i,\T_i+1},T-\T_i\right)\middle|\J\right]\\
\nonumber&\leq \E_{\theta_i,\hat{\theta}_i,}\left[\prod_{t=\T_i+1}^T\max_{\theta:\left\|\theta-\theta^{\oracle}_{i,t}\right\|\leq C}\frac{d\N(\theta^{\oracle}_{i,t},\Sigma'^{\mts}_{i,t})}{d\N(\theta^{\oracle}_{i,t},\Sigma^{\oracle}_{i,t})}\left(\rev_*\left(\theta_i,T-\T_i\right)-\rev\left(\theta_i,\theta^{\oracle}_{i,\T_i+1},\Sigma^{\oracle}_{i,\T_i+1},T-\T_i\right)\right)\middle|\J\right]\\
&\label{eq:cov13}\quad+\sum_{t=\T_i+1}^T\E_{\theta_i,\hat{\theta}_i,}\left[\prod_{t=\T_i+2}^T\max_{\theta:\left\|\theta-\theta^{\oracle}_{i,t}\right\|\leq C}\frac{d\N(\theta^{\oracle}_{i,t},\Sigma^{\mts}_{i,t})}{d\N(\theta^{\oracle}_{i,t},\Sigma^{\oracle}_{i,t})}\int_{\theta:\|\theta\|> C}\rev_*\left(\theta_i,T-t\right)d\N(\theta^{\oracle}_{i,t},\Sigma'^{\mts}_{i,t})\middle|\J\right].
\end{align}
Applying Lemma \ref{lemma:cov_regret1}, we can bound Eq. \eqref{eq:cov13} as
\begin{align}
\nonumber&\E_{\theta_i,\hat{\theta}_i,\chi_i^{\oracle}}\left[\rev_*\left(\theta_i,T-\T_i\right)-\rev\left(\theta_i,\theta^{\oracle}_{i,t_e+1},\Sigma^{\mts}_{i,t_e+1},T-\T_i\right)\middle|\J\right]\\
\nonumber\leq&\left(1+\frac{2c_4d^{5/2}T\log^{3/2}_e(2N^2T)}{\sqrt{i}}\right)\E_{\theta_i,\hat{\theta}_i,}\left[\rev_*\left(\theta_i,T-\T_i\right)-\rev\left(\theta_i,\theta^{\oracle}_{i,\T_i+1},\Sigma^{\oracle}_{i,\T_i+1},T-\T_i\right)\middle|\J\right]\\
\nonumber&+\sum_{t=\T_i+1}^T\E_{\theta_i,\hat{\theta}_i,}\left[3\int_{\theta:\|\theta\|> C}\rev_*\left(\theta_i,T-t\right)d\N(\theta^{\oracle}_{i,t},\Sigma'^{\mts}_{i,t})\middle|\J\right]\\
\nonumber=&\left(1+\frac{2c_4d^{5/2}T\log^{3/2}_e(2N^2T)}{\sqrt{i}}\right)\E_{\theta_i,\hat{\theta}_i,}\left[\rev_*\left(\theta_i,T-\T_i\right)-\rev\left(\theta_i,\theta^{\oracle}_{i,\T_i+1},\Sigma^{\oracle}_{i,\T_i+1},T-\T_i\right)\middle|\J\right]+O\left(\frac{1}{N}\right) \,,
\end{align}
where we used Eq. \eqref{eq:theta_dev} in the last step. Thus, we have expressed the post-alignment meta regret as the sum of a term that is proportional to the true regret of the meta oracle and a negligibly small term. We can now apply Lemma \ref{lemma:align} to further include the meta regret accrued from our prior alignment step to obtain
\begin{align*}
&\E_{\theta_i,\hat{\theta}_i,\chi_i^{\mts}}\left[\rev_*\left(\theta_i,T-\T_i\right)-\rev\left(\theta_i,\theta^{\mts}_{i,\T_i+1},\Sigma^{\mts}_{i,\T_i+1},T-\T_i\right)\middle|\J\right]\\
& \leq \left(1+\frac{16c_3d^{3/2}\T_i\log^{3/2}_e(4dN^2T)}{\sqrt{i}}\right)\left(1+\frac{2c_4d^{5/2}T\log^{3/2}_e(2N^2T)}{\sqrt{i}}\right)\\
&\quad\times\E_{\theta_i,\hat{\theta}_i,}\left[\rev_*\left(\theta_i,T-\T_i\right)-\rev\left(\theta_i,\theta^{\oracle}_{i,\T_i+1},\Sigma^{\oracle}_{i,\T_i+1},T-\T_i\right)\middle|E\right]+O\left(\frac{1}{N}\right) \,.
\end{align*}
As desired, this establishes that the coefficient of our first term decays to $1$ as $i$ grows large. Thus, our meta regret from the first term approaches $0$ for large $i$, and all other terms are clearly negligible.

Noting that $N > N_1 = \tilde{O}(d^4T^2)$ in the ``large N" regime, we can upper bound the meta regret as
\begin{align*}
&\sum_{i=N_1+1}^N\left[\left(1+\frac{16c_3d^{3/2}\T_i\log^{3/2}_e(4dN^2T)}{\sqrt{i}}\right)\left(1+\frac{2c_4d^{5/2}T\log^{3/2}_e(2N^2T)}{\sqrt{i}}\right)-1\right]\\
&\quad\times\E_{\theta_i,\hat{\theta}_i,}\left[\rev_*\left(\theta_i,T-\T_i\right)-\rev\left(\theta_i,\theta^{\oracle}_{i,\T_i+1},\Sigma^{\oracle}_{i,\T_i+1},T-\T_i\right)\middle|\J\right]+O\left(\frac{1}{N}\right)\\
&= \tilde{O}\left(\sum_{i=N_1+1}^N\frac{d^{4}T^{\frac{3}{2}}}{\sqrt{i}}\right) ~=~ \tilde{O}\left(d^4N^{\frac{1}{2}}T^{\frac{3}{2}}\right) ~=~ \tilde{O}\left(d^2(NT)^{\frac{5}{6}}\right) \,.
\end{align*}
\Halmos
\endproof

\section{Extension to Multiple Products with Substitution Effects}\label{sec:multiple_prod}

Thus far, we have considered the setting where the seller offers a single product in each epoch. In practice, there may be many products offered simultaneously in an epoch, and there may be substitution effects across these products (within a single epoch) that must be additionally modeled. We now show that our transfer learning approach extends straightforwardly to this setting.

\subsection{Formulation}

We extend our single-product epoch formulation from Section \ref{sec:problem} to a multi-product epoch formulation, where $K$ products are offered in each epoch. To capture substitution effects \textit{within} an epoch, we will employ an epoch-level joint demand model across all $K$ products. Our demand model is an extension of the multi-product demand model proposed by \cite{keskin2014dynamic}, with the addition of (exogenous, product-specific and customer-specific) features. The seller will now choose a price \textit{vector} (one for each product), observe a demand vector, and estimate the demand function jointly across all products given the price/demand data.

As before, in epoch $i \in [N]$ at time $t \in [T]$, the seller observes a random feature vector $ \vx_{i,t} \in \R^d$, which is sampled i.i.d. from a known distribution $\cP^{mp}_i$. She then chooses a price vector $p^{mp}_{i,t}=\begin{pmatrix}
p^{mp}_{i,t,1}&\ldots p^{mp}_{i,t,K}
\end{pmatrix}^{\top} \in \R^K$, where $p^{mp}_{i,t,k}$ is the chosen price for product $k\in[K]$ in time $t$ and epoch $i$. Recall that, owing to practical constraints, we assume that the allowable price range is bounded across periods and products, \ie, $p^{mp}_{i,t} \in [p_{\min}, 1]^{K}$ and that $0< p_{\min} < 1$.\footnote{Note that we have set $p_{\max}=1$; this is done WLOG since we can always normalize our parameters appropriately.} The seller then observes the resulting induced demand for product $k\in[K]$,
\[ D^{mp}_{i,t,k}(p^{mp}_{i,t},\vx_{i,t}) = \langle\alpha^{mp}_{i,k}, \vx_{i,t}\rangle + \sum_{j=1}^{K}p^{mp}_{i,t,j} \langle\beta^{mp}_{i,k,j}, \vx_{i,t}\rangle + \vep^{mp}_{i,t,k} \,,\]
where $\alpha^{mp}_{i,k} \in \R^{d}$ and $\beta^{mp}_{i,k,j} \in \R^{d}$ are unknown fixed constants throughout epoch $i$, and $\vep^{mp}_{i,t,k}\sim\N(0,\sigma^2)$ is i.i.d. Gaussian noise with variance $\sigma^2$.

Observe that the demand for product $k$ now depends not only on the price of product $k$ but also on the prices of all other products in this epoch --- in particular, $\beta_{i,k,j}$ for $j\neq k$, captures the substitution effects between products $k$ and $j$ under feature vector $\vx_{i,t}$. For ease of notation, we collectively denote the demand vector
\begin{align}
D^{mp}_{i,t}(p^{mp}_{i,t},x_{i,t})=\begin{pmatrix}
D^{mp}_{i,t,1}(p^{mp}_{i,t},x_{i,t})&\ldots&D^{mp}_{i,t,K}(p^{mp}_{i,t},x_{i,t})
\end{pmatrix} \,.
\end{align}

\paragraph{Shared Structure:} For ease of notation, we additionally define the matrix
\[\theta^{mp}_i = \begin{pmatrix}
\alpha^{mp}_{i,1}&\ldots&\alpha^{mp}_{i,K}\\
\beta^{mp}_{i,1,1}&\ldots&\beta^{mp}_{i,K,1}\\
\vdots&\ldots&\vdots\\
\beta^{mp}_{i,1,K}&\ldots&\beta^{mp}_{i,K,K}
\end{pmatrix}\in \R^{(K+1)d\times K} \,, \]
where $\theta^{mp}_i$ is the unknown parameter matrix that must be learned within a given epoch in order for the seller to maximize her revenues over $T$ periods. When there is no shared structure between the $\{\theta^{mp}_i\}_{i=1}^N$, our problem reduces to $N$ independent dynamic pricing problems.

However, as discussed in the main paper, we may have some shared structure that can be related across products. We model the shared structure by positing that product demand parameters $\{\theta^{mp}_i\}_{i=1}^N$ are independent and identically distributed draws from a common unknown matrix normal distribution,\footnote{See, \eg, \cite{GuptaN99} for the definition and properties of a matrix normal distribution.} \ie, $\theta^{mp}_i\sim \MN(\theta^{mp}_*, \Sigma^{mp}_*,I_K)$ for each $i \in [N].$ (The third argument is $I_K$ because the noise terms are uncorrelated by assumption.)

\paragraph{Assumptions:} We impose the same assumptions made in Section \ref{sec:assumption}. However, since we are now learning $(K^2+K)d$ instead of just $2d$ parameters (in the single-product case), we may naturally expect that the constants to differ. Specifically, we take the constants in Assumption \ref{assumption:x_norm} to be $x_{\max}$ and $S^{mp}$; similarly, we take the constant in Assumption \ref{assumption:Sigma} to be $\overline{\lambda}^{mp}$ and $\underline{\lambda}^{mp}$ for the multi-product setting.

\paragraph{Meta Oracle:} As before, we define our meta oracle to be Thompson Sampling with a known prior. Here, our meta oracle is TS$\left(\MN\left(\theta^{mp}_*,\Sigma^{mp}_*,I_K\right),\lambda^{mp}_e\right),$ the Thompson sampling algorithm with prior $\MN\left(\theta^{mp}_*,\Sigma^{mp}_*,I_K\right)$ and an input parameter $\lambda^{mp}_e$. The description is formally given in Algorithm \ref{alg:oracle_multi} below. As before, we perform random price exploration for $\tilde{O}(1)$ time periods by offering initial prices
\begin{align}\label{eq:multiple_ini}
p^{(1)}=\begin{pmatrix}
p_{\min}\\p_{\min}\\p_{\min}\\\vdots\\p_{\min}
\end{pmatrix},\quad p^{(2)}=\begin{pmatrix}
1\\p_{\min}\\p_{\min}\\\vdots\\p_{\min}
\end{pmatrix},\quad p^{(3)}=\begin{pmatrix}
p_{\min}\\1\\p_{\min}\\\vdots\\p_{\min}
\end{pmatrix},\quad\ldots\quad p^{(K+1)}=\begin{pmatrix}
p_{\min}\\p_{\min}\\\vdots\\p_{\min}\\1
\end{pmatrix} \,.
\end{align}
The random exploration period ends once the minimum eigenvalue of the matrix
\[ \sum_{s=1}^t\begin{pmatrix}
x^{\top}_{i,s}&~p^{mp}_{i,s,1}x^{\top}_{i,s}&\ldots p^{mp}_{i,s,K}x^{\top}_{i,s}
\end{pmatrix}^{\top}\begin{pmatrix}
x^{\top}_{i,s}&~p^{mp}_{i,s,1}x^{\top}_{i,s}&\ldots p^{mp}_{i,s,K}x^{\top}_{i,s}
\end{pmatrix} \,, \]
exceeds $\lambda^{mp}_e.$ For each subsequent time step, the meta oracle (1) samples the unknown product demand parameters 
\[ \mathring{\theta}^{mp}_{i,t}= \begin{pmatrix}
\mathring{\alpha}^{mp}_{i,t,1}&\ldots&\mathring{\alpha}^{mp}_{i,t,K}\\
\mathring{\beta}^{mp}_{i,t,1,1}&\ldots&\mathring{\beta}^{mp}_{i,t,K,1}\\
\vdots&\ldots&\vdots\\
\mathring{\beta}^{mp}_{i,t,1,K}&\ldots&\mathring{\beta}^{mp}_{i,t,K,K}
\end{pmatrix} \,,\]
from the posterior $\N\left(\theta^{\oracle}_{i,t},I_K\otimes\Sigma^{\oracle}_{i,t}\right)$, and (2) solves and offers the resulting optimal price based on the demand function given by the sampled parameters
\begin{align}\label{eq:oracle_price}
p_{i,t}^{\oracle} = \argmax_{p\in\left[p_{\min},1\right]^K} ~ \sum_{k=1}^K\left[p_k \left( \left\langle\mathring{\alpha}_{i,t,k}, \vx_{i,t}\right\rangle+ \sum_{j=1}^Kp_j\cdot \left\langle\mathring{\beta}_{i,t,k,j},\vx_{i,t}\right\rangle\right)\right] \,.
\end{align}
Upon observing the actual realized demand $D_{i,t}\left(p_{i,t}^{\oracle},\vx_{i,t}\right)$, the algorithm computes the posterior $\MN\left(\theta^{\oracle}_{i,t+1},\Sigma^{\oracle}_{i,t+1},I_K\right)$ for round $t+1$ \citep{RossiAM05}. The same algorithm is applied independently to each epoch $i \in [N]$.

\begin{algorithm}[!ht]
	\SingleSpacedXI
	\caption{TS$(\MN\left(\theta^{mp}_*,\Sigma^{mp}_*,I_K\right),\lambda^{mp}_e):$ Thompson Sampling Algorithm}
	\label{alg:oracle_multi}
	\begin{algorithmic}[1]
		\State \textbf{Input:} The prior mean matrix $\theta^{mp}_*$ and covariance matrix $\Sigma^{mp}_*,$ the index $i$ of epoch, the length of each epoch $T,$ the noise parameter $\sigma,$ exploration parameter $\lambda_e.$
		\State \textbf{Initialization:} $t\leftarrow1,\left(\theta^{\oracle}_{i,t},\Sigma^{\oracle}_{i,t}\right)\leftarrow\left(\theta^{mp}_*,\Sigma^{mp}_*\right),$
		\While{$\lambda_{\min}\left(\sum_{s=1}^{t-1}\begin{pmatrix}
			x^{\top}_{i,s}&~p_{i,s,1}x^{\top}_{i,s}&\ldots p_{i,s,K}x^{\top}_{i,s}
			\end{pmatrix}^{\top}\begin{pmatrix}
			x^{\top}_{i,s}&~p_{i,s,1}x^{\top}_{i,s}&\ldots p_{i,s,K}x^{\top}_{i,s}
			\end{pmatrix}\right)\leq\lambda_e$}
		\State{Observe feature vector $\vx_{i,t},$ and offer price $p^{\oracle}_{i,t}\leftarrow p^{(t\mod~K)}$}
		\State{Observe demand $D_{i,t}\left(p^{\oracle}_{i,t},\vx_{i,t}\right),$ and compute the posterior $\MN\left(\theta^{\oracle}_{i,t+1},\Sigma^{\oracle}_{i,t+1},I_K\right).$}
		\State{$t\leftarrow t+1.$}
		\EndWhile
		\While{$t\leq T$}
		\State{Observe feature vector $\vx_{i,t}.$}
		\State{Sample parameter $\mathring{\theta}_{i,t}\sim\MN\left(\theta^{\oracle}_{i,t},\Sigma^{\oracle}_{i,t},I_K\right).$}
		\State{Offer $
			p_{i,t}^{\oracle}$ according to eq. \eqref{eq:oracle_price}.}
		\State{Observe demand $D_{i,t}\left(p^{\oracle}_{i,t},\vx_i\right),$ and compute the posterior $\MN\left(\theta^{\oracle}_{i,t+1},\Sigma^{\oracle}_{i,t+1},I_K\right).$}
			\State{$t\leftarrow t+1.$}
		\EndWhile
	\end{algorithmic}
\end{algorithm}
The following theorem bounds the Bayes regret of our meta oracle:

\begin{corollary}[Multi-Product meta oracle]
	\label{theorem:oracle_multi}
	The Bayes regret of Algorithm \ref{alg:oracle_multi} satisfies 
	\begin{align*}
	\textnormal{Bayes Regret}_{N,T}(\pi)=\tilde{O}\left(K^3d^{\frac{3}{2}}N\sqrt{T}\right) \,,
	\end{align*}
	when the prior over the product demand parameters is known.
\end{corollary}
Corollary \ref{theorem:oracle_multi} follows directly from Theorem \ref{theorem:oracle} in the single-product case. This is because, if a matrix $X$ follows the matrix Gaussian distribution $\MN(A,B,C),$ then $\text{vec}(X),$ (\ie, the vectorized version of $X$ that concatenates each column of $X$ to form a vector), follows the multivariate Gaussian distribution $\N(A,C\otimes B)$ \citep{GuptaN99}. Thus, since we still maintain a linear demand model, the only mathematical change is that the unknown parameter has dimension $(K^2+K)d$ instead of $2d$. Thus, the same result applies by replacing the $d$ in Theorem \ref{theorem:oracle} with $(K^2+K)d$.

\subsection{Multi-Product \texttt{Meta-DP} Algorithm}

The multi-product \mpdp~is presented in Algorithm \ref{alg:main_multi}. We first define some additional notation, and then describe the algorithm in detail. 

\paragraph{Additional Notation:} Analogous to our previous notation, we use
\[ m^{mp}_{i,t}=\begin{pmatrix}x_{i,t}\\p^{mp}_{i,t,1}x_{i,t}\\\vdots\\p^{mp}_{i,t,K}x_{i,t}
	\end{pmatrix} \,, \]
to denote the price and feature information and $V^{mp}_{i,t}=\sum_{\tau=1}^tm^{mp}_{i,t}\left(m^{mp}_{i,t}\right)^{\top}$ to denote the Fisher information matrix of round $t$ in epoch $i$ for all $i\in[N]$ and $t\in[T].$

\paragraph{Algorithm Description:} The first $N^{mp}_0$ epochs are treated as exploration epochs, where we define
\begin{align}
\label{eq:N_0_multi}
N^{mp}_0=(c_2^{mp})^2d^2(K^2+K)^2\log_e(4d(K^2+K)N^2T)\log_e(2NT)\lambda^{mp}_e,
\end{align} 
and the constant $c^{mp}_2$ is defined as
	\begin{align*} 
	c^{mp}_2=\frac{32\sqrt{2x^2_{\max}(\sigma^2\left(\lambda_e^{mp}\right)^{-1}+5\overline{\lambda}^{mp})}}{\lambda^{mp}_e\underline{\lambda}^{mp}\sigma^2} \,.
	\end{align*}
As before, the \mpdp~proceeds differently for earlier exploration epochs and later epochs:
\begin{enumerate}
	\item \textbf{Epoch} $\mathbf{i \leq N^{mp}_0}:$ The \mpdp~runs the prior-independent Thompson sampling algorithm \citep{agrawal2013thompson,AbeilleL17} TS$(\MN(0,\Psi^{mp}\cdot I_{(K+1)d},I_K),\lambda_e),$ where $$\Psi^{mp}=\sigma\sqrt{2d\log_e(T(1+2x^2_{\max}T))}+\sqrt{20\overline{\lambda}^{mp}d\log_e(2T)}.$$
	\item \textbf{Epoch} $\mathbf{i > N^{mp}_0}:$ the \mpdp~computes the ordinary least square (OLS) estimate of the parameter vector $\theta_i$ for each of the past epochs; then, it averages these OLS estimates to arrive at an estimate $\hat{\theta}^{mp}_i$ of the prior mean $\theta_*,$ \ie,
		\begin{align}
		\label{eq:ts_mean_update_multi}
		\hat{\theta}^{mp}_{i}=\frac{\sum_{j=1}^{i-1}\left(V^{mp}_{j,T}\right)^{-1}\left(\sum_{t=1}^{T}m^{mp}_{j,t}D^{mp}_{j,t}(p^{mp}_{j,t},x_{j,t})\right)}{i-1}.
		\end{align}
		Then, the \mpdp~runs the Thompson sampling algorithm (see Algorithm \ref{alg:oracle_multi}) with the estimated prior $\MN(\hat{\theta}^{mp}_i,\Sigma^{mp}_*,I_K).$
\end{enumerate}

\begin{algorithm}[h]
	\SingleSpacedXI
	\caption{Meta-Personalized Dynamic Pricing Algorithm}
	\label{alg:main_multi}
	\begin{algorithmic}[1]
		\State \textbf{Input:} The prior covariance matrix $\Sigma^{mp}_*,$ the total number of epochs $N,$ the length of each epoch $T,$ the subgaussian parameter $\sigma,$ and the set of feasible prices $[p_{\min},1].$
		\State \textbf{Initialization:} $N_0\text{ as defined in eq. (\ref{eq:N_0_multi})}.$
		\For{each epoch $i=1,\ldots,N$}
		\If{$i\leq N_0$}
		\State{Run TS$(\MN(0,\Psi^{mp}\cdot I_{(K+1)d},I_K),\lambda^{mp}_e).$}
		\Else
		\State{Update $\hat{\theta}^{mp}_{i}$ according to eq. (\ref{eq:ts_mean_update_multi}), and run TS$\left(\MN\left(\hat{\theta}^{mp}_i,\Sigma^{mp}_*,I_K\right),\lambda^{mp}_e\right).$}
		\EndIf
		\EndFor
	\end{algorithmic}
\end{algorithm}

We now translate our previous upper bound on the meta regret of the single-product \mpdp~to the multi-product setting.
\begin{corollary}[Multi-Product \texttt{Meta-DP}]
	\label{theorem:main_regret_multi}
	The meta regret of multi-product \texttt{Meta-DP} satisfies
	\begin{align*}
	\regret_{N,T}(\mpdp)=\tilde{O}\left(K^4d^2(NT)^{\frac{1}{2}}\right).
	\end{align*}
\end{corollary}
Corollary \ref{theorem:main_regret_multi} is again an immediate consequence of Theorem \ref{theorem:main_regret}. Again, this is because, if a matrix $X$ follows the matrix Gaussian distribution $\MN(A,B,C),$ then $\text{vec}(X),$ (\ie, the vectorized version of $X$ that concatenates each column of $X$ to form a vector), follows the multivariate Gaussian distribution $\N(A,C\otimes B)$ \citep{GuptaN99}. In other words, we can map the multi-product prior $\MN\left(\theta^{mp}_*,\Sigma^{mp}_*,I_K\right)$ to the same form as a single-product prior $\N\left(\theta^{mp}_*,I_k\otimes\Sigma^{mp}_*\right)$, by taking the unknown prior mean to be the vectorized $\text{vec}(\theta^{mp}_*)$ and the prior covariance to be $\begin{pmatrix}1&p^{mp}_{i,t,1}&\ldots&p^{mp}_{i,t,K}\end{pmatrix}^{\top}\otimes x_{i,t}\otimes\bm{1}_K$ ($\bm{1}_K$ is the $K\times 1$ column vector with all entries equal to 1).
Thus, since we still maintain a linear demand model, the only mathematical change is that the unknown parameter has dimension $(K^2+K)d$ instead of $2d$. Thus, the same result applies by replacing the $d$ in Theorem \ref{theorem:main_regret} with $(K^2+K)d$.

\subsection{Multi-Product \covmpdp}

The multi-product \covmpdp~is presented in Algorithm \ref{alg:extension_multi}. We first define some additional notation, and then describe the algorithm in detail.

\paragraph{Algorithm Description:} The first $N^{mp}_1$ epochs are treated as exploration epochs, where we define
\begin{align}
	N^{mp}_1=\max\{&N_0,~32(c^{mp}_3)^2d^3(K^2+K)^3\T^2_e\log^3_e(2d(K^2+K)N^2T), \nonumber \\
	&~(c^{mp}_4)^2d^{4}(K^2+K)^4T^2\log^{3}_e(2N^2T)\} \label{eq:N_1_multi} \,,
\end{align}
and the constants are defined as
{\footnotesize
\begin{align*}
	c^{mp}_3=& \frac{16\sqrt{\sigma^2(\lambda^{mp}_e)^{-1}+5\overline{\lambda}^{mp}}}{\sigma\lambda^{mp}_e\underline{\lambda}^{mp}}+\frac{256(\overline{\lambda}^{mp}(\lambda^{mp}_e)^2+16\sigma^2)}{\left(\lambda_e^{mp}\underline{\lambda}^{mp}\right)^2}\left(\frac{8\sqrt{2} x_{\max}}{\lambda^{mp}_e}+\frac{S^{mp}}{\sigma\lambda^{mp}_e}\right)\,, \quad c_4=\frac{10^4\sigma(\overline{\lambda}^{mp}(\lambda^{mp}_e)^2+16\sigma^2)}{\left(\lambda_e^{mp}\underline{\lambda}^{mp}\right)^2} \,.
\end{align*}}
As before, the \covmpdp~proceeds differently for earlier exploration epochs and later epochs: 
\begin{enumerate}
	\item \textbf{Epoch $\mathbf{i \leq N^{mp}_1}$:} the \covmpdp~runs the prior-independent Thompson sampling algorithm TS$(\MN(0,\Psi^{mp}\cdot I_{(K+1)d},I_K),\lambda_e),$ where 
	\[\Psi^{mp}=\sigma\sqrt{2d\log_e(T(1+2x^2_{\max}T))}+\sqrt{20\overline{\lambda}^{mp}d\log_e(2T)} \,.\]
	\item \textbf{Epoch $\mathbf{i > N^{mp}_1}$:} the \covmpdp~computes an estimator $\hat{\theta}^{mp}_i$ of the prior mean $\theta^{mp}_*$ using Eq. \eqref{eq:ts_mean_update_multi} (same as the multi-product \mpdp), and an estimator $\hat{\Sigma}^{mp}_i$ of the prior covariance $\Sigma^{mp}_*$ as
	\begin{align}\label{eq:ts_cov_update_multi}
		\hat{\Sigma}^{mp}_i=\frac{1}{i-2}\sum_{j=1}^{i-1}\left(\dot{\theta}^{mp}_{j}-\frac{\sum_{k=1}^{i-1}\dot{\theta}^{mp}_k}{i-1}\right)\left(\dot{\theta}^{mp}_{j}-\frac{\sum_{k=1}^{i-1}\dot{\theta}^{mp}_k}{i-1}\right)^{\top}-\frac{\sigma^2\sum_{j=1}^{i-1}\E\left[\left(V^{mp}_{j,\T_j}\right)^{-1}\right]}{i-1} \,,
	\end{align}
where, following the single-product \covmpdp, we define
\begin{align}
	\nonumber\dot{\theta}^{mp}_i=\left(V^{mp}_{i,\T_i}\right)^{-1}\left(\sum_{t=1}^{\T_i}D^{mp}_{i,t}(p^{mp}_{i,t},x_{i,t})m^{mp}_{i,t}\right).
\end{align}
The widened posterior covariance is thus
	\begin{align}\label{eq:widening_multi}
		\hat{\Sigma}^{mp,w}_i ~=~ \hat{\Sigma}_i+\frac{128(\overline{\lambda}^{mp}(\lambda^{mp}_e)^2+8\sigma^2dK(K+1))}{(\lambda_e^{mp})^2}\sqrt{\frac{5dK(K+1)\log_e(2N^2T)}{i}}\cdot I_{K(K+1)d} \,,
	\end{align}
	where $I_{K(K+1)d}$ is the $(K(K+1)d)$-dimensional identity matrix. 
	
	Then, the \covmpdp~runs the Thompson Sampling algorithm (see Algorithm \ref{alg:oracle_multi}) with the estimated prior $\MN\left(\hat{\theta}^{mp}_i,\hat{\Sigma}^{mp,w}_i,I_K\right)$.
\end{enumerate}

\begin{algorithm}[!ht]
	\SingleSpacedXI
	\caption{Meta-Dynamic Pricing++ Algorithm}
	\label{alg:extension_multi}
	\begin{algorithmic}[1]
		\State \textbf{Input:} The total number of products $N,$ the length of each epoch $T,$ the noise parameter $\sigma,$ and the set of feasible prices $[p_{\min},1].$
		\For{epoch $i=1,\ldots,N$}
		\If{$i\leq N^{mp}_1$}
		\State{Run TS$(\MN(0,\Psi^{mp}\cdot I_{(K+1)d},I_K),\lambda^{mp}_e).$}
		\Else
		\State{Update $\hat{\theta}^{mp}_{i}$ and $\hat{\Sigma}^{mp}_i$ according to Eqs. \eqref{eq:ts_mean_update_multi} and \eqref{eq:ts_cov_update_multi} respectively.}
		\State{Compute widened prior mean estimate $\hat{\Sigma}^{mp,w}_i$ according to Eq. \eqref{eq:widening_multi}.}
		\State{Run TS$\left(\MN\left(\hat{\theta}^{mp}_i,\hat\Sigma^{mp,w}_i,I_K\right),\lambda^{mp}_e\right).$}
		\EndIf
		\EndFor
	\end{algorithmic}
\end{algorithm}

We now translate our previous upper bound on the meta regret of the single-product \covmpdp~to the multi-product setting.
\begin{corollary}[Multi-Product \texttt{Meta-DP++}]
	\label{theorem:cov_regret_multi}
	The meta regret of multi-product \texttt{Meta-DP++} satisfies
	\begin{align*}
		\regret_{N,T}(\covmpdp) &= \tilde{O}\left(\min\left\{K^4d^2 N T^{\frac{1}{2}},~ K^8d^4 N^{\frac{1}{2}}T^{\frac{3}{2}} \right\}\right) = \tilde{O}\left(K^6d^3(NT)^{\frac{5}{6}}\right) \,.
	\end{align*}
\end{corollary}
Corollary \ref{theorem:cov_regret_multi} is again an immediate consequence of Theorem \ref{theorem:cov_regret}. The reasoning is exactly the same as for Corollary \ref{theorem:main_regret_multi}, so we omit it. Essentially, we can map the multi-product prior to the same form as a single-product prior, so that the only mathematical change is that the unknown parameter has dimension $(K^2+K)d$ instead of $2d$. Thus, the same result applies by replacing the $d$ in Theorem \ref{theorem:cov_regret} with $(K^2+K)d$.

\section{Additional Numerical Experiments} \label{app:add_numerics}

This section includes a number of additional numerical results, including comparison to a hierarchical Thompson Sampling heuristic (\ref{app:hierarchical}), examining the estimation error of the prior as a function of $N$ (\ref{sec:add_numerical_err}), as well as results under a revenue metric (\ref{sec:add_numerical}).

\subsection{Comparison to Hierarchical Thompson Sampling} \label{app:hierarchical}

As discussed in Section \ref{ssec:remarks}, an alternative heuristic to leverage shared structure is to use hierarchical Thompson Sampling, maintaining a posterior on the shared prior and updating it after each epoch. Instead of using just the point estimate of the prior mean as the \mpdp, hierarchical Thompson sampling maintains a posterior on the shared prior and updates it after each epoch. We now compare the \mpdp~to such a hierarchical approach with unknown prior mean $\theta_*$.

Specifically, for each epoch $i,$ hierarchical TS samples $\tilde{\theta}^{H}_i$ from the posterior $\N(\theta^H_{i},\Sigma^H_{i})$ and runs the Thompson sampling algorithm $TS(\N(\tilde{\theta}^H_i,\Sigma_*),\lambda_e)$ for epoch $i.$ Afterwards, like the \mpdp, it estimates $\theta_i$ via Eq. \eqref{eq:ts_mean_update} and updates the posterior of $\theta_*$ to $\N(\theta^H_{i+1},\Sigma^H_{i+1})$ using the standard Bayesian update rules \cite[see, \eg, Chapter 18 of][]{BolstadC16}, \ie,
\begin{align}\label{eq:hts}
\Sigma^H_{i+1}=\left[\left(\Sigma^H_{i}\right)^{-1}+\Sigma_*^{-1}\right]^{-1},\qquad 	\theta^H_{i+1}=\Sigma^H_{i+1}(\Sigma^H_{i})^{-1}\theta^H_{i}+\Sigma^H_{i+1}\Sigma_*^{-1}\hat{\theta}_i \,.
\end{align} 
Since we begin with a cold start, we follow \cite{agrawal2013thompson} and initialize the prior to $\theta^H_1=0$ and $\Sigma^H_1=\bar{\lambda}\sqrt{48d\log^2_e(T)}I_{2d}.$
The formal description is provided in Algorithm \ref{alg:hts}.
\begin{algorithm}[!ht]
	\SingleSpacedXI
	\caption{Hierarchical Thompson Sampling Algorithm}
	\label{alg:hts}
	\begin{algorithmic}[1]
		\State \textbf{Input:} The prior covariance matrix $\Sigma_*,$ the total number of epochs $N,$ the length of each epoch $T,$ the noise parameter $\sigma,$ and the set of feasible prices $[p_{\min},p_{\max}].$
		\State \textbf{Initialization:} $\theta^H_1=0$ and $\Sigma^H_1=\bar{\lambda}\sqrt{48d\log^2_e(T)}I_{2d}.$
		\For{each epoch $i=1,\ldots,N$}
		\State{Sample $\tilde{\theta}^{H}_i$ from the posterior $\N(\theta^H_{i},\Sigma^H_{i})$ and run TS$\left(\N\left(\tilde{\theta}^{H}_i,\Sigma_*\right),\lambda_e\right).$}
		\State{Update $\hat{\theta}_{i}$ according to Eq. \eqref{eq:ts_mean_update} and update $\theta^H_{i+1}$ and $\Sigma^H_{i+1}$ according to Eq. \eqref{eq:hts}.}
		\EndFor
	\end{algorithmic}
\end{algorithm}

\begin{figure}[!ht]
	\subfigure[$d=1$]{\label{fig:d1}\includegraphics[width=16cm,height=6cm]{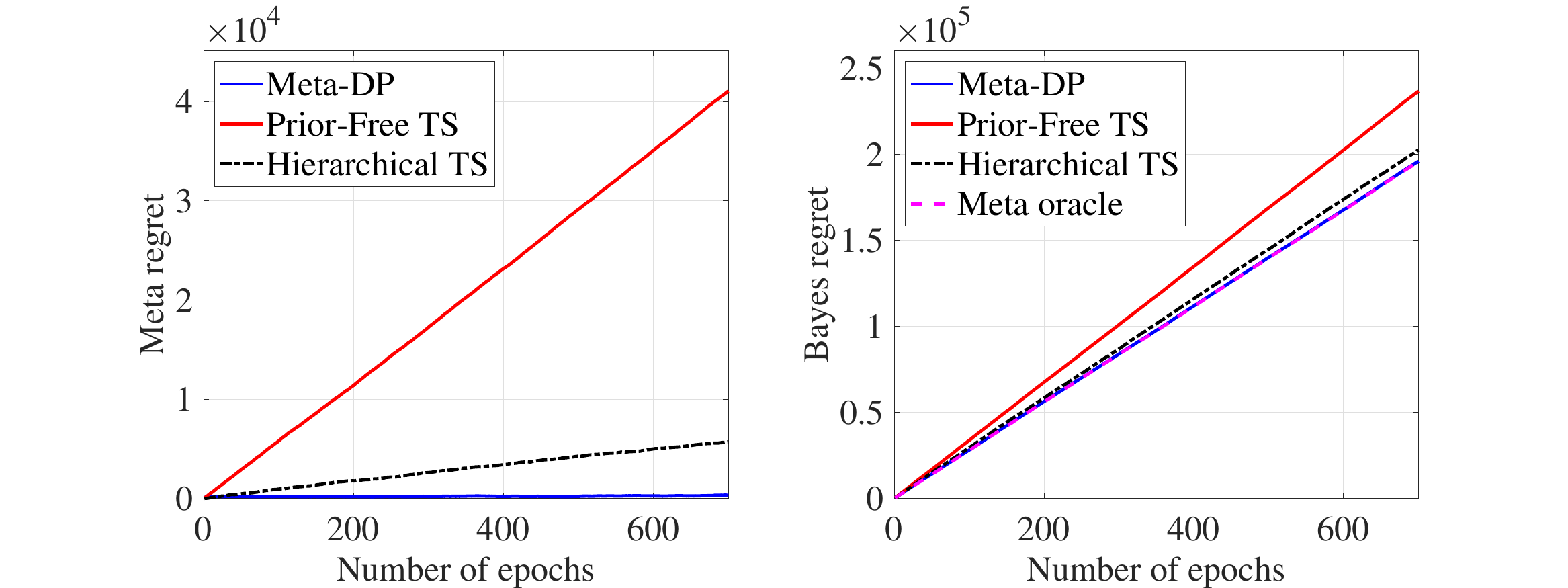}}
	\subfigure[$d=10$]{\label{fig:d10}\includegraphics[width=16cm,height=6cm]{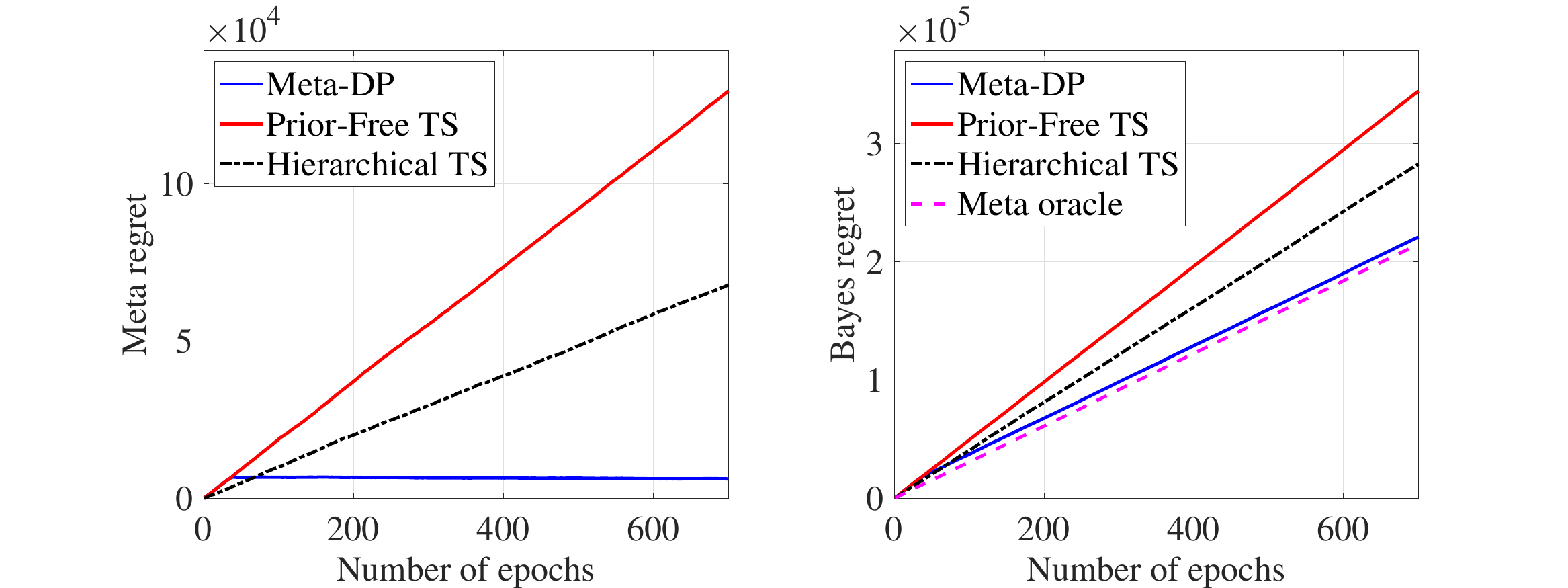}}
	\caption{Cumulative meta regret and Bayes regret for \texttt{Meta-DP}, prior-independent Thompson Sampling, and hierarchical Thompson sampling for feature dimension (a) $d =1$ and (b) $d=10$.}
	\label{fig:hier}
\end{figure}

Following the same setup described in Section \ref{sec:numerics}, Figure \ref{fig:hier} shows results analogous to Figure \ref{fig:d_var} on synthetic data for varying values of the feature dimension $d$. While the hierarchical algorithm significantly outperforms prior-independent Thompson Sampling by leveraging shared structure, we find that it still underperforms compared to the \mpdp~for moderate to large values of $N$. The latter result appears to stem from excessive exploration. In particular, while the \mpdp~uses the point estimate of the prior mean for Thompson Sampling in non-exploration epochs, the hierarchical Thompson sampling algorithm still samples from its posterior, inducing additional unnecessary exploration. Thus, the \mpdp~performs favorably in non-exploration epochs compared to hierarchical Thomspon Sampling.

\subsection{Estimation Error of the Prior}\label{sec:add_numerical_err}

The key ingredient to achieving low regret in the \mpdp~is successfully estimating the prior mean $\theta_*$. We now examine the estimation error $\| \hat{\theta}_i - \theta_*\|$ of the \mpdp~as a function of the number of epochs and various problem parameters.
Figure \ref{fig:err} presents results for varying values of (a) the feature dimension $d$, (b) the variance of the noise $\sigma$, (c) the magnitude of the prior mean $\|\theta_*\|$, and (d) the maximum eigenvalue of the prior covariance matrix $\bar{\lambda}$. We observe that the estimation error increases with the number of features and the noise (because we require more samples for convergence), the width of the prior (because there is more uncertainty), and the magnitude of $\|\theta_*\|$ (because it scales the size of the problem).

\begin{figure}[h]
	\centering
	\subfigure[Dimension of features]{\includegraphics[width=7.4cm,height=5.85cm]{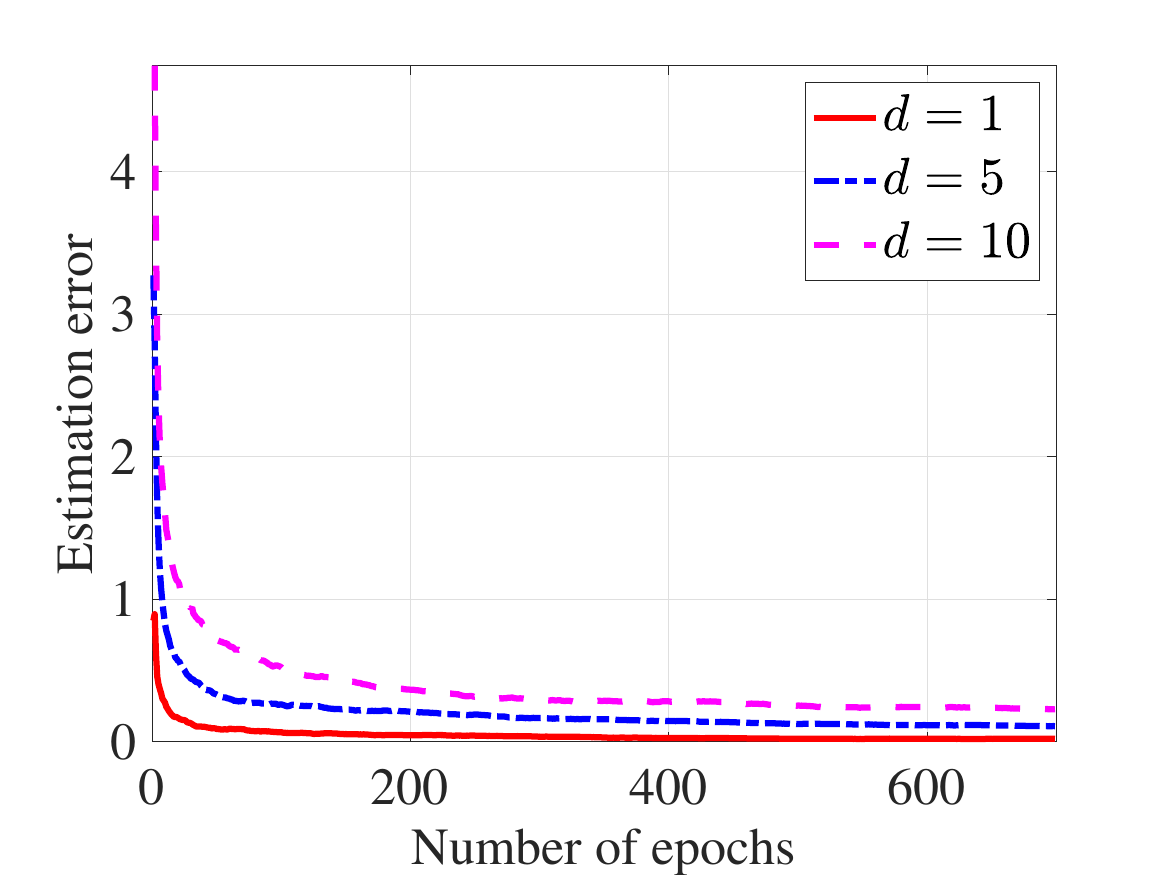}}
	\subfigure[Variance of noise terms]{\includegraphics[width=7.4cm,height=5.85cm]{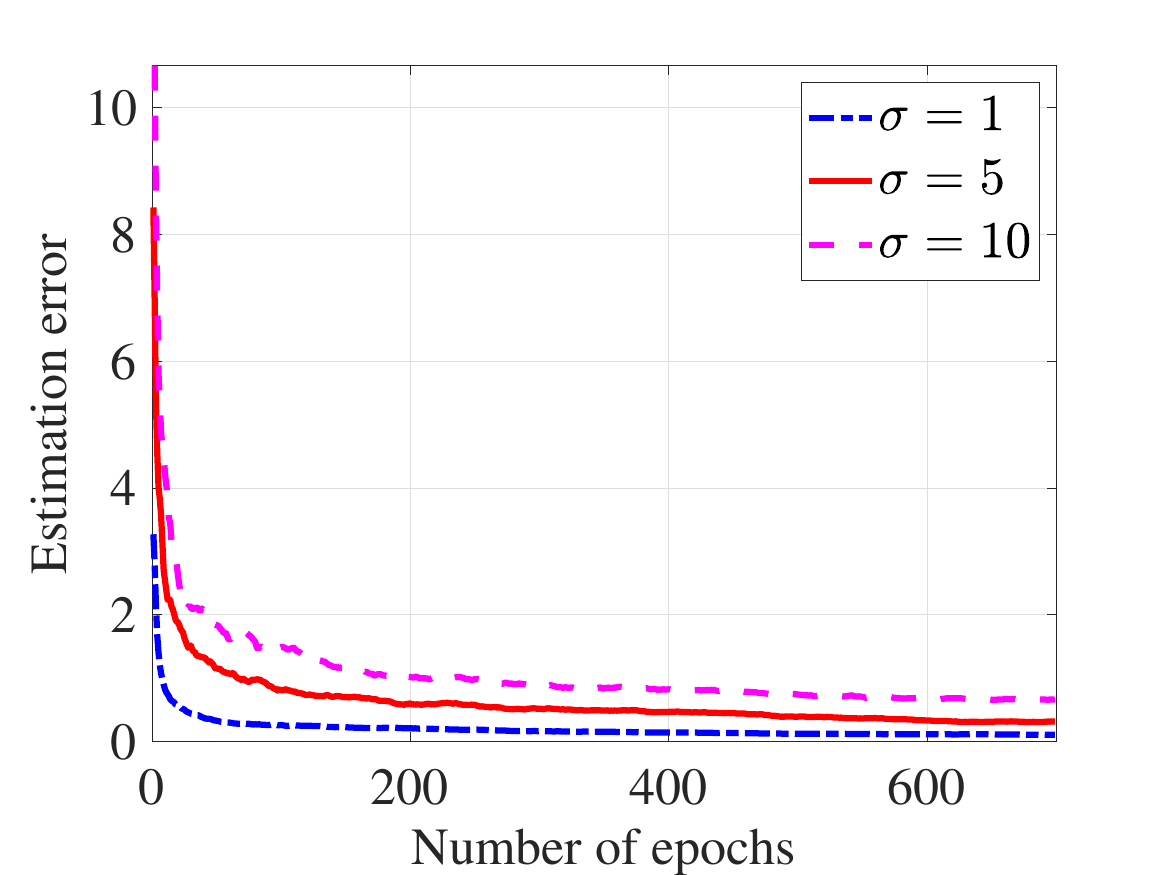}}
	\subfigure[Magnitude of prior mean]{\includegraphics[width=7.4cm,height=5.85cm]{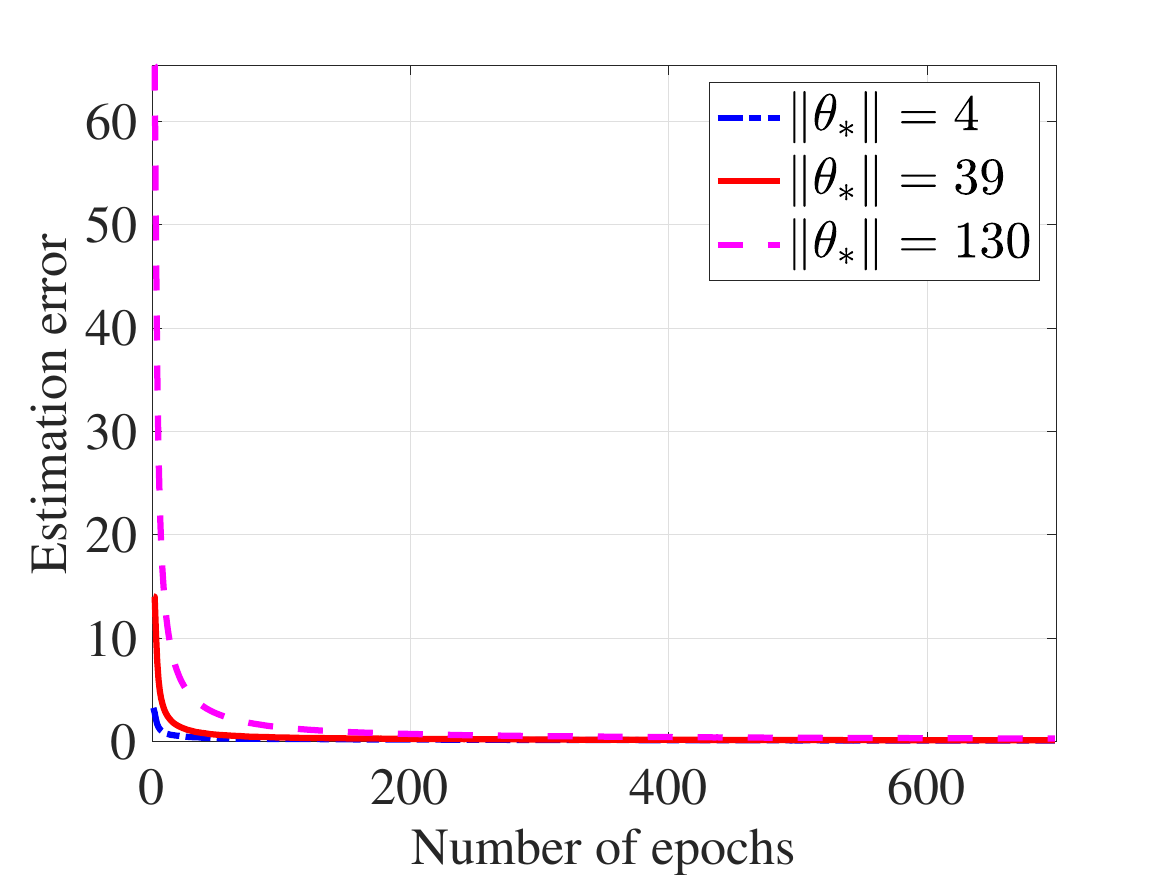}}
	\subfigure[Magnitude of prior coviarnace]{\includegraphics[width=7.4cm,height=5.85cm]{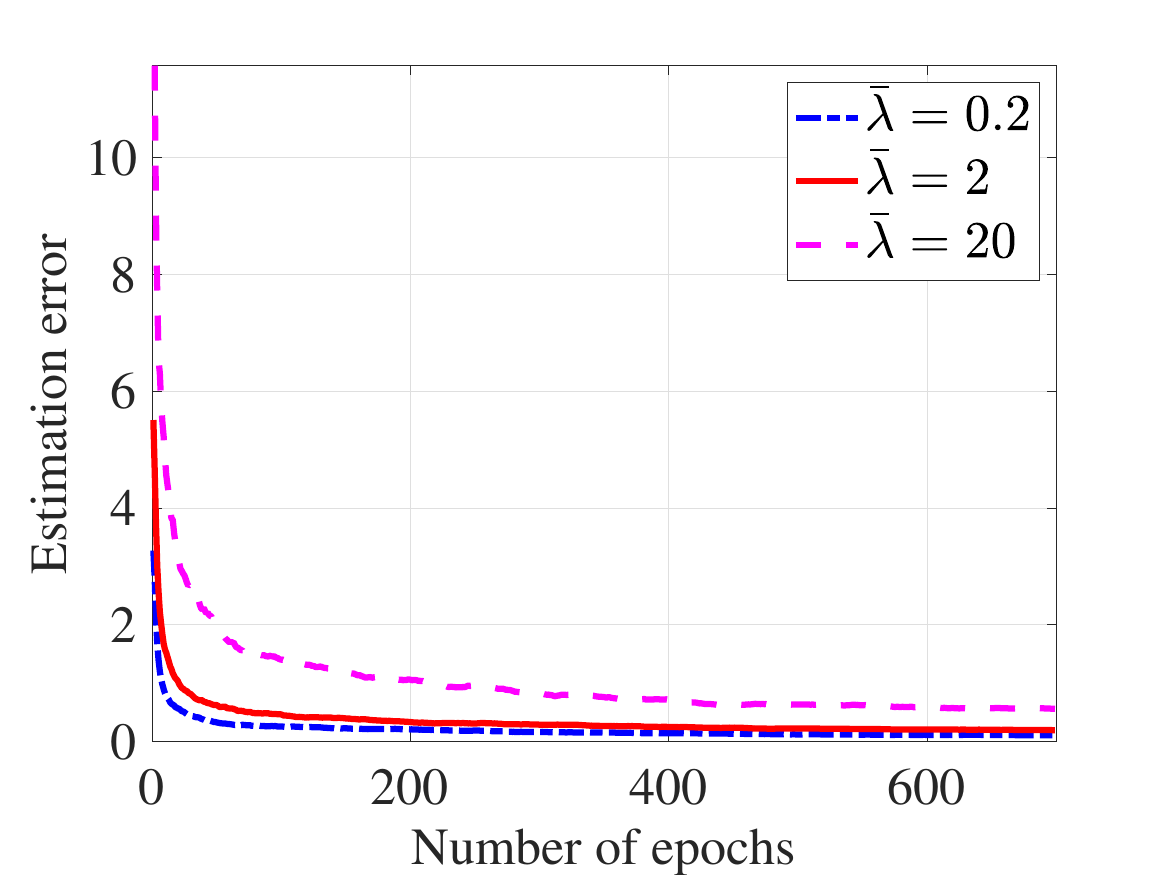}}
	\caption{Estimation errors in prior mean under different problem parameters.}
	\label{fig:err}
\end{figure}

\subsection{Results on Cumulative Revenue}\label{sec:add_numerical}

We now present representative results from the same experiments in Section \ref{sec:numerics}, but compare performance in terms of cumulative revenue. 
Figure \ref{fig:rev} shows results analogous to Figure \ref{fig:d5} for the \mpdp~and Figure \ref{fig:pp} for the \covmpdp~on synthetic data; Figure \ref{fig:auto_rev} shows results analogous to Figure \ref{fig:auto} on a real dataset on auto loans. Our qualitative insights remain the same as discussed in the main paper.
\begin{figure}[h]
	\centering
	\subfigure[Known $\Sigma_*$]{\includegraphics[width=7.4cm,height=5.85cm]{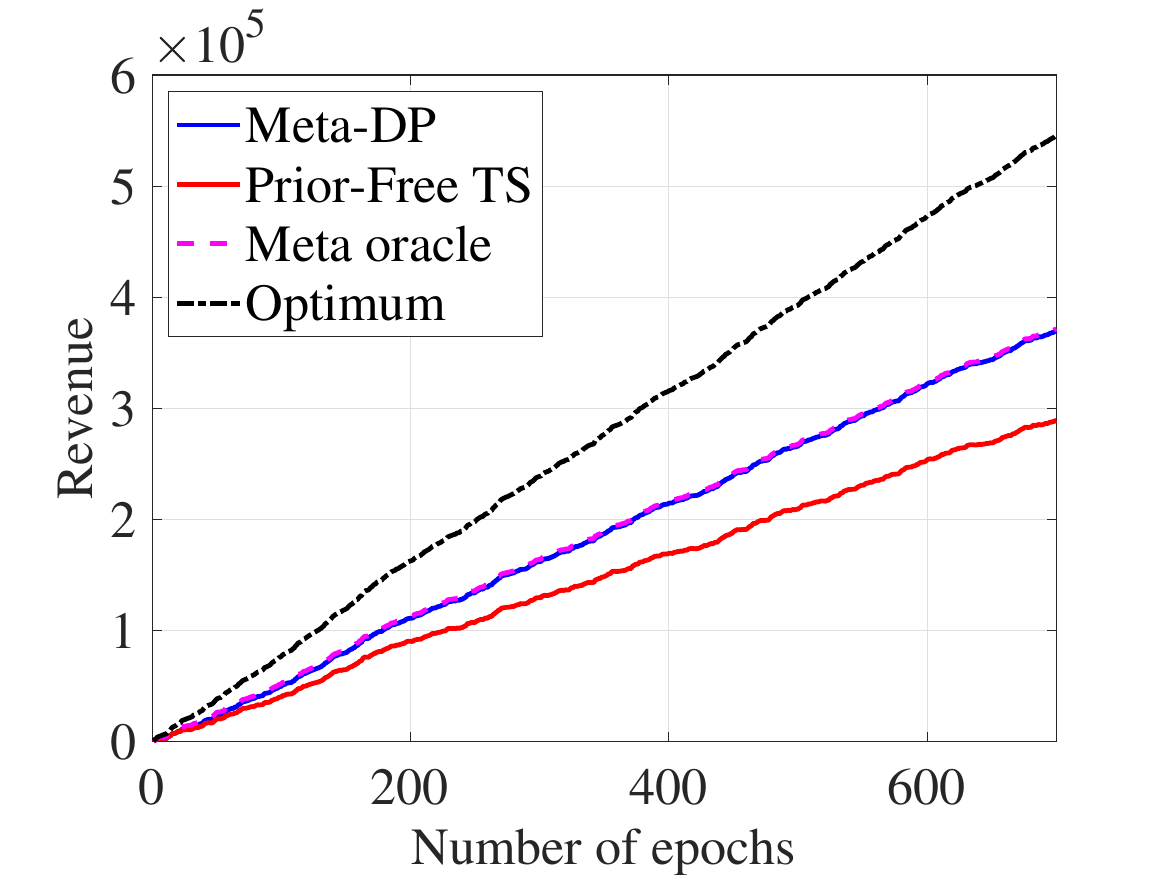}}
	\subfigure[Unknown $\Sigma_*$]{\includegraphics[width=7.4cm,height=5.85cm]{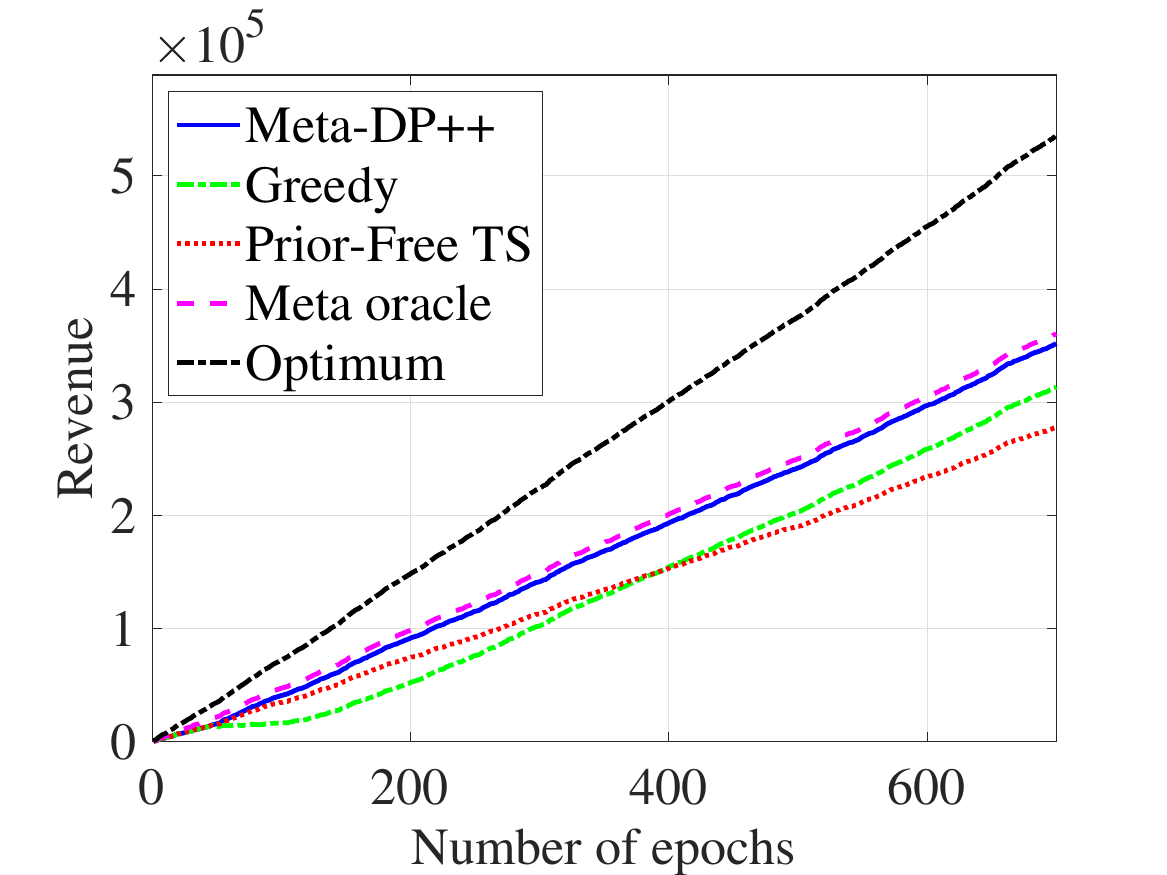}}
	\caption{Cumulative revenue for \texttt{Meta-DP}, \texttt{Meta-DP++}, and benchmark algorithms on synthetic data.}
	\label{fig:rev}
\end{figure}
\begin{figure}
	\centering
	\includegraphics[width=7.4cm,height=5.85cm]{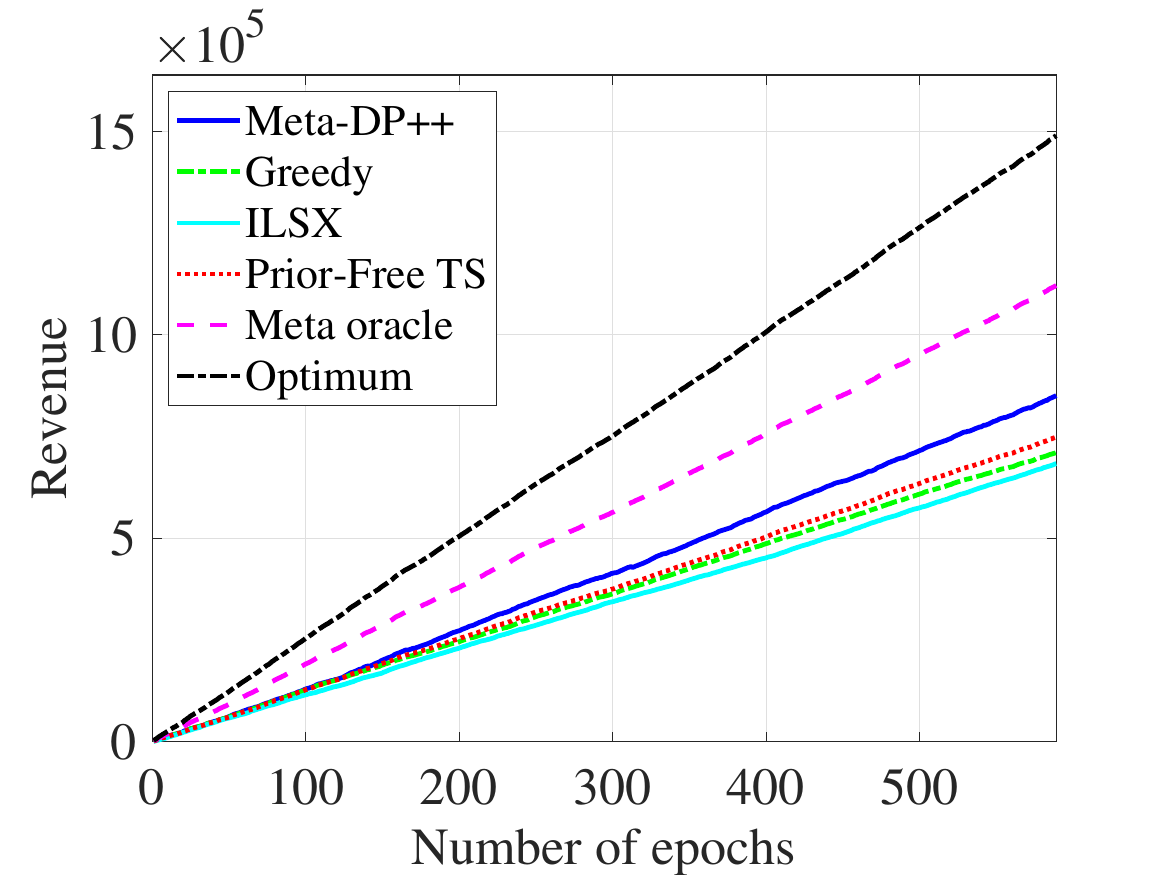}
	\caption{Cumulative revenue for \texttt{Meta-DP++}, and benchmark algorithms on real auto loan data.}
	\label{fig:auto_rev}
\end{figure}
\section{Auxiliary Results}
\label{sec:aux_results}

For completeness, we restate some well-known results from the literature.

The following lemma characterizes the Bayesian regret of Thompson sampling for the linear bandit.

\begin{lemma}[Proposition 3 of \cite{RVR14}]
\label{aux_thm1}
Fix positive constants $\sigma,c,$ and $c'.$ Denote the set of all possible parameters as $\Theta\in\R^d,$ the mean reward function as $f_\theta(a)=\langle\phi(a),\theta\rangle$ for some $\phi:\mathcal{A}\to\R,$ $\sup_{\rho\in\Theta}\|\rho\|\leq c,$ and $\sup_{a\in\mathcal{A}}\|\phi(a)\|\leq c',$ and for each $t,$ the noise term is $\sigma$-subgaussian, then the Bayesian regret of the Thompson sampling algorithm is $\widetilde{O}(d\sqrt{T}).$
\end{lemma}

The following lemma characterizes the eigenvalues of a matrix Kronecker product.
\begin{lemma}[Corollary 13.11 of \cite{Laub04}]\label{lemma:kronecker}
	Let $A$ be a real-valued matrix with singular values $\lambda_1\geq\ldots\geq\lambda_r>0,$ and let $B$ be a real-valued matrix with singular values $\lambda'_1\geq\ldots\geq\lambda'_s>0,$ then $A\otimes B$ has $r\cdot s$ singular values $\lambda_i\lambda'_j~(i\in[r]~j\in[s]).$
\end{lemma}

The following lemma upper bounds the covering number of a $d$-dimensional unit ball.
\begin{lemma}[\citealp{M19}]
	\label{aux_thm3}
	For the $d$-dimensional unit ball, its $\delta$ covering number is upper bounded by $d\log_e(1+2/\delta).$
\end{lemma}

The following lemma provides an upper bound for the quantity $\exp(1/a)$ when $a>1.$
\begin{lemma}
	\label{lemma:exp_algebra} For any number $a\in[0,1]$, $\exp\left(a\right)\leq1+2a$.
\end{lemma}
\begin{proof}{Proof of Lemma \ref{lemma:exp_algebra}.}
	We note that the function $f(a)=\exp(a)-1-2a$ is a convex function as 
	\begin{align}
	f''(a)=e^a>0,
	\end{align}
	as well as that $f(0)=1-1=0$ and $f(1)=e-3<0,$ so $f(a)\leq0$ for all $a\in[0,1].$ \halmos
\end{proof}

The following lemma makes a connection between the tail probability of a random variable and its moment generating function.
\begin{lemma}[Lemma 1.5 of \cite{RH18}]\label{lemma:mgf}
	For a random variable $X\in\R$ such that $\E[X]=0$ and for any $u>0,$
	$$\Pr\left(|X|>u\right)\leq2\exp\left(-\frac{u^2}{2\sigma^2}\right),$$
	we have for any $v\in\R,$
	$$\E[\exp(vX)]\leq\exp(4v^2\sigma^2).$$
\end{lemma}	
	
The following lemma provides a concentration inequality for estimating the empirical covariance matrix.
\begin{lemma}[Theorem 7.1 of \cite{R17} and Theorem 6.5 of \cite{M19}]\label{lemma:cov}
 Let $X_1,\ldots,X_n$ be $n$ i.i.d. copies of the random vector $X$ such that $\E[X]=0,\E[XX^{\top}]=\Sigma,$ and $X$ is $\sigma$-subgaussian vector. Then, the operator norm of the difference between the empirical covariance $\sum_{i=1}^nX_iX_i^{\top}/n$ and $\Sigma$ satisfies
	$$\Pr\left(\left\|\frac{\sum_{i=1}^nX_iX_i^{\top}}{n}-\Sigma\right\|_{op}\leq32\sigma^2\left(\sqrt{\frac{5d+2\log_e(2/\delta)}{n}}\vee\frac{5d+2\log_e(2/\delta)}{n}\right)\right)\geq 1-\delta$$
	for any $\delta\in[0,1].$
\end{lemma}

The following lemma shows that the operator norm of the product of two matrices is upper bounded by the product of the operator norms of those matrices.
\begin{lemma}\label{lemma:op_prod}
	For two matrices $A$ and $B,$ we have
	\begin{align*}
	\|AB\|_{op}\leq\|A\|_{op}\|B\|_{op}.
	\end{align*}
\end{lemma}
\begin{proof}{Proof of Lemma \ref{lemma:op_prod}.}
	The statement can be easily concluded as follows.\begin{align*}
	\|AB\|_{op}=\max_{x:\|x\|=1}\|ABx\|=&\max_{x:\|x\|=1}\frac{\|ABx\|}{\|Bx\|}\|Bx\|\\
	\leq&\max_{x:\|x\|=1}\frac{\|ABx\|}{\|Bx\|}\max_{y:\|y\|=1}\|By\|\\
	=&\max_{Bx:\|x\|=1}\frac{\|ABx/\|Bx\|\|}{\|Bx/\|Bx\|\|}\max_{y:\|y\|=1}\|By\|\\
	=&\|A\|_{op}\|B\|_{op}.
	\end{align*}
	\halmos
\end{proof}

The following lemma compares the determinants of two positive semi-definite matrices.
\begin{lemma}\label{lemma:det}
	For two symmetric positive semi-definite matrices $A$ and $B,$ if $A-B$ is positive semi-definite, then $\det(A)\geq\det(B).$
\end{lemma}
\begin{proof}{Proof of Lemma \ref{lemma:det}.}
Note that
\begin{align}
\nonumber\det(A)=\det(B+(A-B))=&\det\left(B^{\frac{1}{2}}\left(I+B^{-\frac{1}{2}}(A-B)B^{-\frac{1}{2}}\right)B^{\frac{1}{2}}\right)\\
\nonumber=&\det(B)\det\left(\left(I+B^{-\frac{1}{2}}(A-B)B^{-\frac{1}{2}}\right)\right)\\
\label{eq:det1}\geq&\det(B)\left(1+\det\left(B^{-\frac{1}{2}}(A-B)B^{-\frac{1}{2}}\right)\right)\\
\nonumber=&\det(B)+\det(A-B)\\
\label{eq:det2}\geq&\det(B).
\end{align}
Here, inequality \eqref{eq:det1} holds because $\prod_{k=1}^{2d}(1+\mu_k)\geq1+\prod_{k=1}^{2d}\mu_k$ where $\mu_k$ is the $k^{\text{th}}$ eigenvalue of $B^{-\frac{1}{2}}(A-B)B^{-\frac{1}{2}},$ and inequality \eqref{eq:det2} holds because $A-B$ is positive semi-definite. 
\halmos
\end{proof}

\end{APPENDIX}

\end{document}